\documentclass{article}

\usepackage[nonatbib, final]{neurips_2022}

\usepackage[utf8]{inputenc} 
\usepackage[T1]{fontenc}    
\usepackage{hyperref}       
\usepackage{url}            
\usepackage{booktabs}       
\usepackage{amsfonts}       
\usepackage{nicefrac}       
\usepackage{microtype}      
\usepackage{xcolor}         
\usepackage{array,multirow}
\usepackage{textcomp}
\usepackage{multicol} 
\usepackage{rotating}
\usepackage[export]{adjustbox}

\usepackage{graphicx}
\usepackage{amsmath,bm}

\usepackage{algpseudocode}
\usepackage{algorithm, algorithmic}
\newtheorem{theorem}{Theorem}
\newtheorem{lemma}{Lemma}

\usepackage{amsmath,amsfonts,bm}

\def\eqref#1{equation~\ref{#1}}

\def\1{\bm{1}}

\def\rvp{{\mathbf{p}}}

\def\rvv{{\mathbf{v}}}

\def\rvx{{\mathbf{x}}}

\def\rvz{{\mathbf{z}}}

\DeclareMathAlphabet{\mathsfit}{\encodingdefault}{\sfdefault}{m}{sl}
\SetMathAlphabet{\mathsfit}{bold}{\encodingdefault}{\sfdefault}{bx}{n}

\newcommand{\E}{\mathbb{E}}

\newcommand{\KL}[2]{D_{\mathrm{KL}}\left[#1\|#2\right]}

\usepackage{amssymb}
\usepackage{graphicx}
\usepackage{caption}
\usepackage{subcaption}
\usepackage{wrapfig}
\usepackage{adjustbox}
\usepackage{enumitem}

\newcommand{\STAB}[1]{\begin{tabular}{@{}c@{}}#1\end{tabular}}

\newcommand{\Enc}[2]{{q_{\phi}(#1|#2)}}
\newcommand{\HEnc}[2]{{q_{\theta, \phi}(#1|#2)}}
\newcommand{\Dec}[2]{{p_{\theta}(#1|#2)}}

\newcommand{\TV}[2]{{\mathrm{TV}\left[#1 ,#2 \right]}}

\newcommand{\upd}[1]{{\color{black}#1}}
\newcommand{\discussion}[1]{{\color{black}#1}}
\newcommand{\ac}[1]{{\color{black}#1}}

\title{Alleviating Adversarial Attacks on Variational Autoencoders with MCMC}

\author{%
  Anna Kuzina \\
  Vrije Universiteit Amsterdam \\
  \texttt{a.kuzina@vu.nl} \\
   \And
   Max Welling \\
   Universiteit van Amsterdam \\
   \texttt{m.welling@uva.nl} \\
   \And
   Jakub M. ~Tomczak \\
   Vrije Universiteit Amsterdam \\
   \texttt{j.m.tomczak@vu.nl}
}

\begin{document}

\maketitle


\begin{abstract}
Variational autoencoders (VAEs) are latent variable models that can generate complex objects and provide meaningful latent representations. Moreover, they could be further used in downstream tasks such as classification. As previous work has shown, one can easily fool VAEs to produce unexpected latent representations and reconstructions for a visually slightly modified input. Here, we examine several objective functions for adversarial attack construction proposed previously and present a solution to alleviate the effect of these attacks. Our method utilizes the Markov Chain Monte Carlo (MCMC) technique in the inference step that we motivate with a theoretical analysis. Thus, we do not incorporate any extra costs during training, and the performance on non-attacked inputs is not decreased. We validate our approach on a variety of datasets (MNIST, Fashion MNIST, Color MNIST, CelebA) and VAE configurations ($\beta$-VAE, NVAE, $\beta$-TCVAE), and show that our approach consistently improves the model robustness to adversarial attacks.
\end{abstract}

\section{Introduction}\label{sec:intro}

Variational Autoencoders (VAEs) \cite{kingma2013auto, rezende2014stochastic} are latent variable models parameterized by deep neural networks and trained with variational inference. Recently, it has been shown that VAEs with hierarchical structures of latent variables \cite{Ranganath2016-yg}, coupled with skip-connections \cite{Maaloe2019-bp, So_nderby2016-en}, can generate high-quality images \cite{Child2020-ze, Vahdat2020-xe}. An interesting trait of VAEs is that they allow learning meaningful latent space that could be further used in downstream tasks \cite{bengio2013representation, higgins2017darla}. These successes of VAEs motivate us to explore the \textit{robustness} of the resulting latent representations to better understand the capabilities and potential vulnerabilities of VAEs. Here, we focus on \textit{adversarial attacks} on VAEs to verify robustness of latent representations that is especially important in such applications as anomaly detection \cite{an2015variational, Maaloe2019-bp} or data compression \cite{balle2018variational, habibian2019video}. 

The main questions about adversarial attacks for VAEs are mainly focused on how they could be formulated and alleviated. In \cite{Gondim-Ribeiro2018-cu}, it is proposed to minimize the KL-divergence between an adversarial input and a target input to learn an adversarial attack for the vanilla VAE. Further, in \cite{kuzina2021adv}, it is shown that a similar strategy can be used to attack hierarchical VAEs. To counteract the adversarial attacks, the authors of \cite{Willetts2019-mu} suggest using a modified VAE objective, namely, $\beta$-TCVAE, that increases VAE robustness, especially when coupled with a hierarchical structure. It was shown in \cite{camuto2021towards} that $\beta$-VAEs tend to be more robust to adversarial attacks in terms of the $r$-metric proposed therein. The authors of \cite{barrett2021certifiably} presented that the adversarial robustness can be achieved by constraining the Lipschitz constant of the encoder and the decoder. \cite{cemgil2020autoencoding, Cemgil2019-vn} introduced modifications in the VAE framework that allow for better robustness against the adversarial attacks on downstream classification tasks. In our work, we consider $\beta$-VAE, $\beta$-TCVAE and a hierarchical VAE, and outline a defence strategy that improves robustness to attacks on the encoder and the downstream classification task. The proposed method is applied during inference and, therefore, can be combined with other known techniques to get more robust latent representations.

\begin{wrapfigure}{r}{0.53\textwidth}
\begin{center}
\vskip -20pt
\includegraphics[width=0.97\linewidth]{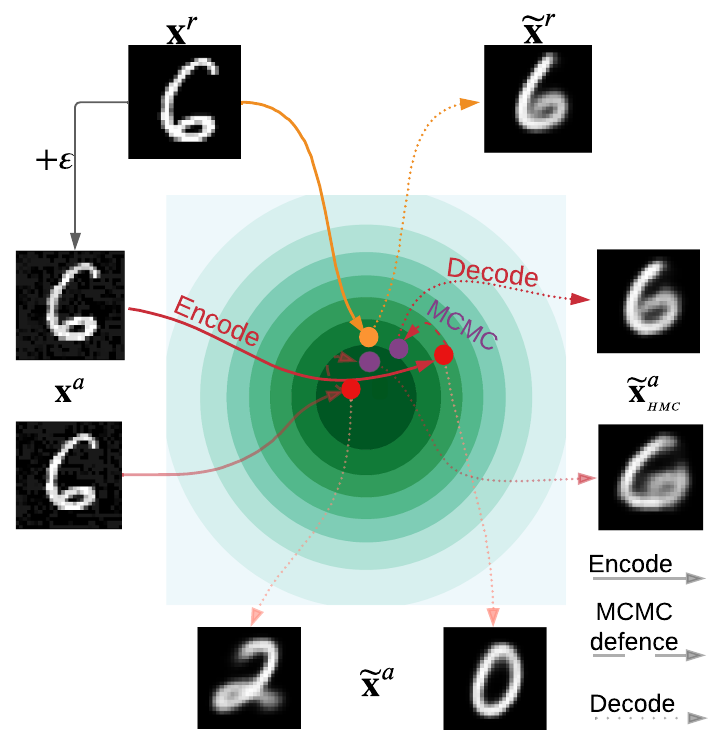}
\caption{An example of an unsupervised encoder attack on VAE with 2D latent space and the proposed defence. Given a single reference point $\rvx^r$ we learn additive perturbation $\varepsilon$, s.t. perturbed input $\rvx^a$ has the most different latent code and, therefore, the reconstruction $\widetilde{\rvx}^a$. We observe that a single reference point can be mapped to extremely different regions of the latent space but using MCMC we are able to move them closer to the initial position so that the reconstruction $\widetilde{\rvx}^a_{\tiny{HMC}}$ is similar to the initial one $\widetilde{\rvx}^r$.}
\label{fig:toy_exaple}
\end{center}
\vskip -20pt
\end{wrapfigure}

An adversarial attack on a VAE is usually formulated as an additive perturbation $\varepsilon$ of the real data point $\rvx^r$ so that the resulting point is perceived by a model as if it is a totally different image (either during reconstruction or in the downstream classification task) \cite{Gondim-Ribeiro2018-cu}. In Figure \ref{fig:toy_exaple}, we depict an example of an attack on the encoder. The reference point $\rvx^r$ and the adversarial point $\rvx^a$ are almost indistinguishable, but they are encoded into different regions in the latent space. As a result, their reconstructions also differ significantly.

In this paper, we propose the method motivated by the following hypothesis: \textit{An adversarial attack maps the input to a latent region with a lower probability mass assigned by the true posterior (proportional to the conditional likelihood times the marginal over latents) and, eventually, we obtain incorrect reconstructions}. Therefore, a potential manner to alleviate the effect of an attack may rely on running a Markov chain to move the latent representation back to a more probable latent region. Such a defence is reasonable because we do not modify the training procedure or the model itself, we only insert a correction procedure. As a result, we propose to counteract adversarial attacks by enhancing the variational inference with Markov Chain Monte Carlo (MCMC) sampling. The \upd{illustrative example} depicted in the Figure \ref{fig:toy_exaple} shows that the latent code of the adversarial input (red circle) moves closer to the latent code of the reference point (orange circle) after applying the MCMC (purple circle). 

The contribution of this work is the following:
\begin{itemize}[leftmargin=*]
    \item We propose to use an MCMC technique during inference to correct adversarial attacks on VAEs.
    \item We show theoretically that the application of an MCMC technique could indeed help to counteract adversarial attacks (Theorem \ref{theorem:main}).
    \item We indicate empirically that the previously proposed strategies to counteract adversarial attacks do not generalize well across various datasets.
    \item We show empirically that the proposed approach (i.e., a VAE with an MCMC during inference) outperforms all baselines by a significant margin.
\end{itemize}



\begin{table}[ht]
\caption{Different types of attacks on the VAE. We denote $g_{\theta}(z)$ the deterministic mapping induced by decoder $p_{\theta}(x|z)$ and as $p_{\psi}(y|z)$ classification model in the latent space (downstream task).\\
\textsuperscript{*} Only used during VAE training}
\vskip -5pt
\label{tab:attack_variation}
\begin{center}
\resizebox{\textwidth}{!}{
\begin{tabular}{llcccc}
\toprule
 & \sc{Reference}  &$f(x)$ & $\Delta\left[A, B\right]$  &  $\|\cdot\|_p$ &  \sc{Type}\\ \midrule
\multirow{3}{*}{Latent Space Attack}
& \multirow{3}{2cm}{\cite{barrett2021certifiably, Gondim-Ribeiro2018-cu, Willetts2019-mu}}
& \multirow{3}{*}{$q_{\phi}(\cdot|x)$} &  \multirow{3}{*}{$\KL{A}{B}$}   & \multirow{3}{*}{2} & \multirow{3}{*}{Supervised}\\
&&&&&\\
&&&&&\\
\multirow{2}{3cm}{Unsupervised Encoder Attack}
& \multirow{2}{2cm}{\cite{kuzina2021adv}}
& \multirow{2}{*}{$q_{\phi}(\cdot|x)$} &  \multirow{2}{*}{$\mathrm{SKL}\left[A\|B\right]$}   & \multirow{2}{*}{2} & \multirow{2}{*}{Unsupervised}\\
&&&&&\\
\multirow{2}{*}{Targeted Output Attack}
& \multirow{2}{*}{\cite{Gondim-Ribeiro2018-cu}}
& \multirow{2}{*}{$g_{\theta}(\tilde{z}), \tilde{z} \sim q_{\phi}(\cdot|x)$}   & 
\multirow{2}{*}{$\|A - B\|_2^2$}  & 
\multirow{2}{*}{2} & \multirow{2}{*}{Supervised}\\
&&&&&\\
\multirow{2}{*}{Maximum Damage Attack}
& \multirow{2}{2cm}{\cite{barrett2021certifiably, camuto2021towards}}
& \multirow{2}{*}{$g_{\theta}(\tilde{z}), \tilde{z} \sim q_{\phi}(\cdot|x)$}   & \multirow{2}{*}{$\|A - B\|_2^2$}  & \multirow{2}{*}{2} & \multirow{2}{*}{Unsupervised}\\
&&&&&\\
\multirow{2}{3cm}{Projected Gradient Descent Attack\textsuperscript{*}}
& \multirow{2}{*}{\cite{Cemgil2019-vn}}
& \multirow{2}{*}{$q_{\phi}(\cdot|x)$} &  \multirow{2}{*}{$\mathcal{WD}\left[A, B\right]$}   & \multirow{2}{*}{$\inf$} & \multirow{2}{*}{Unsupervised}\\
&&&&&\\
\multirow{2}{*}{Adversarial Accuracy}
& \multirow{2}{*}{\cite{cemgil2020autoencoding, Cemgil2019-vn}}
& \multirow{2}{*}{$p_{\psi}(y|\tilde{z}), \tilde{z} \sim q_{\phi}(\cdot|x)$}  & \multirow{2}{*}{\sc{Cross Entropy}} & \multirow{2}{*}{$\inf$} & \multirow{2}{*}{Unsupervised}\\
&&&&&\\
\bottomrule
\end{tabular}}
\end{center}
\vskip -0.25in
\end{table}

\section{Background}

\subsection{Variational Autoencoders}
Let us consider a vector of observable random variables, $\rvx \in \mathcal{X}^{D}$ (e.g., $\mathcal{X} = \mathbb{R}$) sampled from  the empirical distribution $p_{e}(\mathbf{\rvx})$, and vectors of latent variables $\rvz_{k} \in \mathbb{R}^{M_{k}}$, $k=1, 2, \ldots, K$, where $M_k$ is the dimensionality of each latent vector. First, we focus on a model with $K=1$ and the joint distribution $p_{\theta}(\rvx, \rvz) = \Dec{\rvx}{\rvz} p(\rvz)$. The marginal likelihood is then equal to $p_{\theta}(\rvx) = \int p_{\theta}(\rvx, \rvz) \mathrm{d} \rvz$. VAEs exploit variational inference \cite{jordan1999introduction} with a family of variational posteriors $\{\Enc{\rvz}{\rvx}\}$, also referred to as encoders, that results in a tractable objective function, i.e., the Evidence Lower BOund (ELBO): $\mathcal{L}(\phi, \theta)
    = \E_{p_{e}(\mathbf{\rvx})}\left( \E_{\Enc{\rvz}{\rvx}}\ln \Dec{\rvx}{\rvz} - \KL{\Enc{\rvz}{\rvx}}{p(\rvz)} \right)$. 

$\beta$-VAE \cite{higgins2016beta} uses a modified objective by weighting the $D_{\text{KL}}$ term by $\beta > 0$.
In the case of $K>1$, we consider a hierarchical latent structure with the generative model of the following form: $p_{\theta}(\rvx, \rvz_1, \ldots, \rvz_K) = \Dec{\rvx}{\rvz_1, \ldots, \rvz_K} \prod_{k=1}^{K}\Dec{\rvz_{k}}{\rvz_{>k}}$. There are various possible formulations of the family of variational posteriors. However, here we follow the proposition of \cite{So_nderby2016-en} with the autoregressive inference model, namely, $\Enc{\rvz_1, \ldots, \rvz_K}{\rvx} = \Enc{\rvz_K}{\rvx}\prod_{k=1}^{K-1}\HEnc{\rvz_k}{\rvz_{>k}, \rvx}$. 
This formulation was used, among others, in NVAE \cite{Vahdat2020-xe}. Because of the top-down structure, it allows sharing data-dependent information between the inference model and the generative model.

\subsection{Adversarial attacks}

\label{sect:adversarial_attacks}
An \textit{adversarial attack} is a slightly deformed data point that results in an undesired or unpredictable performance of a model \cite{goodfellow2014explaining}. In this work, we focus on the attacks that are constructed as an additive perturbation of the real data point $\rvx^r$ (which we will refer to as \textit{reference}), namely:
\begin{align}
    &\rvx^a = \rvx^r + \varepsilon, \text{where}\\
    &\|\varepsilon\|_{p} \leq \delta ,
\end{align}
where $\delta$ is the radius of the attack. The additive perturbation $\varepsilon$ is chosen in such a way that the attacked point does not differ from the reference point too much in a sense of a given similarity measure. It is a solution to an optimization problem solved by the attacker. The optimization problem could be formulated in various manners by optimizing different objectives and by having specific constraints and/or assumptions.

\paragraph{Attack construction} 
Let $f(x)$ be part of the model available to the attacker. In the case of VAEs, this, for example, may be an encoder network or an encoder with the downstream classifier in the latent space. The attacker uses a similarity measure $\Delta$ to learn an additive perturbation to a reference point. We consider two settings: \textit{unsupervised} and \textit{supervised}. In the former case, the perturbation is supposed to incur the largest possible change in $f$:
\begin{equation}\label{eq:objective_unsup}
    \varepsilon = \arg\max_{\|\varepsilon\|_p \leq \delta }\Delta\left[f(\rvx^r + \varepsilon) , f(\rvx^r) \right].
\end{equation}
The latter setting requires a \textit{target point} $\rvx^t$. The perturbation attempts to match the output for the target and the attacked points, namely:
\begin{equation}\label{eq:objective_sup}
    \varepsilon = \arg\min_{\|\varepsilon\|_p \leq \delta }\Delta\left[ f(\rvx^r + \varepsilon), f(\rvx^t) \right].
\end{equation}
There are different ways to select $f$ and $\Delta$ in the literature. Furthermore, different $L_p$-norms can be used to restrict the radius of the attack. Table \ref{tab:attack_variation} summarizes recent work on the topic. 

In this work, we focus on attacking the encoder and the downstream classification task using unsupervised adversarial attacks. 
In the former, we maximize the symmetric KL-divergence to get the point with the most unexpected latent code and, therefore, the reconstruction. In the latter, the latent code is passed to a classifier. The attack is trained to change the class of the point by maximizing the cross-entropy loss. However, the method we propose in Section \ref{sec:defence} is not limited to these setups since it is agnostic to how the attack was trained. 

\paragraph{Robustness measures} 
To measure the robustness of the VAE as well as the success of the proposed defence strategy, we focus on two metrics: $\mathrm{MSSSIM}$ and Adversarial accuracy.

For latent space attacks we follow \cite{kuzina2021adv} in using Multi-Scale Structural Similarity Index Measure ($\mathrm{MSSSIM}$) \cite{wang2003multiscale}. We calculate $\mathrm{MSSSIM}[\widetilde{\rvx}^{r}, \widetilde{\rvx}^{a}]$, i.e., the similarity between reconstructions of $\rvx^{r}$ and the corresponding $\rvx^{a}$. We do not report the similarity between a reference and the corresponding adversarial input, since this value is the same for all the considered models (for a given attack radius). A successful adversarial attack would have a small value of  $\mathrm{MSSSIM}[\widetilde{\rvx}^{r}, \widetilde{\rvx}^{a}]$. 

For the attacks on the downstream classifier, we follow \cite{cemgil2020autoencoding, Cemgil2019-vn} and calculate the adversarial accuracy. For a given trained VAE model, we first train a linear classifier using latent codes as features. Afterwards, the attack is trained to fool the classifier. Adversarial accuracy is the proportion of points for which the attack was unsuccessful. Namely, when the predicted class of the reference and corresponding adversarial point are the same.  

\section{Preventing adversarial attacks with MCMC} \label{sec:defence}

\paragraph{Assumptions and the hypothesis}

We consider a scenario in which an attacker has access to the VAE encoder and, where relevant, to the downstream classification model. Above, we presented in detail how an attack could be performed. We assume that the defender cannot modify these components, but it is possible to add elements that the attacker has no access to. 

We hypothesize that \textit{the adversarial attacks result in incorrect reconstructions because the latent representation of the adversarial input lands in a region with a lower probability mass assigned by the true posterior. 
}
\ac{This motivates us to use a method which} 
brings the latents "back" to a highly probable region \ac{as a potential defence strategy}. Since the adversarial attack destroys the input irreversibly, at the first sight it seems impossible to reconstruct the latent representation of the reference point. We aim at showing that this is possible to some degree \upd{theoretically and empirically (Appendix \ref{appendix:posterior_ratio})}. 

\paragraph{The proposed solution } In order to steer the latents towards high probability regions, we propose to utilize an MCMC method during inference time. Since we are not allowed to modify the VAE or its learning procedure, this is a reasonable procedure from the defender's perspective. Another positive outcome of such an approach is that in a case of no attack, the latents given by the MCMC sampling will be closer to the mean of the posterior, thus, the reconstruction should be sharper or, at least, not worse. Note that the variational posterior approximates the true posterior from which the MCMC procedure samples. We propose to sample from $q^{(t)}(\rvz|\rvx) = \int Q^{(t)}(\rvz|\rvz_0) q_{\phi}(\rvz_0|\rvx) d\rvz_0$, where $Q^{(t)}(\rvz|\rvz_0)$ is a transition kernel of MCMC with \upd{$t$ steps and} the target distribution $\pi(\rvz) = p_{\theta}(\rvz|\rvx) \propto p_{\theta}(\rvx|\rvz)p(\rvz)$. \discussion{Alternatively, we can use optimization to find the mode of the posterior, however, MCMC can add extra benefits such as exploration of the typical set and randomness (see Appendix \ref{appendix:hmc_vs_opt} for details).}

To further analyze the proposed approach, we start with showing the following lemma:

\begin{lemma}\label{lemma:1}
Consider true posterior distributions of the latent code $\rvz$ for a data point $\rvx$ and its corrupted version $\rvx^a$.  Assume also that $\ln p_{\theta}(\rvz|\rvx)$ \ac{is twice differentiable  over $\rvx$ with continuous derivatives at the neighbourhood around $\rvx=\rvx^r$.} 
Then the KL-divergence between these two posteriors could be expressed using the small $o$ notation of the radius of the attack, namely:
\begin{equation}
     \KL{ p_{\theta}(\rvz|\rvx^r)}{p_{\theta}(\rvz|\rvx^a)} = o(\|\varepsilon\|).
\end{equation}
\end{lemma}
\textit{Proof} See Appendix \ref{appendix:theory}.

According to Lemma \ref{lemma:1}, the difference (in the sense of the Kullback-Leibler divergence) between the true posteriors for a data point $\rvx$ and its corrupted version $\rvx^{a}$ \ac{decreases faster than the norm of the attack radius. However, it is important to note that this result is only valid for asymptotically small attack radius.}

Next, the crucial step to show is whether we can quantify somehow the difference between the distribution over latents for $\rvx^a$ after \upd{running MCMC with $t$ steps}, $q^{(t)}(\rvz|\rvx^a)$, and the variational distribution for $\rvx^r$, $q_{\phi}(\rvz|\rvx^r)$. We provide an important upper-bound for the Total Variation distance (TV)\footnote{The Total Variation fulfills the triangle inequality and it is a proper distance measure.} between $q^{(t)}(\rvz|\rvx^a)$ and $q_{\phi}(\rvz|\rvx^r)$ in the following lemma:

\begin{lemma}\label{lemma:2}
The Total Variation distance ($\mathrm{TV}$) between the variational posterior with MCMC for a given corrupted point $\rvx^a$, $q^{(t)}(\rvz|\rvx^a)$, and the variational posterior for a given data point $\rvx^r$, $q_{\phi}(\rvz|\rvx^r)$, can be upper bounded by the sum of the following three components:
\begin{align}
\TV{q^{(t)}(\rvz|\rvx^a)}{q_{\phi}(\rvz|\rvx^r)}
&\leq
    \TV{q^{(t)}(\rvz|\rvx^a)}{p_{\theta}(\rvz|\rvx^a)} \notag\\
    &+ 
    \sqrt{\tfrac12 \KL{ p_{\theta}(\rvz|\rvx^r)}{p_{\theta}(\rvz|\rvx^a)} } \notag\\
    &+
   \sqrt{\tfrac12  \KL{q_{\phi}(\rvz|\rvx^r)}{p_{\theta}(\rvz|\rvx^r)} }.
\end{align}
\end{lemma}
\textit{Proof} See Appendix \ref{appendix:theory}.

The difference expressed by $\TV{q^{(t)}(\rvz|\rvx^a)}{q_{\phi}(\rvz|\rvx^r)}$ is thus upper-bounded by the following three components:
\begin{itemize}
    \item The $\mathrm{TV}$ between  $q^{(t)}(\rvz|\rvx^a)$ and the real posterior for the corrupted image, $p_{\theta}(\rvz|\rvx^a)$. Theoretically, if $t \rightarrow \infty$, $q^{(\infty)}(\rvz|\rvx^a) = p_{\theta}(\rvz|\rvx^a)$ and, hence, $\TV{ q^{(\infty)}(\rvz|\rvx^a)}{p_{\theta}(\rvz|\rvx^a) } = 0$.
    \item The second component is the square root of the KL-divergence between the real posteriors for the image and its corrupted counterpart. Lemma \ref{lemma:1} gives us information about this quantity.
    \item The last element, the square root of $\KL{ q_{\phi}(\rvz|\rvx^r)}{p_{\theta}(\rvz|\rvx^r)}$, quantifies the \textit{approximation gap} \cite{cremer2018inference}, i.e., the difference between the best variational posterior from a chosen family, and the true posterior. This quantity has no direct connection with adversarial attacks. However, as we can see, using a rich family of variational posteriors can help us to obtain a tighter upper-bound. In other words, taking flexible variational posteriors allows to counteract attacks. This finding is in line with the papers that propose to use hierarchical VAEs as the means for preventing adversarial attacks \cite{Willetts2019-mu}.
\end{itemize}

Eventually, by applying Lemma \ref{lemma:1} to Lemma \ref{lemma:2}, we obtain the following result:

\begin{theorem}\label{theorem:main}
The upper bound on the total variation distance between samples from MCMC for a given corrupted point $\rvx^a$, $q^{(t)}(\rvz|\rvx^a)$, and the variational posterior for the given real point $\rvx^r$, $q_{\phi}(\rvz|\rvx^r)$, is the following:
\begin{align} \label{eq:theorem_1_main}
    \TV{ q^{(t)}(\rvz|\rvx^a)}{q_{\phi}(\rvz|\rvx^r) }
&\leq 
\TV{ q^{(t)}(\rvz|\rvx^a)}{p_{\theta}(\rvz|\rvx^a) } \notag \\
    &+ 
   \sqrt{\tfrac12  \KL{ q_{\phi}(\rvz|\rvx^r) }{ p_{\theta}(\rvz|\rvx^r)} } \notag\\
   &+ o(\sqrt{\|\varepsilon\|}).
\end{align}
\end{theorem}
\textit{Proof} See Appendix \ref{appendix:theory}.

As discussed already, the first component gets smaller with more steps of the MCMC. 
The second component could be treated as a \textit{bias} of the family of variational posteriors. Finally, there is the last element that corresponds to a constant error that is unavoidable. However, this term decays faster than the square root of the attack radius \ac{for the asymptotically small attack radius}.

\paragraph{Specific implementation of the proposed approach}
In this paper, we use a specific MCMC method, namely, the Hamiltonian Monte Carlo (HMC) \cite{betancourt2017conceptual, duane1987hybrid}. Once the VAE is trained, the attacker calculates an adversarial point $\rvx^a$ using the encoder of the VAE. After the attack, the latent representation of $\rvx^a$ is calculated, $\rvz^a$, and used as the initialization of the HMC.

In the HMC, the target (unnormalized) distribution is $p(\mathbf{x}^{a}|\mathbf{z}) p(\mathbf{z})$. The Hamiltonian is then the energy of the joint distribution of $\rvz$ and the auxilary variable $\rvp$, that is:
\begin{align}
    H(\rvz, \rvp) &= U(\rvz) + K(\rvp),\\
    U(\rvz) &= - \ln p_{\theta}(\mathbf{x}^{a}|\mathbf{z}) - \ln p(\mathbf{z}),\\
    K(\rvp) &= -\tfrac12 \rvp^T\rvp.
\end{align}
When applying the proposed defence to hierarchical models, we update all the latent variables simultaneously. That is, we have $\rvz = \{ \rvz_1, \dots, \rvz_K\}$ and
$U(\rvz) = -\ln \Dec{\rvx^a}{\rvz} - \sum_{k=1}^K\ln p_{\theta}(\rvz_k|\rvz_{k+1}). $

\begin{wrapfigure}[17]{R}{0.53\textwidth}
\begin{minipage}{0.54\textwidth}
\vskip -11pt
\begin{algorithm}[H]
	\caption{One Step of HMC.}
	\label{alg:hmc_step}
	\begin{algorithmic} 
	   \State \hskip-3mm \textbf{Input}: $\rvz, \eta, L$
		\State $\rvp \sim \mathcal{N}(0,I)$. \Comment{Sample the auxiliary variable }
		\State $\rvz^{(0)} := \rvz, \rvp^{(0)}:=\rvp$.
		\For{$l = 1 \dots, L$} \Comment{Make $L$ steps of \textit{leapfrog}.}
        \State $ \rvp^{(l)} = \rvp^{(l-1)} - \frac{\eta}{2}\nabla_{\rvz} U(\mathbf{z}^{(l)})$.
		\State $\rvz^{(l)} = \rvz^{(l)} + \eta  \nabla_{\rvp} K(\rvp^{(l)})$.
		\State $\rvp^{(l)} = \rvp^{(l)} - \frac{\eta}{2}\nabla_{\rvz} U(\mathbf{z}^{(l)})$.
		\EndFor
		\State \Comment{Accept new point with prob. $\alpha$.}
		\State $\alpha = \min\left(1, \exp\left(- H(\rvz^{(L)}, \rvp^{(L)}) + H(\rvz^{(0)}, \rvp^{(0)})\right) \right)$
		\State $\rvz =
\begin{cases}
\rvz^{(L)}\, \text{with probability}\, \alpha ,\\
\rvz^{(0)}\, \text{otherwise}.
\end{cases}$ 
        \State  \hskip-3mm \textbf{Return}: $\rvz$
	\end{algorithmic}
\end{algorithm}
\end{minipage}
\end{wrapfigure}
Eventually, the resulting latents from the HMC are decoded. The steps of the whole process are presented below:
\begin{enumerate}[wide, labelwidth=0pt, labelindent=0pt]
    \item (\textit{Defender}) Train a VAE: $q_{\phi}(\rvz|\rvx)$, $p(\rvz)$, $p_{\theta}(\rvx|\rvz)$.
    \item (\textit{Attacker}) For given $\rvx^r$, calculate the attack $\rvx^{a}$ using the criterion \eqref{eq:objective_unsup} or \eqref{eq:objective_sup}.
    \item (\textit{Defender}) Initialize the latent code $\rvz := \rvz_0$, where $\rvz_{0} \sim q_{\phi}(\rvz|\rvx^{a})$. Then, run $T$ steps of HMC (Algorithm \ref{alg:hmc_step}) with the step size $\eta$ and $L$ \textit{leapfrog} steps.
\end{enumerate}

The resulting latent code $\rvz$ can be passed to the decoder to get a reconstruction or to the downstream classification task.



\section{Experiments} \label{sec:experiments}
\subsection{Posterior ratio}\label{sec:exp_ratio}

\begin{wrapfigure}{r}{0.55\textwidth}
    \centering
        \includegraphics[width=0.45\textwidth]{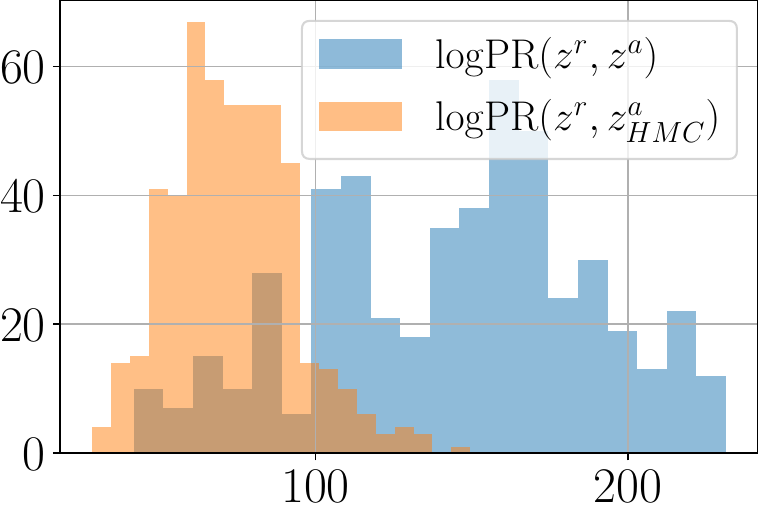}
    \caption{Histograms of the log posterior ratios \textcolor{blue}{before HMC (blue)} and \textcolor{orange}{after HMC (orange)} evaluated on the MNIST dataset.}
    \label{fig:mnist_post_ratio_main}
\end{wrapfigure}
We motivate our method by the hypothesis that the adversarial attack "shifts" a latent code to the region of a lower posterior density, while our approach moves it back to a high posterior probability region. In Section \ref{sec:defence} we theoretically justify our hypothesis, while here we provide an additional empirical evidence.
The true posterior $p(\rvz | \rvx^{r})$ is not available due to the cumbersome marginal distribution $p(\rvx^{r})$, however, we can calculate the ratio of posteriors because the marginal will cancel out. In our case, we are interested in calculating the posterior ratio between the reference and adversarial latent codes ($\rvz_1 = \rvz^r$,  $\rvz_2 = \rvz^a$) as the baseline, and the posterior ratio between the reference and adversarial code after applying the HMC ($\rvz_1 = \rvz^r$ , $\rvz_2 = \rvz^a_{\text{HMC}}$). The lower the posterior ratio, the better. For practical reasons, we use the logarithm of the posterior ratio since the logarithm does not change the monotonicity and turns products to sums:
\begin{equation}
    \log \text{PR}(\rvz_1, \rvz_2) = \log p_{\theta}(\rvx^r|\rvz_1) + \log p(\rvz_1) - \log p_{\theta}(\rvx^r| \rvz_2) - \log p(\rvz_2) .
\end{equation}

In Figure \ref{fig:mnist_post_ratio_main} we show a plot with two histograms: one with the posterior ratio between the reference and adversarial latent codes ($\rvz_1 = \rvz^r$,  $\rvz_2 = \rvz^a$) in blue, and the second histogram of the posterior ratio between the reference and adversarial code after applying the HMC ($\rvz_1 = \rvz^r$ , $\rvz_2 = \rvz^a_{\text{HMC}}$) in orange. We observe that the histogram has moved to the left after applying the HMC. This indicates that posterior of the adversarial (in the denominator) is increasing when the HMC is used. This is precisely the effect we hoped for and this result provides an empirical evidence in favor of our hypothesis. For more details see \ref{appendix:posterior_ratio}.


\subsection{VAE, $\beta$-VAE and $\beta$-TCVAE} 
All implementation details and hyperparameters are included in the Appendix ~\ref{appendix:experimental_details} and code repository~\footnote{\url{https://github.com/AKuzina/defend_vae_mcmc}}. 
\paragraph{Datasets}
VAEs are trained on the MNIST, Fashion MNIST \cite{xiao2017fashion} and Color MNIST datasets. Following \cite{Cemgil2019-vn}, we construct the Color MNIST dataset from MNIST by artificially coloring each image with seven colors (all corners of RGB cube except for black). 

\begin{wrapfigure}{r}{0.55\textwidth}
\vskip -5pt
    \begin{tabular}{lllll}
        \includegraphics[width=0.16\linewidth]{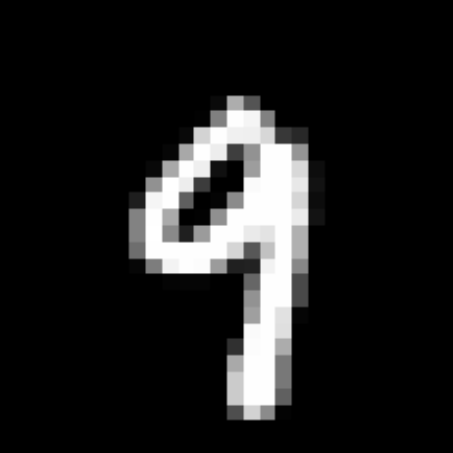} & \includegraphics[width=0.16\linewidth]{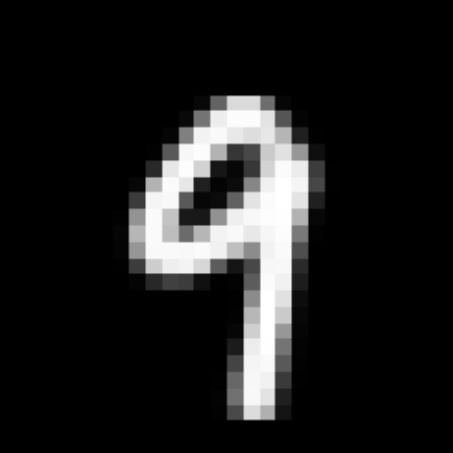} &
        \includegraphics[width=0.16\linewidth]{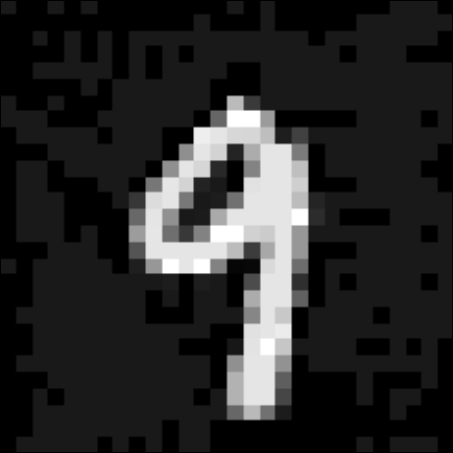} & \includegraphics[width=0.16\linewidth]{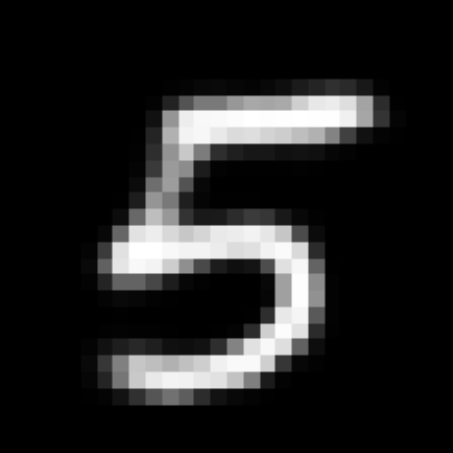} & 
        \includegraphics[width=0.16\linewidth]{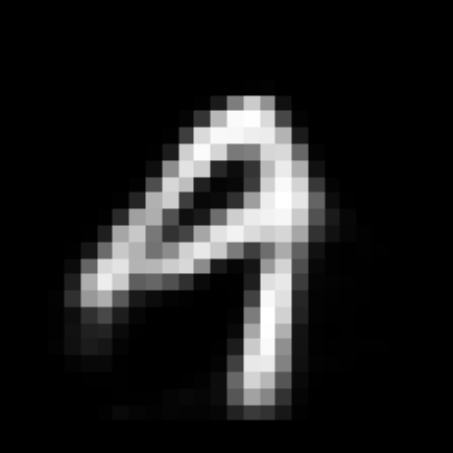}\\
        \includegraphics[width=0.16\linewidth]{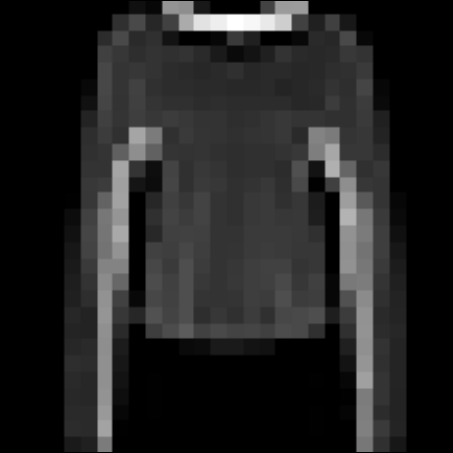} & \includegraphics[width=0.16\linewidth]{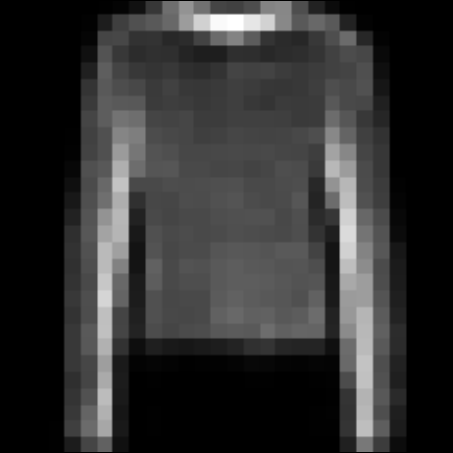} &
        \includegraphics[width=0.16\linewidth]{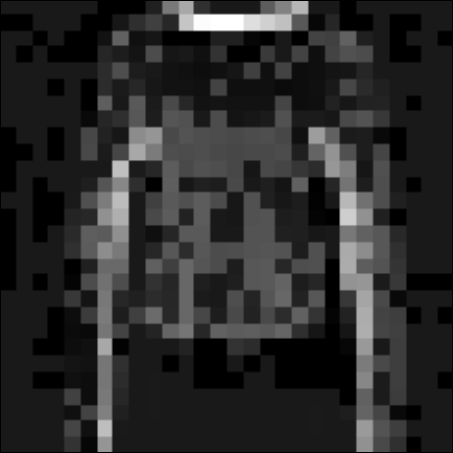} & \includegraphics[width=0.16\linewidth]{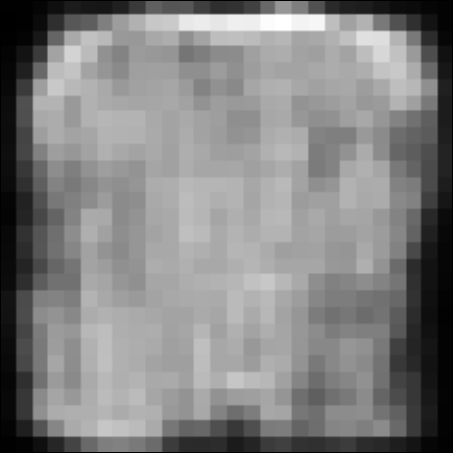} & 
        \includegraphics[width=0.16\linewidth]{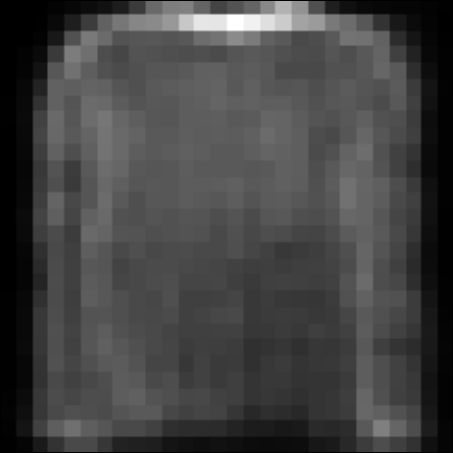}\\
        \includegraphics[width=0.16\linewidth]{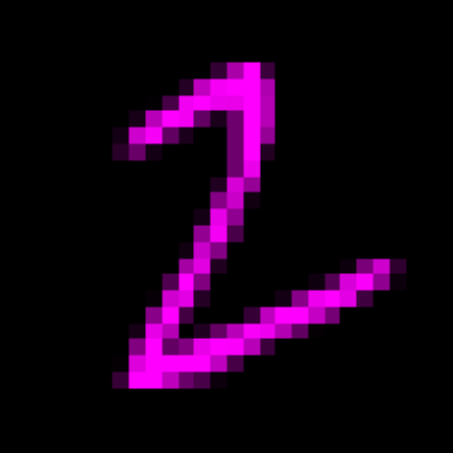} & \includegraphics[width=0.16\linewidth]{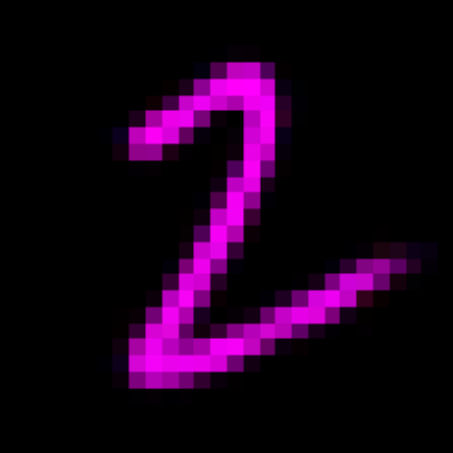} &
        \includegraphics[width=0.16\linewidth]{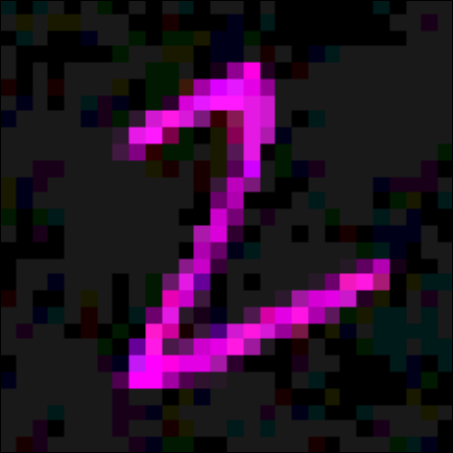} & \includegraphics[width=0.16\linewidth]{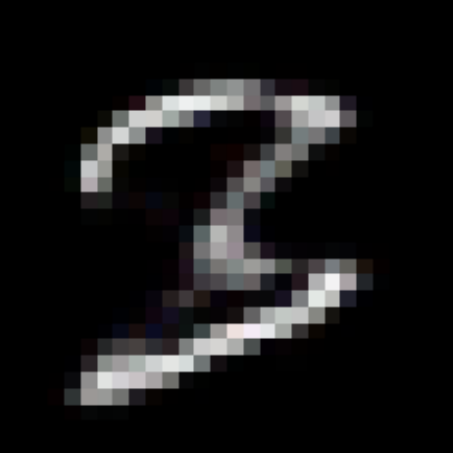} & 
        \includegraphics[width=0.16\linewidth]{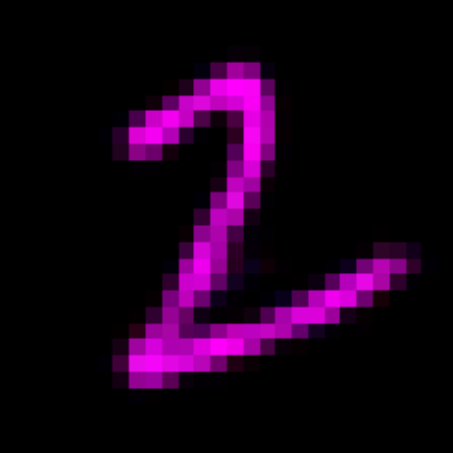}\\
        \includegraphics[width=0.16\linewidth]{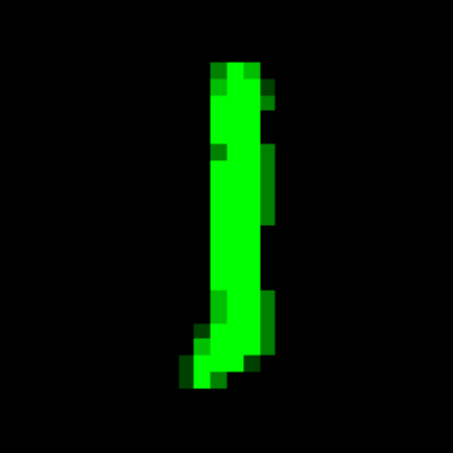} & \includegraphics[width=0.16\linewidth]{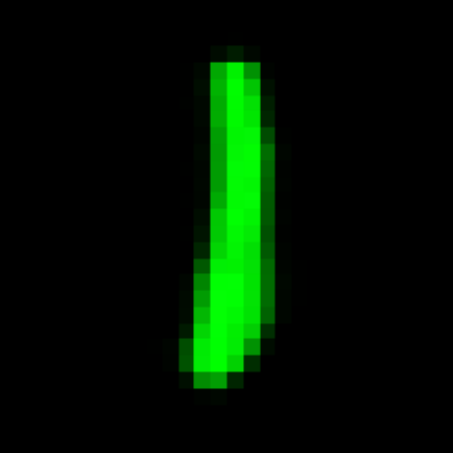} &
        \includegraphics[width=0.16\linewidth]{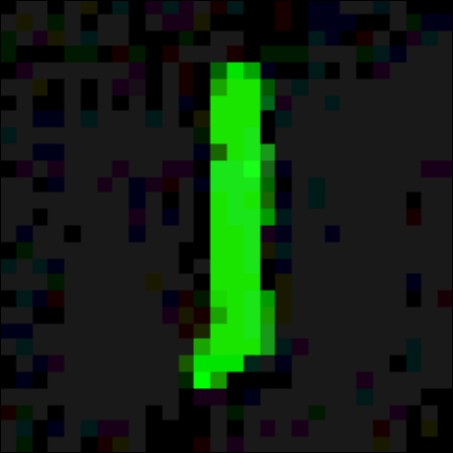} & \includegraphics[width=0.16\linewidth]{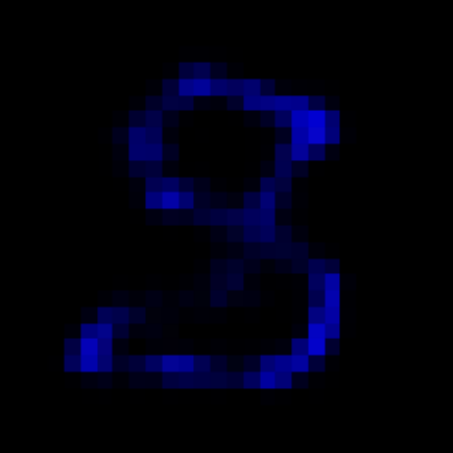} & 
        \includegraphics[width=0.16\linewidth]{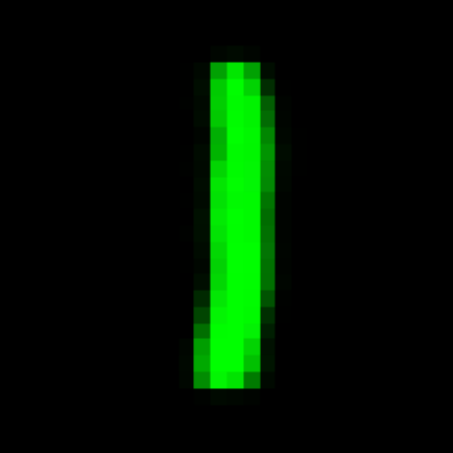}\\
        \includegraphics[width=0.16\linewidth]{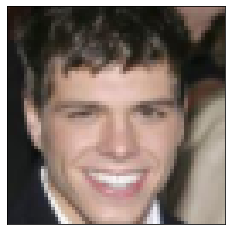} & \includegraphics[width=0.16\linewidth]{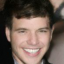} &
        \includegraphics[width=0.16\linewidth]{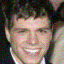} & \includegraphics[width=0.16\linewidth]{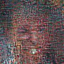} & 
        \includegraphics[width=0.16\linewidth]{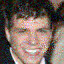}\\
        \includegraphics[width=0.16\linewidth]{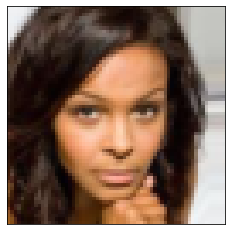} & \includegraphics[width=0.16\linewidth]{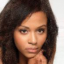} &
        \includegraphics[width=0.16\linewidth]{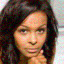} & \includegraphics[width=0.16\linewidth]{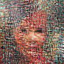} & 
        \includegraphics[width=0.16\linewidth]{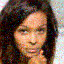}\\
        (a) $\rvx^r$ & (b) $\widetilde{\rvx}^r$ & (c) $\rvx^a$ & (d) $\widetilde{\rvx}^a$ & (e) $\widetilde{\rvx}^\text{a}_{\text{HMC}}$ \\
    \end{tabular}
    \caption{Examples of (a) reference points, (b) reconstructions of the reference points, (c) adversarial points, (d) reconstructions of the adversarial points, (e) reconstructions of the adversarial points after the proposed defence (HMC). All the adversarial examples are unsupervised attacks on the encoder. \upd{Last two rows contain examples for the NVAE model discussed in Section \ref{sec:nvae}}}
    \vskip -15pt
    \label{fig:adversarial_examples}
\end{wrapfigure}

\paragraph{Models}
We train vanilla fully convolutional VAEs, as well as $\beta$-VAE \cite{higgins2016beta} and $\beta$-TCVAE \cite{chen2018isolating, kim2018disentangling}. Both $\beta$-VAE and $\beta$-TCVAE modify the ELBO objective to encourage disentanglement. $\beta$-VAE weighs the KL-term in the ELBO with $\beta > 0$. It is said that the larger values of $\beta$ encourage disentangling of latent representations \cite{chen2018isolating} and improve the model robustness as observed by \cite{camuto2021towards}. $\beta$-TCVAE puts a higher weight on the total correlation (TC) term of the ELBO. Penalization of the total correlation was shown to increase the robustness of VAE adversarial attacks \cite{Willetts2019-mu}.

In Appendix \ref{appendix:beta_vae_param} we provide details of the architecture, optimization, and results on the test dataset for VAE trained with different values of $\beta$. 
We note that the optimal value in terms of the negative log-likelihood (NLL) is always $\beta=1$. Larger values of $\beta$ are supposed to improve robustness in exchange for the reconstruction quality. When evaluating the robustness of $\beta$-VAE and $\beta$-TCVAE, we train models with $\beta \in \{2, 5, 10\}$. Then, we select the value of $\beta$ that provides the most robust model in terms of the used metric. Next, we apply our defence strategy to this model to observe the potential performance improvement.

\begin{table*}[h]
\caption{Results for unsupervised attack with radius $0.1$ and $0.2$ on MNIST and Fashion MNIST datasets. We attack the encoder (left) and the downstream classification task (right).
\\\textsuperscript{$\dagger$} Our implementation.}
\vskip -0.2cm
\label{tab:mnist_attack_result}
\begin{center}
\begin{small}
\begin{sc}
\resizebox{1.0\textwidth}{!}{
\begin{tabular}{lllcc|ccc|ll}
\toprule
& & & \multicolumn{2}{c|}{$\mathrm{MSSSIM}[\widetilde{\rvx}^{r}, \widetilde{\rvx}^{a}]$ $\uparrow$} & \multicolumn{3}{c|}{Adversarial Accuracy $\uparrow$} & \multirow{2}{*}{MSE \upd{$\downarrow$}}  \\
\multicolumn{3}{c}{\upd{$\|\varepsilon\|$}} & 0.1 & 0.2 & 0.0 & 0.1 & 0.2           & &\\ 
\midrule
 & \multirow{3}{*}{\STAB{\rotatebox[origin=c]{90}{\small{MNIST}}}} 
 & VAE & 0.70 \tiny{(0.02)} & 0.36 \tiny{(0.03)} & \textbf{0.90} \tiny{(0.04)} & 0.08 \tiny{(0.04)} & 0.05 \tiny{(0.03)} & 578.7 \\
&& \textbf{VAE + HMC} & \textbf{0.88} \tiny{(0.01)} & \textbf{0.76} \tiny{(0.02)} & \upd{0.76 \tiny{(0.01)}} & \textbf{0.25} \tiny{(0.03)} & \textbf{0.19} \tiny{(0.03)} & \upd{\textbf{478.1}} 
\\
&& $\beta$-VAE & 0.75 \tiny{(0.01)} & 0.50 \tiny{(0.03)} & 0.90 \tiny{(0.05)} & 0.11 \tiny{(0.04)} & 0.01 \tiny{(0.01)} & 824.2 \\
&& $\beta$-TCVAE\textsuperscript{$\dagger$} 
& 0.70 \tiny{(0.02)} & 0.46 \tiny{(0.03)} & 0.86 \tiny{(0.05)} & 0.05 \tiny{(0.03)} & 0.03 \tiny{(0.02)} & 828.4
\\ \midrule 
\multirow{3}{*}{\STAB{\rotatebox[origin=c]{90}{\small{Fashion}}}} &\multirow{3}{*}{\STAB{\rotatebox[origin=c]{90}{\small{MNIST}}}} 
&VAE           & 0.59 \tiny{(0.03)} & 0.47 \tiny{(0.03)} & 0.78 \tiny{(0.06)} & 0.00 \tiny{(0.01)} & 0.01 \tiny{(0.01)} & 814.2  \\
&& \textbf{VAE + HMC}    & \textbf{0.66} \tiny{(0.03)} & \textbf{0.54} \tiny{(0.03)} & \upd{0.56 \tiny{(0.01)}} & \textbf{0.14} \tiny{(0.02)} & \textbf{0.13} \tiny{(0.02)} & \upd{\textbf{764.2}} \\ 
&& $\beta$-VAE & 0.52 \tiny{(0.03)} & 0.41 \tiny{(0.03)} & 0.80 \tiny{(0.05)} & 0.00 \tiny{(0.01)} & 0.00 \tiny{(0.01)} & 
1021.1\\
&& $\beta$-TCVAE\textsuperscript{$\dagger$} & 0.52 \tiny{(0.03)} & 0.42 \tiny{(0.03)} & \textbf{0.84} \tiny{(0.05)} & 0.00 \tiny{(0.01)} & 0.02 \tiny{(0.02)} & 980.4\\ 
\bottomrule
\end{tabular}
}
\end{sc}
\end{small}
\end{center}
\vskip -5pt
\end{table*}


\begin{table*}[ht]
\caption{Results for unsupervised attack with radius $0.1$ and $0.2$ on ColorMNIST dataset. We attack the encoder (left) and the downstream classification task (right).
\\ \textsuperscript{$\dagger$} Our implementation. \\
\textsuperscript{*} Values reported in \cite{cemgil2020autoencoding}, VAE implementation and evaluation protocol may differ. }
\vskip -0.2cm
\label{tab:color_mnist_attack_results}
\begin{center}
\begin{small}
\begin{sc}
\resizebox{1.03\textwidth}{!}{
\begin{tabular}{lcc|cccccc|ll}
\toprule
 & \multicolumn{2}{c|}{$\mathrm{MSSSIM}[\widetilde{\rvx}^{r}, \widetilde{\rvx}^{a}]$ $\uparrow$} & \multicolumn{6}{c|}{Adversarial Accuracy $\uparrow$} & \multirow{3}{*}{MSE\upd{$\downarrow$}} &  \multirow{3}{*}{FID\upd{$\downarrow$}} \\
 &     &     & \multicolumn{3}{c}{Digit} &  \multicolumn{3}{c|}{Color} & & \\
 $\|\varepsilon\|$& 0.1 & 0.2 & 0.0 & 0.1 & 0.2           & 0.0 & 0.1 & 0.2             & & \\ \midrule
VAE & 0.36 \tiny{(0.03)} & 0.19 \tiny{(0.02)} & \textbf{1.00 \tiny{(0.00)}} & 0.04 \tiny{(0.03)} & 0.02 \tiny{(0.02)} & 1.00 \tiny{(0.00)} & 0.06 \tiny{(0.03)} & 0.06 \tiny{(0.03)} & \textbf{261} & \textbf{2.1}\\
\textbf{VAE + HMC}
& \textbf{0.96 \tiny{(0.01)}} & \textbf{0.90 \tiny{(0.01)}} & \upd{0.42 \tiny{(0.01)}} & 0.16 \tiny{(0.02)} & 0.11 \tiny{(0.02)} & \upd{1.00 \tiny{(0.00)}} & 0.68 \tiny{(0.03)} & 0.62 \tiny{(0.03)} & \textbf{\upd{206}} & \textbf{2.1}\\
\midrule
$\beta$-VAE 
& 
0.75 \tiny{(0.01)} & 0.5 \tiny{(0.03)} & 0.88 \tiny{(0.04)} & 0.08 \tiny{(0.04)} & 0.05 \tiny{(0.03)} & 1.00 \tiny{(0.00)} & 0.21 \tiny{(0.06)} & 0.18 \tiny{(0.05)} & 366 & 2.4\\
$\beta$-TCVAE\textsuperscript{$\dagger$} 
& 
0.35 \tiny{(0.02)} & 0.23 \tiny{(0.02)} & 0.94 \tiny{(0.04)} & 0.08 \tiny{(0.04)} & 0.05 \tiny{(0.03)} & 1.00 \tiny{(0.00)} & 0.06 \tiny{(0.03)} & 0.05 \tiny{(0.02)} & 366 & 3.0\\
$\text{SE}_{0.1}$\textsuperscript{*}
& N/A                & N/A                & 0.94 \tiny{(N/A)}  & 0.89 \tiny{(N/A)}  & 0.02 \tiny{(N/A)}  & 1.00 \tiny{(N/A)}  & 1.00 \tiny{(N/A)}  & 0.22 \tiny{(N/A)}  & 1372 & 13.0\\
$\text{SE}_{0.2}$\textsuperscript{*}
& N/A                & N/A                & 0.95 \tiny{(N/A)}  & 0.92 \tiny{(N/A)}  & \textbf{0.87 \tiny{(N/A)}}  & 1.00 \tiny{(N/A)}  & 1.00 \tiny{(N/A)}  & \textbf{1.00 \tiny{(N/A)}}  & 1375 & 11.7 \\
AVAE\textsuperscript{*} 
& N/A                & N/A                & 0.97 \tiny{(N/A)}  & 0.88 \tiny{(N/A)}  & 0.55 \tiny{(N/A)}  & 1.00 \tiny{(N/A)}  & 1.00 \tiny{(N/A)} & 0.88 \tiny{(N/A)}  & 1372 & 15.5\\
$\text{SE}_{0.1}$-AVAE\textsuperscript{*}
& N/A                & N/A                & 0.97 \tiny{(N/A)}  & \textbf{0.94 \tiny{(N/A)}}  & 0.25 \tiny{(N/A)}  & 1.00 \tiny{(N/A)}  & 1.00 \tiny{(N/A)}  & 0.60 \tiny{(N/A)}  & 1373 & 13.9\\
$\text{SE}_{0.2}$-AVAE\textsuperscript{*} 
& N/A                & N/A                & 0.98 \tiny{(N/A)}  & \textbf{0.94 \tiny{(N/A)}}  & 0.80 \tiny{(N/A)}  & 1.00 \tiny{(N/A)}  & 1.00 \tiny{(N/A)}  & 0.83 \tiny{(N/A)}  & 1374 & 13.9\\
AVAE-SS\textsuperscript{*} 
& N/A                & N/A                & 0.94 \tiny{(N/A)}  & 0.73 \tiny{(N/A)}  & 0.21 \tiny{(N/A)}  & 1.00 \tiny{(N/A)}  & 1.00 \tiny{(N/A)}  & 0.57 \tiny{(N/A)}  & 1379 & 12.4\\
\bottomrule
\end{tabular}}
\end{sc}
\end{small}
\end{center}
\vskip -10pt
\end{table*}

\paragraph{Attacks on the Encoder} 
In the first setup, we assume that the attacker has access to the encoder of the model $\Enc{z}{x}$ {\cite{barrett2021certifiably, Gondim-Ribeiro2018-cu, Willetts2019-mu}. We use the projected gradient descent (PGD) with 50 steps to maximize the symmetric KL-divergence in the unsupervised setting (\eqref{eq:objective_unsup}). We train 10 adversarial attacks (with different random initialization) for each of 50 reference points. See Appendix \ref{appendix:vae_attack} for the details. We report similarity between the reconstruction of the adversarial and reference point as a measure of the robustness (see Section \ref{sect:adversarial_attacks}). 

\paragraph{Attacks on the downstream task} 
In this setup, we examine how the proposed approach can aleviate the effect of the attack on the downstream tasks in the latent space. For this purpose, we follow the procedure from \cite{cemgil2020autoencoding, Cemgil2019-vn}. Once the VAE is trained, we learn a linear classifier using the mean mappings as features. For the MNIST and Fashion MNIST datasets, we have the 10-class classification problem (digits in the former and pieces of clothing in the latter case). For the ColorMNIST dataset, we consider two classification tasks: the digit classification (10 classes), and the color classification (7 classes). We construct the attack to fool the classifier. See Appendix \ref{appendix:vae_attack} for the details.


\paragraph{Results} 
In Tables \ref{tab:mnist_attack_result} and \ref{tab:color_mnist_attack_results}, we compare our method (VAE + HMC) with the vanilla VAE with other methods in the literature. We report more results and the extended comparison in the Appendix \ref{appendix:beta_vae} where we show that our method combined with $\beta$-VAE and $\beta$-TCVAE leads to the increased robustness. For MNIST and Fashion MNIST (Table \ref{tab:mnist_attack_result}), we observe that the vanilla VAE with the HMC is more robust than $\beta$-VAE and $\beta$-TCVAE. The latter model was shown to be more robust to the latent space attack \cite{Willetts2019-mu}. Still, in our experiments (on different datasets), we could not observe the consistent improvement over the vanilla VAE, when using it with a single level of latent variables. 

In Table \ref{tab:color_mnist_attack_results} we report the result on the ColorMNIST dataset. Here, we additionally compare the adversarial accuracy for our method with the Smooth Encoders (SE) and Autoencoding Variational Autoencoder (AVAE) methods \cite{cemgil2020autoencoding, Cemgil2019-vn}.
We notice that these methods provide higher adversarial accuracy. However, we have also observed a large discrepancy in terms of the MSE and FID scores of the model itself compared to our experiments, which we suspect might be a result of a mistake in \cite{cemgil2020autoencoding, Cemgil2019-vn}.
\footnote{We have gotten in touch with the authors, who are kindly working with us to address this issue}


Lastly, we would like to highlight that our defence strategy can be also combined with all the above VAE modifications. One advantage of our approach is that it does not require changing the training procedure of a VAE and, as a result, it does not decrease the quality of the generated images. \upd{Moreover, we can apply the same procedure to reconstruct the non-corrupted points and it will improve the reconstruction error. This result can be seen in the Tables \ref{tab:mnist_attack_result} and \ref{tab:color_mnist_attack_results} and it goes in line with the results of the \cite{salimans2015markov}, where MCMC was used to improve the VAE performance. }


\begin{figure*}[t]
    \begin{adjustbox}{center}
    \begin{tabular}{llll}
        \includegraphics[width=0.25\textwidth]{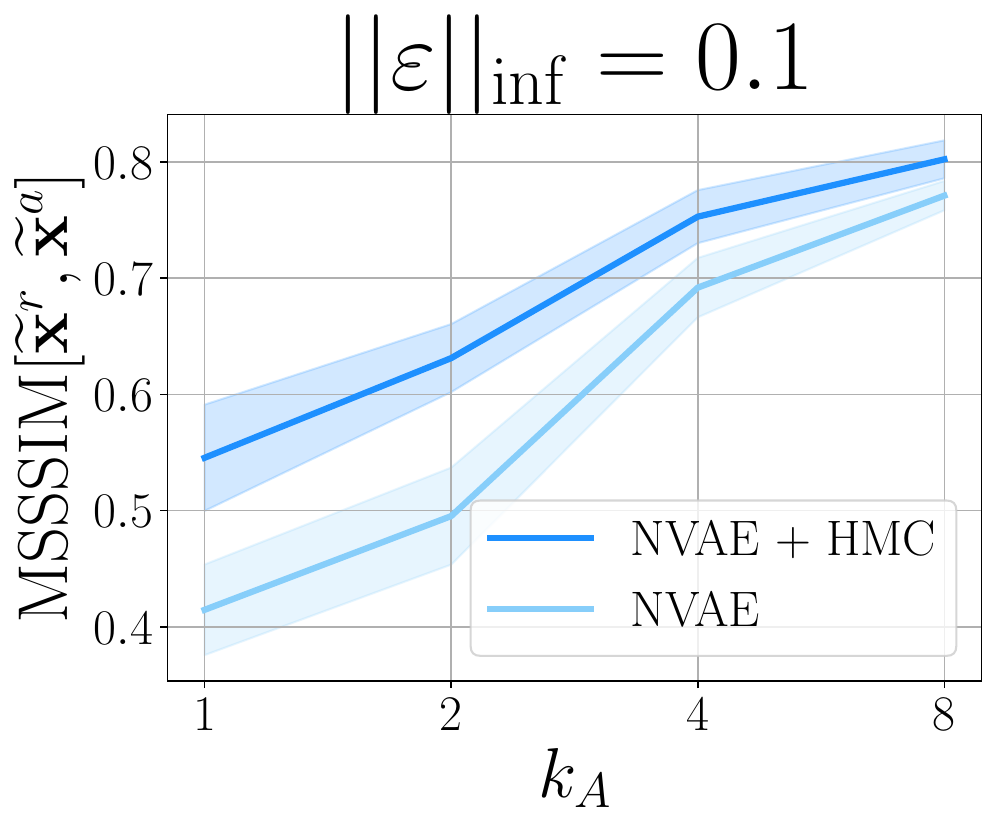} & 
        \includegraphics[width=0.25\textwidth]{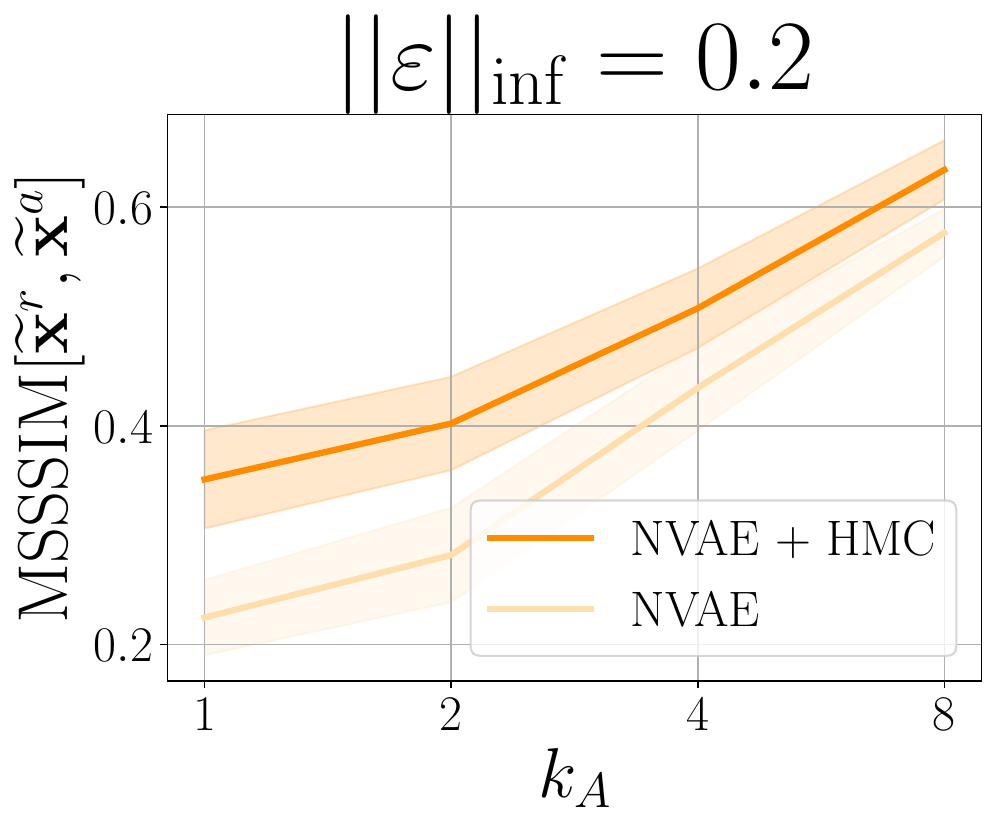} &
        \includegraphics[width=0.25\textwidth]{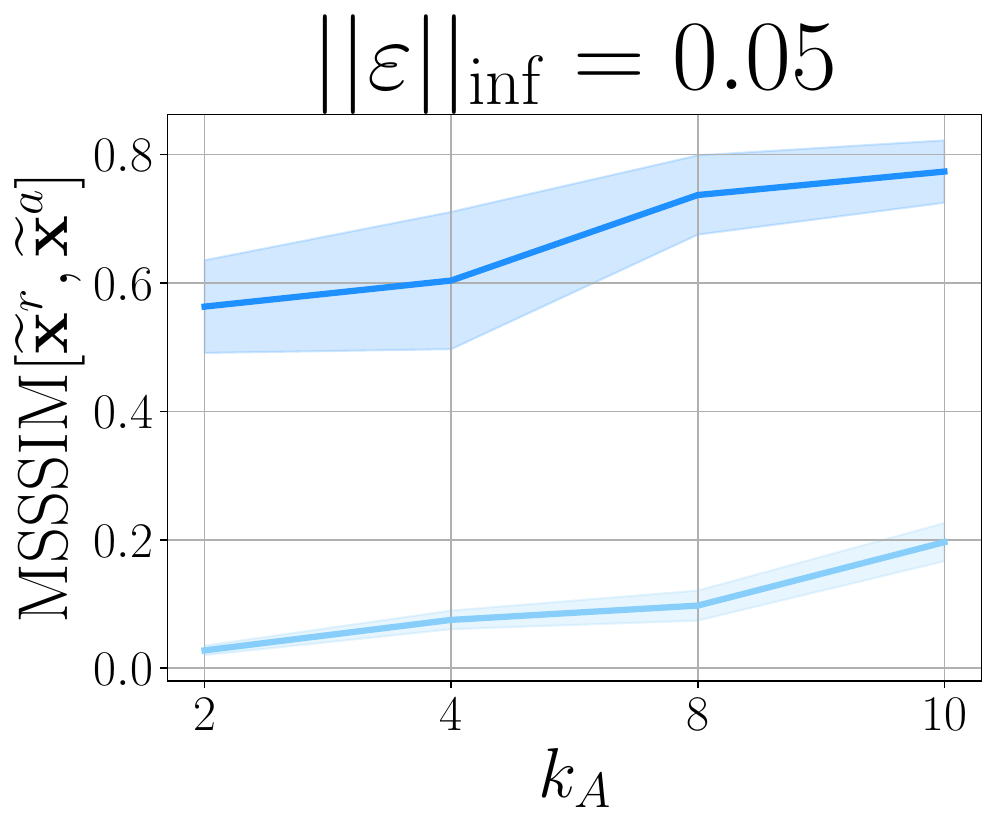} & 
        \includegraphics[width=0.25\textwidth]{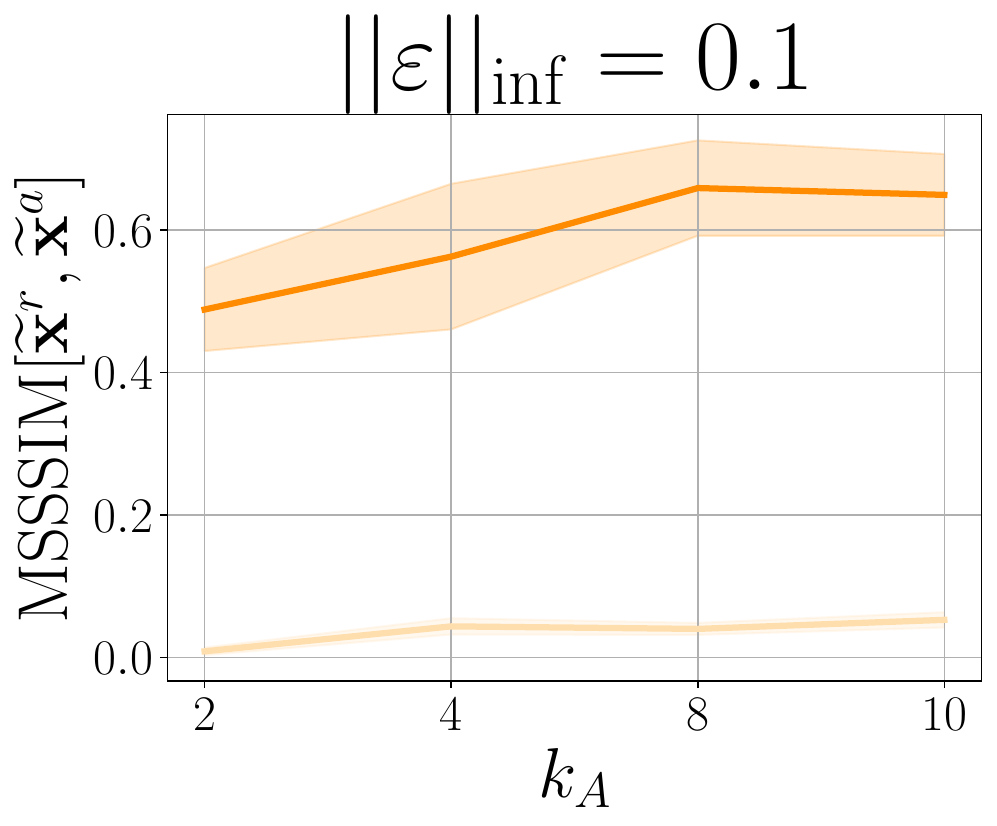}\\
        \multicolumn{2}{c}{(a) The reconstruction similarity for MNIST} &
         \multicolumn{2}{c}{(b) The reconstruction similarity for CelebA}\\
    \end{tabular}
    \end{adjustbox}
    \caption{The robustness improvement for the hierarchical model (NVAE) on (a) MNIST and (b) CelebA for two different attack radii. Higher values correspond to more robust representations.}
    \label{fig:nvae_res}
\end{figure*}

\subsection{Hierarchical VAE: NVAE}\label{sec:nvae}

\paragraph{Model and datasets} In this section, we explore the robustness of the deep hierarchical VAE (NVAE) \cite{Vahdat2020-xe}, a specific implementation of a hierarchical VAE that works well for high-dimensional data. We attack models trained on MNIST and CelebA \cite{liu2015faceattributes} datasets. We use the weights of the pre-trained model provided in the official NVAE implementation\footnote{The code and model weights were taken from \texttt{https://github.com/NVlabs/NVAE}}. 

\paragraph{Attacks construction}
Following \cite{kuzina2021adv} we construct adversarial attacks on the hierarchical VAE by considering higher-level latent variables. That being said, we use latent variables $\{\rvz_{L-k_A}, \rvz_{L-k_A+1}, \dots, \rvz_L\}$ when constructing an attack (\eqref{eq:objective_unsup}). Otherwise, we follow the same procedure we used for VAEs with a single level of latent variables. We assume that the attacker has access to the model's encoder and uses the symmetric KL-divergence as the objective. The radius of an attack is measured with the $L_{\inf}$ norm. For optimization, we use the projected gradient descent with the number of iterations limited to 50 per point. Further details are reported in the Appendix \ref{appendix:vae_attack}.

\paragraph{Results}
In Figure \ref{fig:nvae_res}, we present reconstruction similarity of the reference and adversarial points for both datasets. We observe that the proposed method consistently improves the robustness of the model to the adversarial attack. This result is in line with our theoretical considerations where a flexible class of variational posteriors could help to counteract adversarial attacks and, eventually, deacrease the bias of the class of models measured in terms of the KL-divergence. Additionally, applying the MCMC can further help us to counteract the attack. In Figure \ref{fig:adversarial_examples} (the two bottom rows), we show how an adversarial point (c) is reconstructed without any defence (d) and with the proposed defence (e). In the depicted samples, we have used the top 10 latent variables ($k_A = 10$) to construct the attack with a radius of 0.05.

\begin{figure}[t]
\vskip - 15pt
    \centering
    \begin{tabular}{cc}
     \includegraphics[width=0.38\linewidth]{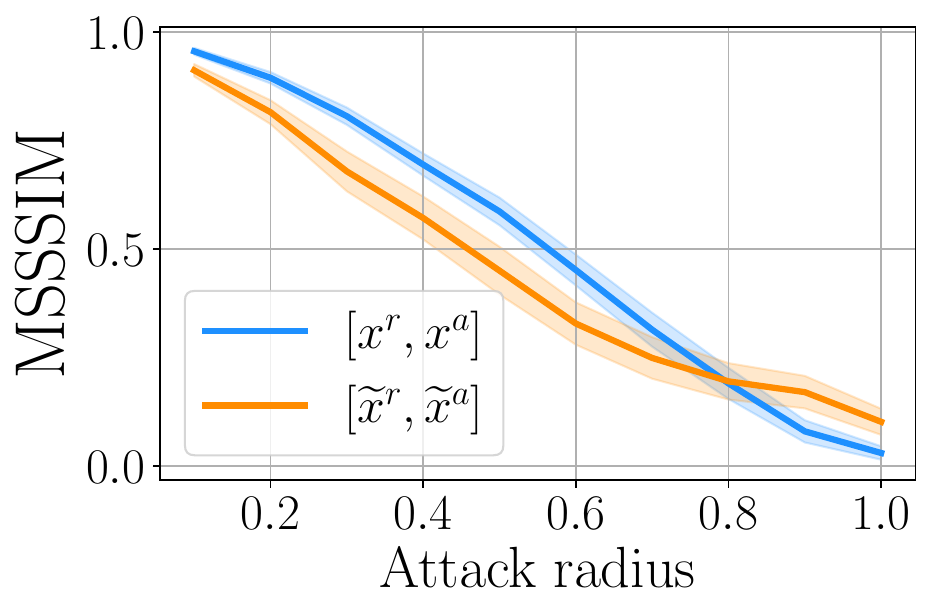} &
     \includegraphics[width=0.38\linewidth]{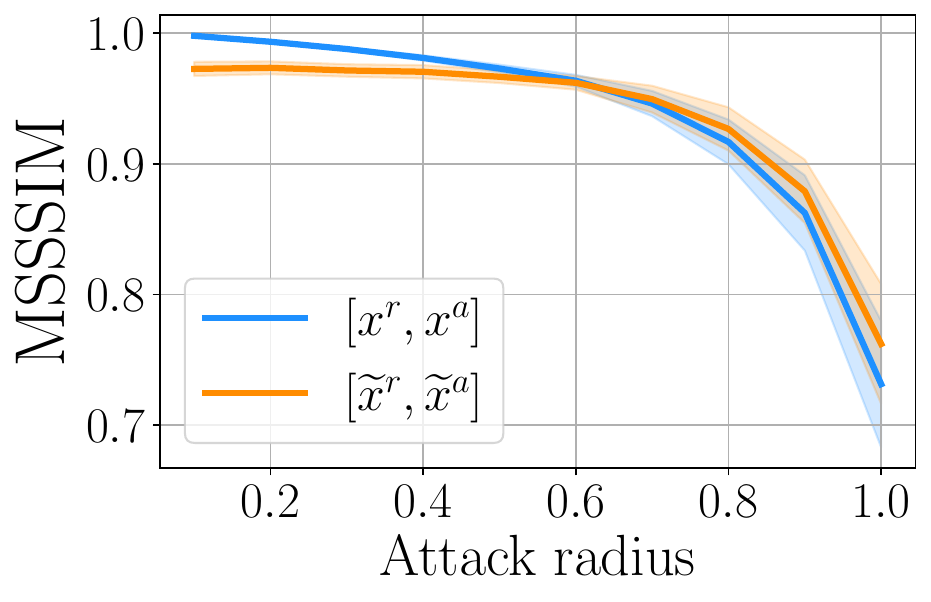} \\
     \multirow{2}{0.5\linewidth}{(a) An attacker does not know the defence strategy}&
     \multirow{2}{0.5\linewidth}{(b) An attacker knows the defence strategy }\\
     \\
    \end{tabular}
    \caption{Robustness to adversarial attack (with HMC defence). We report similarity of the reference and adversarial point (blue) and their reconstructions (orange).}
    \label{fig:attack_mcmc}
    \vskip - 12pt
\end{figure}


\subsection{Ablation Study: What if the attacker knows the defence strategy?} \label{sec:ablation}
In our experiments, we rely on the assumption that the attacker does not take into account the defence strategy that we use. We believe that it is reasonable, since the defence requires access to the decoder part of the model, $p_{\theta}(x|z)$, which is not necessarily available to the attacker. 

In this ablation study we verify how the robustness results change if we construct the attack with the access to the defence strategy. We train an adversarial attack with the modified objective \ref{eq:objective_unsup}, which takes into account the HMC step. See Appendix \ref{appendix:attack_mcmc} for more details on the experimental setup. 

In Figure \ref{fig:attack_mcmc} we show the experiment results for various attack radii between 0 and 1. We observe that constructing an attack with such an objective is much harder (Figure \ref{fig:attack_mcmc} (b)).

\section{Discussion}
Following the previous works on attacking VAEs \cite{barrett2021certifiably, camuto2021towards, cemgil2020autoencoding, Cemgil2019-vn, Gondim-Ribeiro2018-cu, Willetts2019-mu}, we only consider the projected gradient descent as a way to construct the attack. However, more sophisticated adaptive methods \cite{ athalye2018obfuscated, tramer2020adaptive} were proposed to attack discriminative model and can be potentially applied to VAEs as well.
We believe that it is an interesting direction for the future work. 

\paragraph{Objective Function}
For the unsupervised attack on the encoder, we use the symmetric KL-divergence to measure the dissimilarity. However, other options are possible, e.g., the forward or reverse KL-divergence or even $L_2$ distance between the means (see Table \ref{tab:attack_variation}). In our comparative experiments (see Appendix \ref{appendix:objectives}), we observe that no single objective consistently performs better than others.


\paragraph{Attack radius}
During the attack construction we seek to obtain a point that will have the most different latent representation or a different predicted class. However, it is also important that the point itself is as similar to the initial reference point as possible. In Appendix \ref{appendix:attack_radius}, we visualize how the attacks of different radii influence the similarity between the adversarial and reference points. Based on these results, we have chosen the radii which do not allow adversarial points to deviate a lot from the reference (as measured by \textsc{MSSSIM}): $\|\varepsilon\|_{\inf} \leq 0.2$ for the MNIST dataset (which goes in line with the previous works \cite{cemgil2020autoencoding, Cemgil2019-vn}) and $\|\varepsilon\|_{\inf} \leq 0.1$ for the CelebA dataset.

\begin{wrapfigure}{r}{0.54\textwidth}
\vskip -10pt
    \begin{tabular}{c}
        \includegraphics[width=0.8\linewidth]{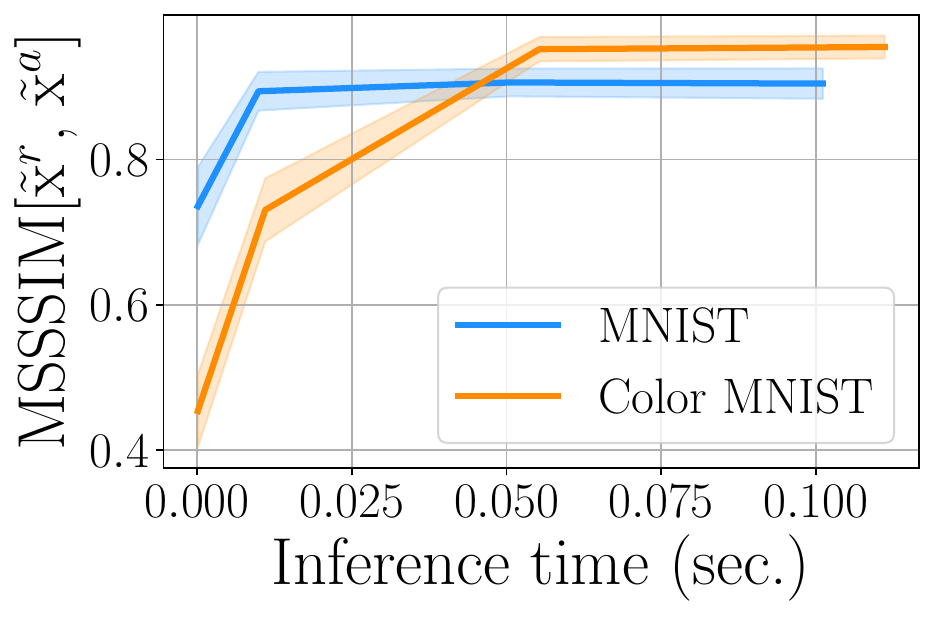} \\
    \end{tabular}
    \vskip -5pt
    \caption{Trade-off between robustness and inference time.}
    \vskip -28pt
    \label{fig:time_vs_performance}
\end{wrapfigure}

\paragraph{\discussion{Number of MCMC steps and Inference Time}}
In our approach, we have to select the number of MCMC steps that the defender performs. This parameter potentially can be critical as it influences both the inference time (see Appendix \ref{appendix:inference_time}) and the performance (see Appendix \ref{appendix:hmc_steps}). \discussion{In Figure \ref{fig:time_vs_performance} we show the trade-off between the reconstruction similarity and the inference time. The increase in the inference time is cause by a larger number of HMC steps used (we consider 0, 100, 500 and 1000 steps for this experiment).}


\section{Conclusion}
In this work, we explore the robustness of VAEs to adversarial attacks.  We propose a theoretically justified method that allows alleviating the effect of attacks on the latent representations by improving the reconstructions of the adversarial inputs and the downstream tasks accuracy. We experimentally validate our approach on a variety of datasets: both grey-scale (MNIST, Fashion MNIST) and colored (ColorMNIST, CelebA) data. We show that the proposed method improves the robustness of the vanilla VAE models and its various modifications, i.e., $\beta$-VAE, $\beta$-TCVAE and NVAE.

\section*{Acknowledgements}

Anna Kuzina is funded by the Hybrid Intelligence Center, a 10-year programme funded by the Dutch Ministry of Education, Culture and Science through the Netherlands Organisation for Scientific Research. Experiments were carried out on the Dutch national e-infrastructure with the support of SURF Cooperative

\bibliographystyle{abbrv}
\bibliography{bibliography.bib}

\begin{thebibliography}{10}

\bibitem{an2015variational}
J.~An and S.~Cho.
\newblock Variational autoencoder based anomaly detection using reconstruction
  probability.
\newblock {\em Special Lecture on IE}, 2(1):1--18, 2015.

\bibitem{andrieu2003introduction}
C.~Andrieu, N.~De~Freitas, A.~Doucet, and M.~I. Jordan.
\newblock An introduction to mcmc for machine learning.
\newblock {\em Machine learning}, 50(1):5--43, 2003.

\bibitem{athalye2018obfuscated}
A.~Athalye, N.~Carlini, and D.~Wagner.
\newblock Obfuscated gradients give a false sense of security: Circumventing
  defenses to adversarial examples.
\newblock In {\em International conference on machine learning}, pages
  274--283. PMLR, 2018.

\bibitem{auzina2021approximate}
I.~A. Auzina and J.~M. Tomczak.
\newblock Approximate bayesian computation for discrete spaces.
\newblock {\em Entropy}, 23(3):312, 2021.

\bibitem{balle2018variational}
J.~Ball{\'e}, D.~Minnen, S.~Singh, S.~J. Hwang, and N.~Johnston.
\newblock Variational image compression with a scale hyperprior.
\newblock In {\em International Conference on Learning Representations}, 2018.

\bibitem{barrett2021certifiably}
B.~Barrett, A.~Camuto, M.~Willetts, and T.~Rainforth.
\newblock Certifiably robust variational autoencoders.
\newblock {\em International Conference on Artificial Intelligence and
  Statistics}, pages 3663--3683, 2022.

\bibitem{bengio2013representation}
Y.~Bengio, A.~Courville, and P.~Vincent.
\newblock Representation learning: A review and new perspectives.
\newblock {\em IEEE transactions on pattern analysis and machine intelligence},
  35(8):1798--1828, 2013.

\bibitem{betancourt2017conceptual}
M.~Betancourt.
\newblock A conceptual introduction to hamiltonian monte carlo.
\newblock {\em arXiv preprint arXiv:1701.02434}, 2017.

\bibitem{burda2015importance}
Y.~Burda, R.~B. Grosse, and R.~Salakhutdinov.
\newblock Importance weighted autoencoders.
\newblock In {\em International Conference on Learning Representations}, 2016.

\bibitem{camuto2021towards}
A.~Camuto, M.~Willetts, S.~Roberts, C.~Holmes, and T.~Rainforth.
\newblock Towards a theoretical understanding of the robustness of variational
  autoencoders.
\newblock In {\em International Conference on Artificial Intelligence and
  Statistics}, pages 3565--3573. PMLR, 2021.

\bibitem{caterini2018hamiltonian}
A.~L. Caterini, A.~Doucet, and D.~Sejdinovic.
\newblock Hamiltonian variational auto-encoder.
\newblock {\em Advances in Neural Information Processing Systems}, 31, 2018.

\bibitem{cemgil2020autoencoding}
T.~Cemgil, S.~Ghaisas, K.~Dvijotham, S.~Gowal, and P.~Kohli.
\newblock The autoencoding variational autoencoder.
\newblock {\em Advances in Neural Information Processing Systems}, 33, 2020.

\bibitem{Cemgil2019-vn}
T.~Cemgil, S.~Ghaisas, K.~D. Dvijotham, and P.~Kohli.
\newblock Adversarially robust representations with smooth encoders.
\newblock In {\em International Conference on Learning Representations}.
  openreview.net, 2019.

\bibitem{chen2018isolating}
R.~T. Chen, X.~Li, R.~B. Grosse, and D.~K. Duvenaud.
\newblock Isolating sources of disentanglement in variational autoencoders.
\newblock {\em Advances in neural information processing systems}, 31, 2018.

\bibitem{Child2020-ze}
R.~Child.
\newblock Very deep {VAEs} generalize autoregressive models and can outperform
  them on images.
\newblock In {\em International Conference on Learning Representations}, 2021.

\bibitem{cremer2018inference}
C.~Cremer, X.~Li, and D.~Duvenaud.
\newblock Inference suboptimality in variational autoencoders.
\newblock In {\em International Conference on Machine Learning}, pages
  1078--1086. PMLR, 2018.

\bibitem{duane1987hybrid}
S.~Duane, A.~D. Kennedy, B.~J. Pendleton, and D.~Roweth.
\newblock Hybrid monte carlo.
\newblock {\em Physics letters B}, 195(2):216--222, 1987.

\bibitem{Gondim-Ribeiro2018-cu}
G.~Gondim-Ribeiro, P.~Tabacof, and E.~Valle.
\newblock Adversarial attacks on variational autoencoders.
\newblock {\em ArXiv Preprint}, June 2018.

\bibitem{goodfellow2014explaining}
I.~J. Goodfellow, J.~Shlens, and C.~Szegedy.
\newblock Explaining and harnessing adversarial examples.
\newblock {\em arXiv preprint arXiv:1412.6572}, 2014.

\bibitem{habibian2019video}
A.~Habibian, T.~v. Rozendaal, J.~M. Tomczak, and T.~S. Cohen.
\newblock Video compression with rate-distortion autoencoders.
\newblock In {\em Proceedings of the IEEE/CVF International Conference on
  Computer Vision}, pages 7033--7042, 2019.

\bibitem{higgins2016beta}
I.~Higgins, L.~Matthey, A.~Pal, C.~Burgess, X.~Glorot, M.~Botvinick,
  S.~Mohamed, and A.~Lerchner.
\newblock beta-vae: Learning basic visual concepts with a constrained
  variational framework.
\newblock {\em International Conference on Learning Representations (ICLR)},
  2017.

\bibitem{higgins2017darla}
I.~Higgins, A.~Pal, A.~Rusu, L.~Matthey, C.~Burgess, A.~Pritzel, M.~Botvinick,
  C.~Blundell, and A.~Lerchner.
\newblock Darla: Improving zero-shot transfer in reinforcement learning.
\newblock In {\em International Conference on Machine Learning}, pages
  1480--1490. PMLR, 2017.

\bibitem{hoffman2017learning}
M.~D. Hoffman.
\newblock Learning deep latent gaussian models with markov chain monte carlo.
\newblock In {\em International conference on machine learning}, pages
  1510--1519. PMLR, 2017.

\bibitem{izmailov2021bayesian}
P.~Izmailov, S.~Vikram, M.~D. Hoffman, and A.~G.~G. Wilson.
\newblock What are bayesian neural network posteriors really like?
\newblock In {\em International conference on machine learning}, pages
  4629--4640. PMLR, 2021.

\bibitem{jordan1999introduction}
M.~I. Jordan, Z.~Ghahramani, T.~S. Jaakkola, and L.~K. Saul.
\newblock An introduction to variational methods for graphical models.
\newblock {\em Machine learning}, 37(2):183--233, 1999.

\bibitem{kim2018disentangling}
H.~Kim and A.~Mnih.
\newblock Disentangling by factorising.
\newblock In {\em International Conference on Machine Learning}, pages
  2649--2658. PMLR, 2018.

\bibitem{kingma2013auto}
D.~P. Kingma and M.~Welling.
\newblock Auto-encoding variational bayes.
\newblock In {\em International Conference on Learning Representations, {ICLR}
  2014}, 2014.

\bibitem{kuzina2021adv}
A.~Kuzina, M.~Welling, and J.~M. Tomczak.
\newblock Diagnosing vulnerability of variational auto-encoders to adversarial
  attacks.
\newblock {\em ICLR, RobustML workshop}, 2021.

\bibitem{liu2015faceattributes}
Z.~Liu, P.~Luo, X.~Wang, and X.~Tang.
\newblock Deep learning face attributes in the wild.
\newblock In {\em Proceedings of International Conference on Computer Vision
  (ICCV)}, December 2015.

\bibitem{Maaloe2019-bp}
L.~Maal{\o}e, M.~Fraccaro, V.~Li{\'e}vin, and O.~Winther.
\newblock {BIVA}: A very deep hierarchy of latent variables for generative
  modeling.
\newblock In {\em Advances in Neural Information Processing Systems}, Feb.
  2019.

\bibitem{neal2011mcmc}
R.~M. Neal et~al.
\newblock Mcmc using hamiltonian dynamics.
\newblock {\em Handbook of markov chain monte carlo}, 2(11):2, 2011.

\bibitem{nishimura2020discontinuous}
A.~Nishimura, D.~B. Dunson, and J.~Lu.
\newblock Discontinuous hamiltonian monte carlo for discrete parameters and
  discontinuous likelihoods.
\newblock {\em Biometrika}, 107(2):365--380, 2020.

\bibitem{Ranganath2016-yg}
R.~Ranganath, D.~Tran, and D.~Blei.
\newblock Hierarchical variational models.
\newblock In {\em Proceedings of The 33rd International Conference on Machine
  Learning}, volume~48 of {\em Proceedings of Machine Learning Research}, pages
  324--333, New York, New York, USA, 2016. PMLR.

\bibitem{rezende2014stochastic}
D.~J. Rezende, S.~Mohamed, and D.~Wierstra.
\newblock Stochastic backpropagation and approximate inference in deep
  generative models.
\newblock In {\em International Conference on Machine Learning}, pages
  1278--1286. PMLR, 2014.

\bibitem{ruiz2019contrastive}
F.~Ruiz and M.~Titsias.
\newblock A contrastive divergence for combining variational inference and
  mcmc.
\newblock In {\em International Conference on Machine Learning}, pages
  5537--5545. PMLR, 2019.

\bibitem{salimans2015markov}
T.~Salimans, D.~Kingma, and M.~Welling.
\newblock Markov chain monte carlo and variational inference: Bridging the gap.
\newblock In {\em International conference on machine learning}, pages
  1218--1226. PMLR, 2015.

\bibitem{So_nderby2016-en}
C.~K. S{\o}nderby, T.~Raiko, L.~Maal{\o}~e, S.~R.~K. S{\o}~nderby, and
  O.~Winther.
\newblock Ladder variational autoencoders.
\newblock In {\em Advances in Neural Information Processing Systems},
  volume~29, pages 3738--3746. Curran Associates, Inc., 2016.

\bibitem{tramer2020adaptive}
F.~Tramer, N.~Carlini, W.~Brendel, and A.~Madry.
\newblock On adaptive attacks to adversarial example defenses.
\newblock {\em Advances in Neural Information Processing Systems},
  33:1633--1645, 2020.

\bibitem{Vahdat2020-xe}
A.~Vahdat and J.~Kautz.
\newblock {NVAE}: A deep hierarchical variational autoencoder.
\newblock In {\em Neural Information Processing Systems (NeurIPS)}, 2020.

\bibitem{van2017neural}
A.~Van Den~Oord, O.~Vinyals, et~al.
\newblock Neural discrete representation learning.
\newblock {\em Advances in neural information processing systems}, 30, 2017.

\bibitem{wang2003multiscale}
Z.~Wang, E.~P. Simoncelli, and A.~C. Bovik.
\newblock Multiscale structural similarity for image quality assessment.
\newblock In {\em The Thrity-Seventh Asilomar Conference on Signals, Systems \&
  Computers, 2003}, volume~2, pages 1398--1402. Ieee, 2003.

\bibitem{Willetts2019-mu}
M.~Willetts, A.~Camuto, T.~Rainforth, S.~Roberts, and C.~Holmes.
\newblock Improving {VAEs'} robustness to adversarial attack.
\newblock In {\em International Conference on Learning Representations}, 2021.

\bibitem{wolf2016variational}
C.~Wolf, M.~Karl, and P.~van~der Smagt.
\newblock Variational inference with hamiltonian monte carlo.
\newblock {\em arXiv preprint arXiv:1609.08203}, 2016.

\bibitem{xiao2017fashion}
H.~Xiao, K.~Rasul, and R.~Vollgraf.
\newblock Fashion-mnist: a novel image dataset for benchmarking machine
  learning algorithms.
\newblock {\em arXiv preprint arXiv:1708.07747}, 2017.

\bibitem{zhang2022langevin}
R.~Zhang, X.~Liu, and Q.~Liu.
\newblock A langevin-like sampler for discrete distributions.
\newblock In {\em International Conference on Machine Learning}, pages
  26375--26396. PMLR, 2022.

\end{thebibliography}

\newpage
\section*{Checklist}


\begin{enumerate}

\item For all authors...
\begin{enumerate}
  \item Do the main claims made in the abstract and introduction accurately reflect the paper's contributions and scope?
    \answerYes{}
  \item Did you describe the limitations of your work?
    \answerYes{}
  \item Did you discuss any potential negative societal impacts of your work?
    \answerNA{}
  \item Have you read the ethics review guidelines and ensured that your paper conforms to them?
    \answerYes{}
\end{enumerate}

\item If you are including theoretical results...
\begin{enumerate}
  \item Did you state the full set of assumptions of all theoretical results?
    \answerYes{See Section~\ref{sec:defence}}
        \item Did you include complete proofs of all theoretical results?
    \answerYes{See Section~\ref{sec:defence} and Appendix~\ref{appendix:theory}}
\end{enumerate}

\item If you ran experiments...
\begin{enumerate}
  \item Did you include the code, data, and instructions needed to reproduce the main experimental results (either in the supplemental material or as a URL)?
    \answerYes{}
  \item Did you specify all the training details (e.g., data splits, hyperparameters, how they were chosen)?
    \answerYes{See Appendix~\ref{appendix:experimental_details}}
        \item Did you report error bars (e.g., with respect to the random seed after running experiments multiple times)?
    \answerYes{}
        \item Did you include the total amount of compute and the type of resources used (e.g., type of GPUs, internal cluster, or cloud provider)?
    \answerNo{}{}
\end{enumerate}

\item If you are using existing assets (e.g., code, data, models) or curating/releasing new assets...
\begin{enumerate}
  \item If your work uses existing assets, did you cite the creators?
    \answerYes{We use NVAE code and pretrained that are published on github}
  \item Did you mention the license of the assets?
    \answerYes{The licence allows using code for non-commercial purposes}
  \item Did you include any new assets either in the supplemental material or as a URL?
    \answerNA{}
  \item Did you discuss whether and how consent was obtained from people whose data you're using/curating?
    \answerNA{}
  \item Did you discuss whether the data you are using/curating contains personally identifiable information or offensive content?
    \answerNA{}
\end{enumerate}

\item If you used crowdsourcing or conducted research with human subjects...
\begin{enumerate}
  \item Did you include the full text of instructions given to participants and screenshots, if applicable?
    \answerNA{}
  \item Did you describe any potential participant risks, with links to Institutional Review Board (IRB) approvals, if applicable?
    \answerNA{}
  \item Did you include the estimated hourly wage paid to participants and the total amount spent on participant compensation?
    \answerNA{}
\end{enumerate}

\end{enumerate}


\newpage
\appendix

\onecolumn
\section{Theory}\label{appendix:theory}
We consider an attack, which has an additive structure:
\begin{align}
    \rvx^a = \rvx^r + \varepsilon, \\
    \text{such that }\|\varepsilon\| \leq \delta,
\end{align}
where $\delta$ is a radius of the attack. 

In the vanilla VAE setup we will get the latent code by sampling from $q_{\phi}(\rvz|\rvx)$. With our approach, instead, we get a sample from the following distribution:
\begin{align}
    q^{(t)}(\rvz|\rvx) = \int Q^{(t)}(\rvz|\rvz_0) q_{\phi}(\rvz_0|\rvx) d\rvz_0,
\end{align}

where $Q^{(t)}(\rvz|\rvz_0)$ is a transition kernel of MCMC with the target distribution $\pi(\rvz) = p_{\theta}(\rvz|\rvx) \propto p_{\theta}(\rvx|\rvz)p(\rvz)$.

\paragraph{Lemma 1}
Consider true posterior distributions of the latent code $\rvz$ for a data point $\rvx$ and its corrupted version $\rvx^a$.  Assume also that $\ln p_{\theta}(\rvz|\rvx)$ \ac{is twice differentiable over $\rvx$ with continuous derivatives at the neighbourhood around $\rvx=\rvx^r$.} 
Then the KL-divergence between these two posteriors could be expressed using the small $o$ notation of the radius of the attack, namely:
\begin{equation}
     \KL {p_{\theta}(\rvz|\rvx^r)} {p_{\theta}(\rvz|\rvx^a)} = o(\|\varepsilon\|).
\end{equation}
\textit{Proof}\\
Let us use definition of the KL-divergence:
\begin{align} \label{eq:kl_def}
     \KL{p_{\theta}(\rvz|\rvx^r)}{p_{\theta}(\rvz|\rvx^a)} &= \E_{p_{\theta}(\rvz|\rvx^r)} \ln \frac{p_{\theta}(\rvz|\rvx^r)}{p_{\theta}(\rvz|\rvx^a)}  
\end{align}

Let us introduce $\ln p_{\theta}(\rvz|\rvx) = g(\rvx, \rvz)$. Assume that this function is differentiable at $\rvx = \rvx^r$. Then, we can apply Taylor expansion to $g(\rvx, \rvz)$ in the point $\rvx^r$ which yields:
\begin{equation}
    g(\rvx, \rvz) =  g(\rvx^r, \rvz) + (\rvx - \rvx^r)^T \nabla_{\rvx} g(\rvx, \rvz) \Big|_{\rvx^r} + R_1(\rvx, \rvx^r) .
\end{equation}

The remainder term in the Lagrange form can be written as
\begin{equation}\label{eq:remainder}
    R_1(\rvx, \rvx^r) = \tfrac{1}{2}(\rvx - \rvx^r)^T \nabla^2_{\rvx\rvx} g(\rvx + \theta (\rvx - \rvx^r), \rvz) (\rvx - \rvx^r), \, \theta \in (0, 1)
\end{equation}
\ac{Under the assumption that $g$ is twice differentiable with the continuous derivatives on the segment around $\rvx = \rvx^r$ the remainder term asymptotically converges to zero with $\rvx \rightarrow \rvx^r$. 
\begin{equation}
    R_1(\rvx, \rvx^r) = o(\|\rvx - \rvx^r\|).
\end{equation}}

Then, the log-ratio of two distributions is the following:
\begin{align}
    \ln \frac{p_{\theta}(\rvz| \rvx^r)}{p_{\theta}(\rvz|\rvx^a)}
    &= g(\rvx^r, \rvz) - g(\rvx^a, \rvz) \\
    &= g(\rvx^r, \rvz) - \left( g(\rvx^r, \rvz) + (\rvx^a - \rvx^r)^T \nabla_{\rvx} g(\rvx, \rvz) \Big|_{\rvx^r} + o(\|\rvx^a - \rvx^r\|)\right). \\
    &=  -\varepsilon^T\nabla_{\rvx} g(\rvx, \rvz) \Big|_{\rvx^r} + o(\|\varepsilon\|) .
\end{align}
Notice that $\varepsilon^T \nabla_{\rvx} g(\rvx, \rvz) \Big|_{\rvx^r} $ is the dot product between $\varepsilon$ and $\nabla_{\rvx} g(\rvx, \rvz) \Big|_{\rvx^r} $, i.e., $\varepsilon^T \nabla_{\rvx} g(\rvx, \rvz) \Big|_{\rvx^r}  = \langle \varepsilon, \nabla_{\rvx} g(\rvx, \rvz) \Big|_{\rvx^r}  \rangle$.

We can now plug this into the KL-divergence definition (\ref{eq:kl_def}):
\begin{align}
     \KL{p_{\theta}(\rvz|\rvx^r)}{p_{\theta}(\rvz|\rvx^a)} &= \E_{p_{\theta}(\rvz|\rvx^r)} \left[ -\langle \varepsilon, \nabla_{\rvx} \ln p_{\theta}(\rvz|\rvx) \Big|_{\rvx^r}  \rangle + o(\|\varepsilon\|)\right] \\
     &=   -\langle \varepsilon, \underbrace{ \E_{p_{\theta}(\rvz|\rvx^r)}  \nabla_{\rvx} \ln p_{\theta}(\rvz|\rvx) \Big|_{\rvx^r}}_{\text{A}(\rvx^r)}  \rangle + o(\|\varepsilon\|) \label{eq:kl_in_lemma1}
\end{align}

Note that for (\ref{eq:kl_in_lemma1}) to hold we need to make sure that $\E_z R_1 = o(\|\varepsilon\|)$. As follows from (\ref{eq:remainder}), this requirement is satisfied if $\E_{z}\nabla^2_{\rvx\rvx} g(\rvx + \theta (\rvx - \rvx^r), \rvz)$ is bounded around $\rvx = \rvx^r$.

Let us take a closer to look at the term $\text{A}(\rvx^r)$ in the equation above:

\begin{align}
\text{A}(\rvx^r) &=    \E_{p_{\theta}(\rvz|\rvx^r)}  \nabla_{\rvx} \ln p_{\theta}(\rvz|\rvx) \Big|_{\rvx^r} \\
&=  \E_{p_{\theta}(\rvz|\rvx^r)}  \nabla_{\rvx} \ln \frac{p_{\theta}(\rvx|\rvz)p(\rvz)}{p_{\theta}(\rvx)} \Big|_{\rvx^r}   \label{eq:bayes_rule}  \\
&= \E_{p_{\theta}(\rvz|\rvx^r)}  \nabla_{\rvx} \ln p_{\theta}(\rvx|\rvz) \Big|_{\rvx^r} - \E_{p_{\theta}(\rvz|\rvx^r)}  \nabla_{\rvx} \ln p_{\theta}(\rvx) \Big|_{\rvx^r} \\
&= \int p_{\theta}(\rvz|\rvx^r)\frac{\nabla_{\rvx} p_{\theta}(\rvx|\rvz)\Big|_{\rvx^r}}{p_{\theta}(\rvx^r|\rvz)}d\rvz - \int p_{\theta}(\rvz|\rvx^r)\frac{\nabla_{\rvx} p_{\theta}(\rvx)\Big|_{\rvx^r}}{p_{\theta}(\rvx^r)} d\rvz  
\label{eq:log_derivative}\\
&= \int \frac{p_{\theta}(\rvz)}{p_{\theta}(\rvx^r)} \nabla_{\rvx} p_{\theta}(\rvx|\rvz)\Big|_{\rvx^r} d\rvz - \frac{\nabla_{\rvx} p_{\theta}(\rvx)\Big|_{\rvx^r}}{p_{\theta}(\rvx^r)}  \underbrace{\int p_{\theta}(\rvz|\rvx^r)d\rvz }_{=1}  \label{eq:bayes_rule_2} \\
&= \frac{1}{p_{\theta}(\rvx^r)} \left[ \int p(\rvz) \nabla_{\rvx} p_{\theta}(\rvx|\rvz)\Big|_{\rvx^r} d\rvz - \nabla_{\rvx} p_{\theta}(\rvx)\Big|_{\rvx^r} \right]\\
&= \frac{1}{p_{\theta}(\rvx^r)} \left[ \E_{p(\rvz)} \nabla_{\rvx} p_{\theta}(\rvx|\rvz)\Big|_{\rvx^r} - \nabla_{\rvx} \E_{p(\rvz)}p_{\theta}(\rvx|\rvz)\Big|_{\rvx^r} \right]\\
&= \frac{1}{p_{\theta}(\rvx^r)}  \underbrace{\left[ \E_{p(\rvz)} \nabla_{\rvx} p_{\theta}(\rvx|\rvz)\Big|_{\rvx^r} -  \E_{p(\rvz)}\nabla_{\rvx} p_{\theta}(\rvx|\rvz)\Big|_{\rvx^r} \right]}_{=0} = 0.
\end{align}

where we use Bayes rule in \ref{eq:bayes_rule}, \ref{eq:bayes_rule_2}, log-derivative trick in \ref{eq:log_derivative}. 

We have shown that $\text{A}(\rvx^r) = 0$, therefore, from \eqref{eq:kl_in_lemma1} we have:
 
\begin{align}
    \KL{p_{\theta}(\rvz|\rvx^r)}{p_{\theta}(\rvz|\rvx^a)} &= -\langle \varepsilon, \text{A}(\rvx^r) \rangle + o(\|\varepsilon\|) = o(\|\varepsilon\|). 
\end{align}
$\hfill\blacksquare$

\paragraph{Lemma 2}
The Total Variation distance (TV) between the variational posterior with MCMC for a given corrupted point $\rvx^a$, $q^{(t)}(\rvz|\rvx^a)$, and the variational posterior for a given data point $\rvx^r$, $q_{\phi}(\rvz|\rvx^r)$, can be upper bounded by the sum of the following three components:
\begin{align}
\TV{q^{(t)}(\rvz|\rvx^a)}{q_{\phi}(\rvz|\rvx^r)} &\leq 
    \TV{q^{(t)}(\rvz|\rvx^a)}{p_{\theta}(\rvz|\rvx^a)} \\
     &+ 
    \sqrt{\tfrac12 \KL{p_{\theta}(\rvz|\rvx^r)}{p_{\theta}(\rvz|\rvx^a)}  } \\
    &+
   \sqrt{\tfrac12  \KL{q_{\phi}(\rvz|\rvx^r)}{p_{\theta}(\rvz|\rvx^r)}}.
\end{align}
\textit{Proof}\\
Total variation is a proper distance, thus, the triangular inequality holds for it. For the proof, we apply the triangular inequality twice. First, we use the triangle inequality for $\text{TV}\left[ q^{(t)}(\rvz|\rvx^a), q_{\phi}(\rvz|\rvx^r) \right]$, namely: 
\begin{align}
    \TV{q^{(t)}(\rvz|\rvx^a)}{q_{\phi}(\rvz|\rvx^r)} \leq 
    \TV{q^{(t)}(\rvz|\rvx^a)}{p_{\theta}(\rvz|\rvx^r)} + 
    \TV{p_{\theta}(\rvz|\rvx^r)}{q_{\phi}(\rvz|\rvx^r)}.
\end{align}
Second, we utilize the triangle inequality for $\TV{q^{(t)}(\rvz|\rvx^a)} {p_{\theta}(\rvz|\rvx^r) }$, that is:
\begin{align}
    \TV{q^{(t)}(\rvz|\rvx^a)}{p_{\theta}(\rvz|\rvx^r)}
    \leq 
    \TV{q^{(t)}(\rvz|\rvx^a)}{p_{\theta}(\rvz|\rvx^a)} + 
    \TV{p_{\theta}(\rvz|\rvx^a)}{p_{\theta}(\rvz|\rvx^r)} .
\end{align}

Combining the two gives us the following upper bound on the initial total variation:

\begin{align}
\TV{q^{(t)}(\rvz|\rvx^a)}{q_{\phi}(\rvz|\rvx^r)}
&\leq 
    \TV{q^{(t)}(\rvz|\rvx^a)}{p_{\theta}(\rvz|\rvx^a)} \\
     &+ 
    \TV{p_{\theta}(\rvz|\rvx^a)}{p_{\theta}(\rvz|\rvx^r)} \label{eq:three_comp_2}\\
    &+
    \TV{p_{\theta}(\rvz|\rvx^r)}{q_{\phi}(\rvz|\rvx^r)} \label{eq:three_comp_3}
\end{align}

Moreover, the Total Variation distance is a lower bound of the KL-divergence (by Pinsker inequality):
\begin{equation}
    \TV{p(\rvx)}{q(\rvx)}
    \leq \sqrt{\tfrac12 \KL{p(\rvx)}{q(\rvx)}} .
\end{equation}

Applying Pinsker inequality to 
\ref{eq:three_comp_2} and \ref{eq:three_comp_3}
yields:
\begin{align}
\TV{q^{(t)}(\rvz|\rvx^a)}{q_{\phi}(\rvz|\rvx^r)}
&\leq 
    \TV{q^{(t)}(\rvz|\rvx^a)}{p_{\theta}(\rvz|\rvx^a)} \\
     &+ 
    \sqrt{\tfrac12 \KL{p_{\theta}(\rvz|\rvx^r)}{p_{\theta}(\rvz|\rvx^a)}  } \\
    &+
   \sqrt{\tfrac12  \KL{q_{\phi}(\rvz|\rvx^r)}{p_{\theta}(\rvz|\rvx^r)}} .
\end{align}
$\hfill\blacksquare$

\paragraph{Theorem 1}
The upper bound on the total variation distance between samples from MCMC for a given corrupted point $\rvx^a$, $q^{(t)}(\rvz|\rvx^a)$, and the variational posterior for the given real point $\rvx$, $q_{\phi}(\rvz|\rvx)$, is the following:

\begin{align}
    \TV{q^{(t)}(\rvz|\rvx^a)}{q_{\phi}(\rvz|\rvx^r)}
\leq 
\TV{q^{(t)}(\rvz|\rvx^a)}{p_{\theta}(\rvz|\rvx^a)}
     + 
   \sqrt{\tfrac12  \KL{q_{\phi}(\rvz|\rvx^r)}{p_{\theta}(\rvz|\rvx^r)}} + o(\sqrt{\|\varepsilon\|}).
\end{align}
\textit{Proof}\\
Combining \textbf{Lemma 1} and \textbf{2} we get:
\begin{align}
    \TV{q^{(t)}(\rvz|\rvx^a)}{q_{\phi}(\rvz|\rvx^r)}
\underbrace{\leq }_{\text{Lemma 2}}
    &\TV{q^{(t)}(\rvz|\rvx^a)}{p_{\theta}(\rvz|\rvx^a)} \\
     &+ 
    \sqrt{\tfrac12 \KL{p_{\theta}(\rvz|\rvx^r)}{p_{\theta}(\rvz|\rvx^a)}} \\
    &+
   \sqrt{\tfrac12  \KL{q_{\phi}(\rvz|\rvx^r)}{p_{\theta}(\rvz|\rvx^r)}} \\
   \underbrace{=}_{\substack{\text{Lemma 1}}}
 &\TV{q^{(t)}(\rvz|\rvx^a)}{p_{\theta}(\rvz|\rvx^a)}\\
     &+ 
    \sqrt{\tfrac12 o(\|\varepsilon\|)  }\\
    &+
   \sqrt{\tfrac12  \KL{q_{\phi}(\rvz|\rvx^r)}{p_{\theta}(\rvz|\rvx^r)}}.
\end{align}

Note that $\sqrt{\tfrac12 o(\|\varepsilon\|)  } =  o(\sqrt{\|\varepsilon\|}) $ that gives us the final expression:

\begin{align}
    \TV{q^{(t)}(\rvz|\rvx^a)}{q_{\phi}(\rvz|\rvx^r)}
\leq 
 \TV{q^{(t)}(\rvz|\rvx^a)}{p_{\theta}(\rvz|\rvx^a)}
    +
   \sqrt{\tfrac12  \KL{q_{\phi}(\rvz|\rvx^r)}{p_{\theta}(\rvz|\rvx^r)}} + o(\sqrt{\|\varepsilon\|}).
\end{align}
$\hfill\blacksquare$


\newpage
\upd{
\section{Background on MCMC}\label{appendix:background}

\subsection{Sampling from an unnormalized density with MCMC}
Markov Chain Monte Carlo (MCMC) is a class of methods that are used to obtain samples from the density $p(\rvv)$ (also referred to as \textbf{target}), which is only known up to a normalizing constant. That is, we have access to $\tilde{p}(\rvv)$, such that $p(\rvv) = \frac{\tilde{p}(\rvv)}{Z}$ and $Z$ is a typically unknown and hard to estimate normalizing constant. Thus, we construct a Markov Chain with samples $\{\rvv^{(t)}\}_{t=1}^T$ so that they mimic the samples from $p(\rvv)$. To ensure they are proper samples, the stationary distribution of the constructed Markov Chain should be the target distribution $p(\rvv)$.

The most popular way of constructing such Markov Chains is the Metropolis-Hastings (MH) method. The majority of the MCMC methods used in practice can be formulated as a special case of the MH \cite{andrieu2003introduction}. In the MH method, we introduce a proposal distribution $q(\rvv^{t+1}|\rvv^t)$ to obtain a new sample and then accept it with the following probability:
\begin{equation}\label{eq:accept}
\mathcal{A}(\rvv^t, \rvv^{t+1})= \min\{ 1, \tfrac{p(\rvv^t)q(\rvv^{t+1}|\rvv^{t})}{p(\rvv^{t+1})q(\rvv^{t}|\rvv^{t+1})} \} .    
\end{equation}

If the point is not accepted, we reuse the previous point, i.e., $\rvv^{t+1} = \rvv^{t}$. It can be proven that the resulting chain of correlated samples converges in the distribution to the target density \cite{andrieu2003introduction}. 

It is worth mentioning that the performance of the method strongly depends on the choice of the proposal distribution. In higher dimensional spaces, it is especially important to incorporate the information about the geometry of the target distribution into the proposal density to improve the convergence time. The Hamiltonian Monte-Carlo (HMC) \cite{neal2011mcmc} is known to be one of the most efficient MCMC methods. It uses gradient of a target distribution in the proposal to incorporate the information about the geometry of the space.

The idea of the HMC is to introduce an auxiliary variable $\rvp$ with a known density (usually assumed to be the standard Gaussian) and the joint distribution formulated as follows: 

\begin{align}\label{eq:hmc}
p(\rvv, \rvp) &= \frac{1}{Z}\exp(-U(\rvv)) \exp(-K(\rvp)),   \\
\text{with:}& \notag\\
K(\rvp) &= - \frac{1}{2}\rvp^T\rvp, \\
U(\rvv) &= -\log \tilde{p}(\rvv). 
\end{align}

We obtain samples $(\rvv, \rvp)$ using the Hamiltonian dynamics \cite{neal2011mcmc} that describes how the $\rvv$ and $\rvp$ change over time for the given Hamiltonian $H(\rvv, \rvp) = U(\rvv) + K(\rvp)$, namely:

\begin{align}
\dot{\rvv} &= \frac{\partial H}{\partial \rvp},\\
\dot{\rvp} &= - \frac{\partial H}{\partial \rvv}.
\end{align}

For the practical implementation, these continuous-time equations are approximated by discretizing the time using $L$ small steps of size $\eta$. The discretization method that is often used is called the \textit{leapfrog}.

\subsection{The MCMC and Variational Autoencoders}
In this paper, we use the MCMC to sample from the posterior distribution $p_{\theta}(\rvz|\rvx^a)$. That is, in our case $\rvv = \rvz$ and $ \tilde{p}(\rvv) = p_{\theta}(\rvx^a|\rvz)p(\rvz)$. The HMC is a widely applied method to sampling from an unknown posterior distribution in deep learning (e.g. \cite{izmailov2021bayesian}). 
\ac{
A lot of effort was already done in combining variational inference with MCMC (and more specifically with HMC). Hamiltonian Variational Inference \cite{salimans2015markov, wolf2016variational} was proposed in order to obtain a more flexible variational approximation. Different approaches were proposed to use HMC during VAE training. \cite{hoffman2017learning} approximate the gradients of the likelihood and avoid the use of variational approximation. \cite{caterini2018hamiltonian} propose an unbiased estimate for the ELBO gradient, which allows training Hamiltonian Variational Autoencoder. \cite{ruiz2019contrastive} propose an alternative objective, which uses a contrastive divergence instead of standard KL-divergence. 

In this work we are not changing the training procedure, instead, we propose to only use HMC during evaluation. 

}

\paragraph{A possible extension to discrete latent spaces}
Some VAEs operate on discrete latent spaces, a very popular example would be a VQ-VAE \cite{van2017neural}. However, the classical HMC that we use in our experiments is not able to sample from a discrete distribution. Therefore, other MCMC methods should be used in this case, such as population-based MCMC \cite{auzina2021approximate}, modifications of HMC\cite{nishimura2020discontinuous} or Langevin Dynamics \cite{zhang2022langevin}.





}

\discussion{
\subsection{Mode optimization}\label{appendix:hmc_vs_opt}

In this work we hypothesise that adversarial attacks move latent codes to the region of low probability and we use HMC to get a sample from the high posterior probability region. However, another strategy can be to find the posterior mode instead. Here we explain, what was our motivation to not use this approach. 

\paragraph{Posterior modes similarity}
Ideally, we would like to obtain a sample from the variational posterior $q_{\phi}(\rvz|\rvx^r)$, because our decoder was trained to produce reconstructions from such latent codes. At the same time, VAE was trained to match this variational posterior to the true one $p_{\theta}(\rvz|\rvx^r)$. However, both these distributions are not available to us, since we observe attacked point $\rvx^a$ instead of the reference $\rvx^r$. 

Instead, we sample from $p_{\theta}(\rvz|\rvx^a)$ and show theoretically that the resulting samples are close (in terms of total variation distance) to the "goal" ones. However, that does guarantee that their modes are the same. Therefore, obtaining the mode of $p_{\theta}(\rvz|\rvx^a)$  is not necessarily a mode of $q_{\phi}(\rvz|\rvx^r)$. Thus, the fact that the HMC allows us to “wander” around that mode may be beneficial. 

\paragraph{Concentration of measure}
During reconstruction, we get a sample from $q(\rvz|\rvx)$ and pass it to the decoder, thus, a mode can actually be a bad latent code for these purposes. Instead, ideally, we want to get a sample from the typical set where most of the probability mass is concentrated. In theory, the HMC allows us to do that.

\paragraph{Randomness}
The HMC adds a source of randomness to our defence strategy that potentially makes it harder to attack. This is supported by our experiment in Section \ref{sec:ablation}
}

\newpage

\section{Additional results}
\upd{\subsection{Posterior ratio} \label{appendix:posterior_ratio}}

\upd{
We motivate our method by the hypothesis that the adversarial attack "shifts" a latent code to the region of a lower posterior density, while our approach moves it back to a high posterior probability region. In Section \ref{sec:defence} we theoretically justify our hypothesis, while here we provide an additional empirical evidence.

In order to verify our claim that applying an MCMC method allows us to counteract attacks by moving a latent code from a region of a lower posterior probability mass to a region of a higher density, we propose to quantify this effect by measuring the ratio of posteriors for $\rvz_1$ and $\rvz_2$. The true posterior $p(\rvz | \rvx^{r})$ is not available due to the cumbersome marginal distribution $p(\rvx^{r})$, however, we can calculate the ratio of posteriors because the marginal will cancel out, namely:

\begin{align}
\text{PR}(\rvz_1, \rvz_2) &= \tfrac{p_{\theta}(\rvz_1|\rvx^r)}{p_{\theta}(\rvz_2|\rvx^r)} \\
&= \tfrac{p_{\theta}(\rvx^r|\rvz_1)p(\rvz_1)}{p_{\theta}(\rvx^r| \rvz_2)p(\rvz_2)} .
\end{align}

In our case, we are interested in calculating the posterior ratio between the reference and adversarial latent codes ($\rvz_1 = \rvz^r$,  $\rvz_2 = \rvz^a$) as the baseline, and the posterior ratio between the reference and adversarial code after applying the HMC ($\rvz_1 = \rvz^r$ , $\rvz_2 = \rvz^a_{\text{HMC}}$). The lower the posterior ratio, the better. For practical reasons, we use the logarithm of the posterior ratio (the logarithm does not change the monotonicity and turns products to sums):
\begin{equation}
    \log \text{PR}(\rvz_1, \rvz_2) = \log p_{\theta}(\rvx^r|\rvz_1) + \log p(\rvz_1) - \log p_{\theta}(\rvx^r| \rvz_2) - \log p(\rvz_2) .
\end{equation}

We present results on the log-posterior-ratio calculated on the MNIST dataset. In Figure \ref{fig:mnist_post_ratio} we show a plot with two histograms: one with the posterior ratio between the reference and adversarial latent codes ($\rvz_1 = \rvz^r$,  $\rvz_2 = \rvz^a$) in blue, and the second histogram of the posterior ratio between the reference and adversarial code after applying the HMC ($\rvz_1 = \rvz^r$ , $\rvz_2 = \rvz^a_{\text{HMC}}$) in orange. 



We observe that the histogram has moved to the left after applying the HMC. This indicates that posterior of the adversarial (in the denominator) is increasing when the HMC is used. This is precisely the effect we hoped for and this result provides an empirical evidence in favor of our hypothesis.
}

\begin{figure}[ht]
    \centering
        \includegraphics[width=0.45\textwidth]{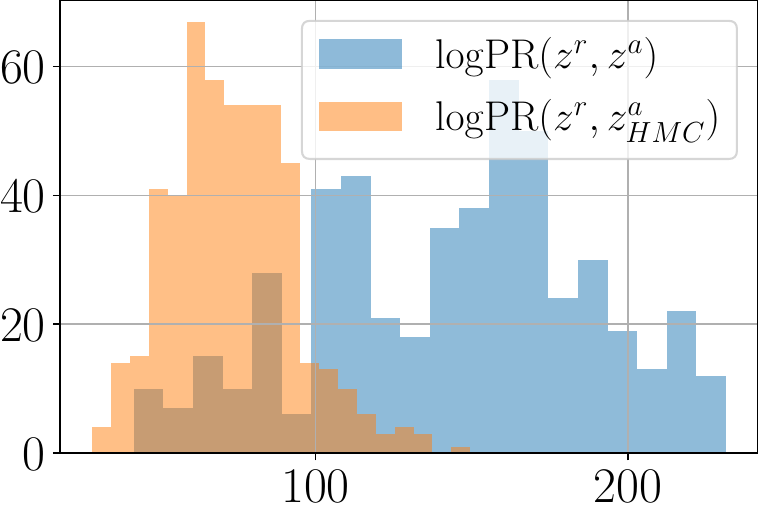}
    \caption{Histograms of the log posterior ratios without HMC (blue) and with HMC (orange) evaluated on MNIST dataset.}
    \label{fig:mnist_post_ratio}
\end{figure}

\discussion{
\paragraph{Experimental Details}
For this experiment we construct 500 adversarial attacks with the radius 0.1 on the encoder of the VAE trained on MNIST dataset. We run 500 HMC steps with the same hyperparameters as mentioned in Table \ref{tab:setup} to obtain $\rvz^a_{HMC}$.

\paragraph{Statistical Analisys}
We performed a two-sample Kolmagorov-Smirnov test with the null hypothesis that two histograms are drawn from the same distribution. As an alternative hypothesis is that the underlying distributions are different. Choosing the confidence level of 95\% results in the rejection of the null hypothesis (p-value is equal to 0.029) in favour of the alternative: two histograms were not drawn from the same distribution.

}

\newpage
\subsection{Detailed results for $\beta$-VAE and $\beta$-TCVAE} \label{appendix:beta_vae}
In this section we report extended results for MNIST, FashionMNIST and ColorMNIST datasets. We train VAE,  $\beta$-VAE and $\beta$-TCVAE on three datasets: MNIST, FashionMNIST and ColorMNIST. Then, we compare the robustness to adversarial attack with and without HMC. We present all the result with the standard error in Table $\ref{tab:mnist_results}$. On Figures \ref{fig:mnist_rec_similarity}, \ref{fig:mnist_adv_acc} we show visually how HMC improve the robustness for each dataset and model. 

\begin{table*}[ht]
\caption{Robustness results on MNIST, Fashion MNIST and Color MNIST datasets. We perform unsupervised attack with radius $0.1$ (top) and $0.2$ (bottom). We attack the encoder (left) and the downstream classification task (right). Higher values correspond to more robust models.\\
}
\vskip -0.2cm
\label{tab:mnist_results}
\begin{center}
\begin{sc}
\resizebox{.99\textwidth}{!}{
\begin{tabular}{llccc|cccc}
\toprule
& & \multicolumn{3}{c|}{$\mathrm{MSSSIM}[\widetilde{\rvx}^{r}, \widetilde{\rvx}^{a}]$ $\uparrow$} & \multicolumn{3}{c}{Adversarial Accuracy $\uparrow$} \\
 & & \multirow{2}{*}{\sc{MNIST}} &  \multirow{2}{*}{\sc{Fashion MNIST}} &  \multirow{2}{*}{\sc{Color MNIST}}  &  \multirow{2}{*}{\sc{MNIST}} &  \multirow{2}{*}{\sc{Fashion MNIST}} & \multicolumn{2}{c}{\sc{Color MNIST}} \\
 & &  &  &   & &  & Digit & Color \\
\midrule
\multirow{7}{*}{\STAB{\rotatebox[origin=c]{90}{$\|\varepsilon \|_{\inf} = 0.1$}}} 
& VAE              & 0.70 \tiny{(0.02)}          & 0.59 \tiny{(0.03)}          & 0.36 \tiny{(0.03)}           & 0.08 \tiny{(0.04)}          & 0.00 \tiny{(0.01)}          & 0.04 \tiny{(0.03)}          & 0.06 \tiny{(0.03)}  \\
& \quad\quad + HMC & \textbf{0.88 \tiny{(0.01)}} & \textbf{0.66 \tiny{(0.03)}} & \textbf{0.96 \tiny{(0.01)}}  & 0.25 \tiny{(0.03)}          & \textbf{0.14 \tiny{(0.02)}} & 0.16 \tiny{(0.02)}          & 0.68 \tiny{(0.03)} \\
& $\beta$-VAE      & 0.75 \tiny{(0.01)}          & 0.52 \tiny{(0.03)}          & 0.50 \tiny{(0.04)}           & 0.11 \tiny{(0.04)}          & 0.00 \tiny{(0.02)}          & 0.08 \tiny{(0.04)}          & 0.21 \tiny{(0.06)} \\
& \quad\quad + HMC & 0.84 \tiny{(0.01)}          & 0.64 \tiny{(0.03)}          & 0.92 \tiny{(0.03)}           & \textbf{0.30 \tiny{(0.03)}} & 0.13 \tiny{(0.02)}          & 0.14 \tiny{(0.02)}          & 0.66 \tiny{(0.04)} \\
& $\beta$-TCVAE    & 0.70 \tiny{(0.02)}          & 0.52 \tiny{(0.03)}          & 0.35 \tiny{(0.02)}           & 0.05 \tiny{(0.03)}          & 0.01 \tiny{(0.01)}          & 0.08 \tiny{(0.04)}          & 0.06 \tiny{(0.03)} \\
& \quad\quad + HMC & 0.79 \tiny{(0.02)}          & \textbf{0.66 \tiny{(0.03)}} & \textbf{0.96 \tiny{(0.01)}}  & 0.25 \tiny{(0.04)}          & 0.13 \tiny{(0.02)}          & \textbf{0.22 \tiny{(0.03)}} & \textbf{0.81 \tiny{(0.02)}} \\ 
\midrule
\multirow{6}{*}{\STAB{\rotatebox[origin=c]{90}{$\|\varepsilon \|_{\inf} = 0.2$}}} 
& VAE              & 0.36 \tiny{(0.03)}          & 0.47 \tiny{(0.03)}          & 0.19 \tiny{(0.02)}          & 0.05 \tiny{(0.03)}          & 0.01 \tiny{(0.01)}          & 0.02 \tiny{(0.02)}          & 0.06 \tiny{(0.03)}\\
& \quad\quad + HMC & \textbf{0.76 \tiny{(0.02)}} & \textbf{0.54 \tiny{(0.03)}} & \textbf{0.90 \tiny{(0.01)}} & \textbf{0.19 \tiny{(0.03)}} & \textbf{0.13 \tiny{(0.02)}} & 0.11 \tiny{(0.02)}          & 0.62 \tiny{(0.03)}\\
& $\beta$-VAE      & 0.50 \tiny{(0.03)}          & 0.41 \tiny{(0.03)}          & 0.38 \tiny{(0.04)}          & 0.01 \tiny{(0.01)}          & 0.00 \tiny{(0.01)}          & 0.05 \tiny{(0.03)}          & 0.18 \tiny{(0.05)}\\
& \quad\quad + HMC & 0.69 \tiny{(0.03)}          & 0.50 \tiny{(0.03)}          & 0.87 \tiny{(0.01)}          & 0.16 \tiny{(0.03)}          & 0.12 \tiny{(0.02)}          & \textbf{0.15 \tiny{(0.02)}} & 0.56 \tiny{(0.04)}\\
& $\beta$-TCVAE    & 0.45 \tiny{(0.03)}          & 0.42 \tiny{(0.03)}          & 0.20 \tiny{(0.02)}          & 0.03 \tiny{(0.02)}          & 0.02 \tiny{(0.02)}          & 0.05 \tiny{(0.03)}          & 0.05 \tiny{(0.03)}\\
& \quad\quad + HMC & 0.65 \tiny{(0.03)}          & 0.52 \tiny{(0.03)}          & 0.87 \tiny{(0.01)}          & 0.16 \tiny{(0.04)}          & 0.11 \tiny{(0.02)}          & 0.14 \tiny{(0.02)}          & \textbf{0.72 \tiny{(0.03)}} \\
\bottomrule
\end{tabular}}
\end{sc}
\end{center}
\vskip -0.1in
\end{table*}

\begin{figure}[ht]
    \centering
    \begin{tabular}{ll}
        \includegraphics[width=0.5\textwidth]{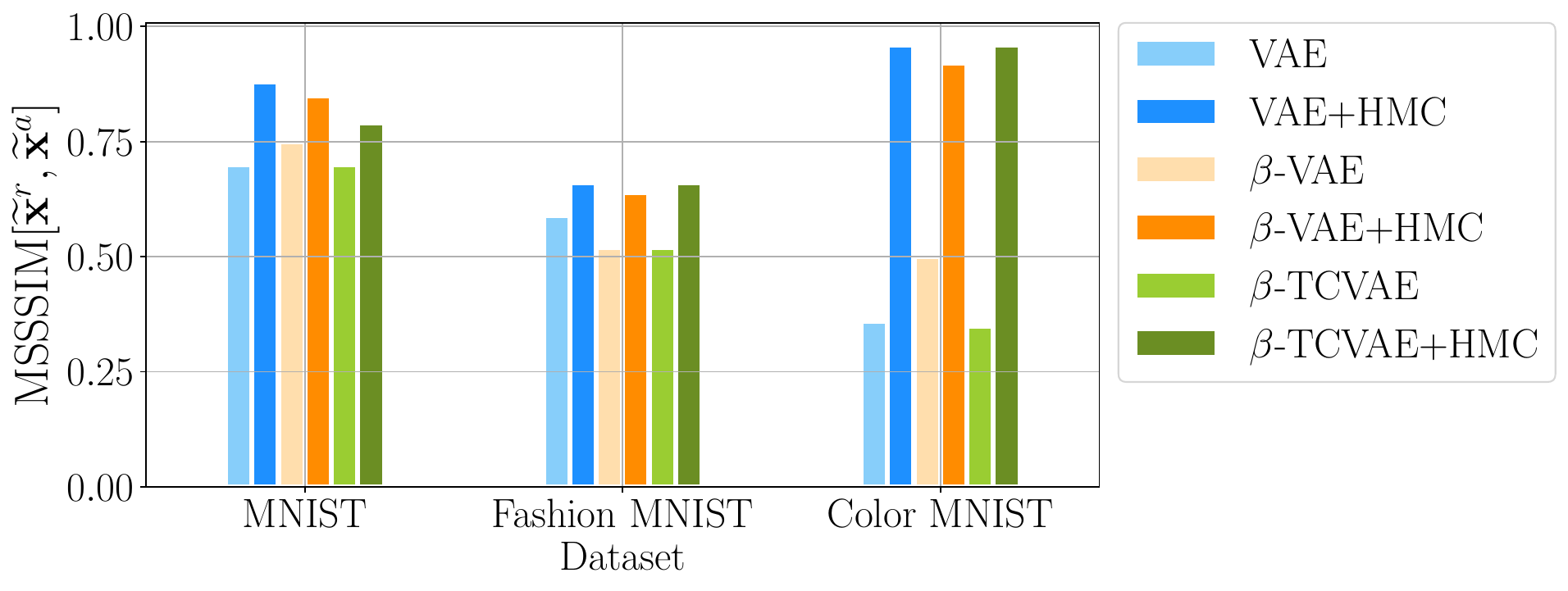} & 
        \includegraphics[width=0.5\textwidth]{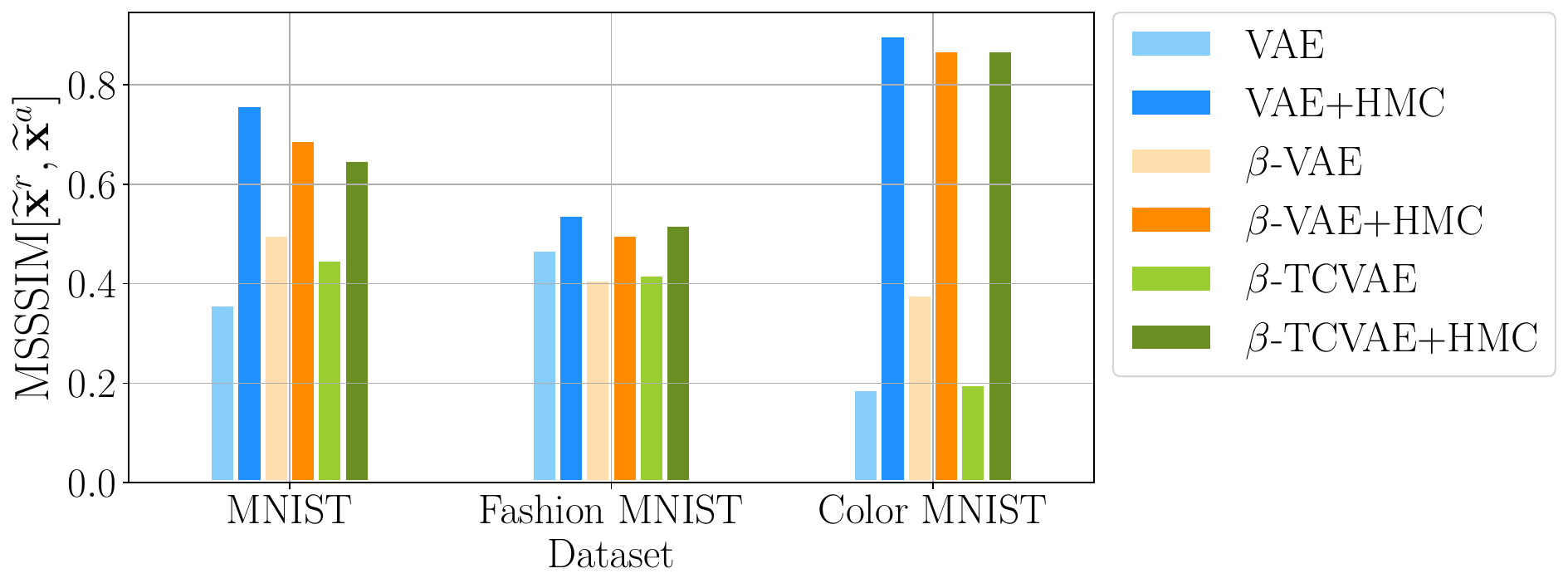} \\
        \multicolumn{1}{c}{(a) $||\varepsilon ||_{\inf} = 0.1$} &
        \multicolumn{1}{c}{(b) $||\varepsilon ||_{\inf} = 0.2$} \\
    \end{tabular}
    \caption{Improvement of the Reconstruction Similarity after the proposed defence. We fix the attack radius to be equal to (a) 0.1 and (b) 0.2. Higher values correspond to a more robust representations.}
    \vskip -10pt
    \label{fig:mnist_rec_similarity}
\end{figure}

\begin{figure}[ht]
    \centering
    \begin{tabular}{ll}
        \includegraphics[width=0.5\textwidth]{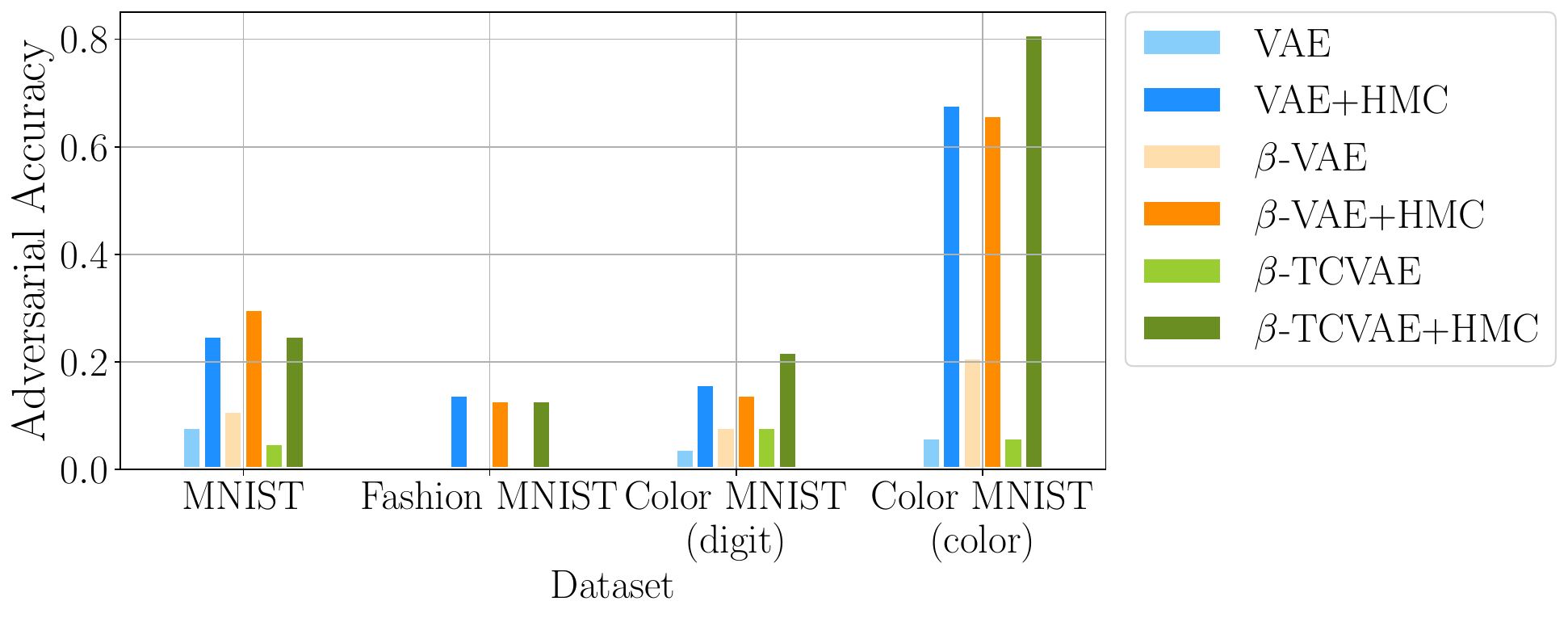} &
        \includegraphics[width=0.5\textwidth]{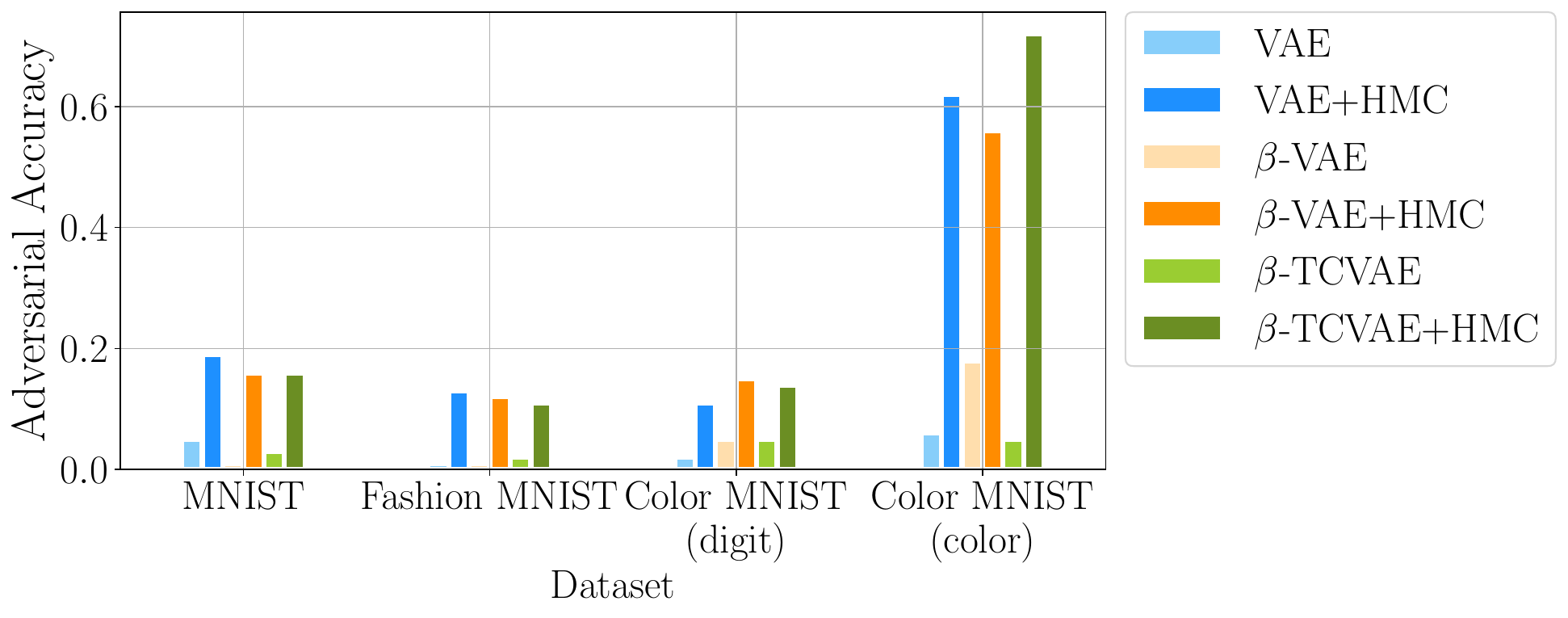} \\
        \multicolumn{1}{c}{(a) $||\varepsilon ||_{\inf} = 0.1$} &
        \multicolumn{1}{c}{(b) $||\varepsilon ||_{\inf} = 0.2$} \\
    \end{tabular}
    \caption{Improvement of the Adversarial Accuracy after proposed defence. We fix the attack radius to be equal to (a) 0.1 and (b) 0.2.}
    \vskip -10pt
    \label{fig:mnist_adv_acc}
\end{figure}

\newpage

\subsection{What if the attacker knows the defence strategy?} \label{appendix:attack_mcmc}
In our experiments we relied on the assumption that attack does not take into account the defence strategy that we use. We believe that it is reasonable, since defence requires access to the decoder part of the model ($p_{\theta}(x|z)$), which is not necessarily available to the attacker. 

However, one may assume that the defence strategy is known to the attacker. In this case, it is reasonable to verify whether the robustness results change. In the conducted experiment we show that it is vastly more complicated to attack the encoder with taking the MCMC defence into account. We train the unsupervised attack (\ref{eq:objective_unsup}). The attack has access to the encoder and MCMC defence:
\begin{equation}
f(x) = q^{(t)}(\rvz|\rvx) = \int Q^{(t)}(\rvz|\rvz_0) q_{\phi}(\rvz_0|\rvx) d\rvz_0,
\end{equation}
where $Q^{(t)}(\rvz|\rvz_0)$ is MCMC kernel. 

Then, given the attack radius $\delta$, we train the attack using the following objective:
\begin{align}
    \varepsilon^* &= \arg\max_{\|\varepsilon\|_{\inf} < \delta}\|\widetilde{z}^a - \widetilde{z}^r\|^2, \\
    \widetilde{z}^a &\sim q^{(t)}(\rvz|\rvx^r + \varepsilon), \label{eq:z_a_mcmc}\\
    \widetilde{z}^r &\sim q^{(t)}(\rvz|\rvx^r). \label{eq:z_r_mcmc}
\end{align}

The similarity results of these attacks are plotted in Figure \ref{fig:mcmc_attack}. We observe that the reconstructed reference and adversarial points have approximately the same similarity (measured by MSSSIM) as the initial points $\rvx^r$ and $\rvx^a$, which indicates that the attacks were unsuccessful. 

However, if we use the same objective, but omit the MCMC step (e.g $t=0$ in eq. \ref{eq:z_a_mcmc} and \ref{eq:z_r_mcmc}), then, as observed in Figure \ref{fig:no_mcmc_attack}, the attack becomes much more successful (Figure \ref{fig:no_mcmc_attack} (a)), but we can fix it with the proposed defence (Figure \ref{fig:no_mcmc_attack} (b)).

It is interesting to compare how the attacked points look in both cases, especially as we increase the radius of the attack. In Figure \ref{fig:mcmc_attack_example}, we plot attack on two reference points for radius values in $\{0.1, 0.6, 0.8, 1.0\}$. When the attacker does not use MCMC (left), it just learns to add more and more noise to the image, which eventually makes it meaningless. 
 
 When we use MCMC during an attack, the situation is different. The adversarial input is almost indistinguishable from the reference point for a small radius. After each gradient update, the attacker runs a new MCMC, which moves point closer to the region of high posterior probability, but may follow a different trajectory every time. Eventually, it makes it harder to learn an additive perturbation $\varepsilon$. However, as we increase the attack radius, we observe a very interesting effect. Instead of meaningless noise, the attacker learns to change the digit. For instance, we see how $4$ is transformed into $0$ in the first example and into $9$ in the second. This way, the attacker ensures that the MCMC will move the latent far away from the reference latent code, which now has a different posterior distribution.

\begin{figure}[ht]
    \centering
    \begin{tabular}{cc}
      \includegraphics[width=0.35\columnwidth]{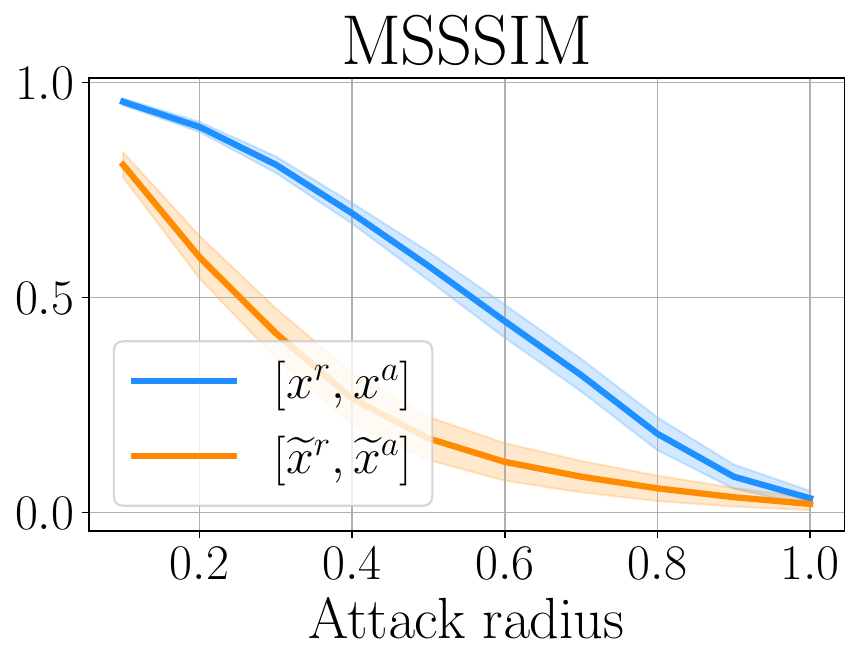} & \qquad
     \includegraphics[width=0.35\columnwidth]{img/attack_mcmc_0_100.pdf} \\
     \Large{(a) No defence} & \Large{(b) MCMC defence} \\
    \end{tabular}
    \caption{Adversarial attack, if attacker \textbf{does not use MCMC}. We report similarity of the reference and adversarial point before forward pass (blue) and after forward pass (orange). }
    \label{fig:no_mcmc_attack}
\end{figure}

\begin{figure}[ht]
    \centering
    \begin{tabular}{cc}
      \includegraphics[width=0.35\columnwidth]{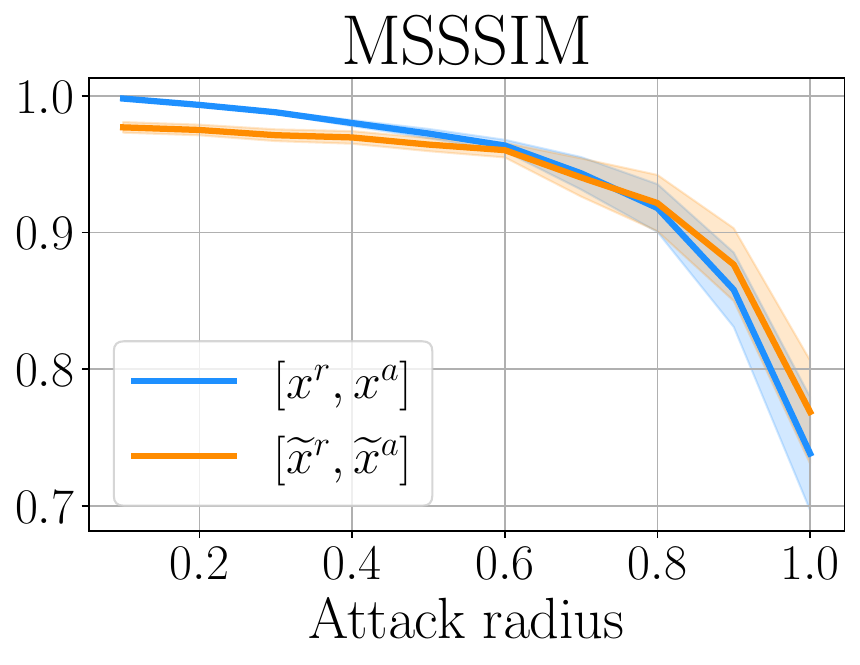} & \qquad
     \includegraphics[width=0.35\columnwidth]{img/attack_mcmc_1_100.pdf} \\
     \Large{(a) No defence} & \Large{(b) MCMC defence} \\
    \end{tabular}
    \caption{Adversarial attack, if attacker \textbf{uses MCMC}. We report similarity of the reference and adversarial point before forward pass (blue) and after forward pass (orange). }
    \label{fig:mcmc_attack}
\end{figure}

\begin{figure}[H]
    \centering
    \begin{tabular}{cccc|ccc}
    \multirow{2}{*}{Radius} & \multicolumn{3}{c|}{Attacker doe not use MCMC} & \multicolumn{3}{c}{Attacker uses MCMC} \\
        & $\rvx^a$ & $\widetilde{\rvx}^a$ & $\widetilde{\rvx}^a_{\text{HMC}}$ & $\rvx^a$ & $\widetilde{\rvx}^a$ & $\widetilde{\rvx}^a_{\text{HMC}}$ \\
        0.1 & \includegraphics[width=0.1\linewidth]{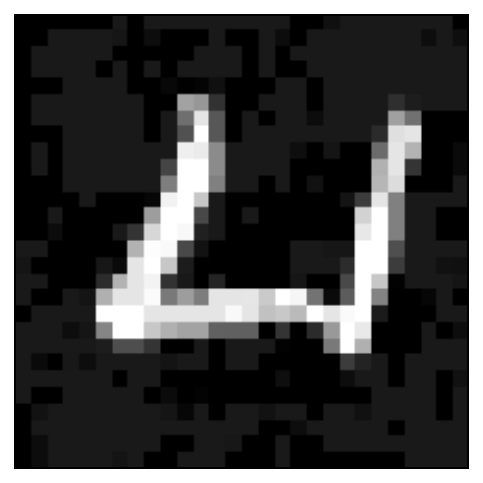} &
        \includegraphics[width=0.1\linewidth]{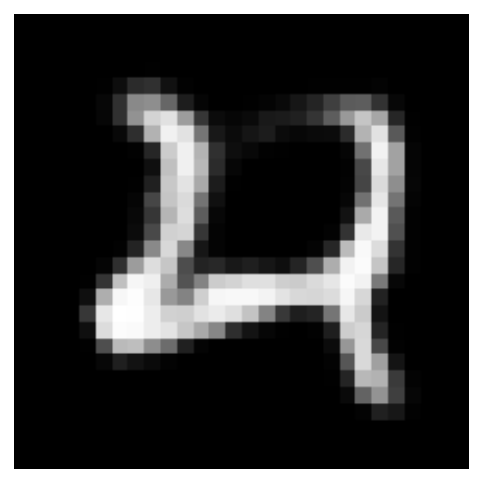} &
        \includegraphics[width=0.1\linewidth]{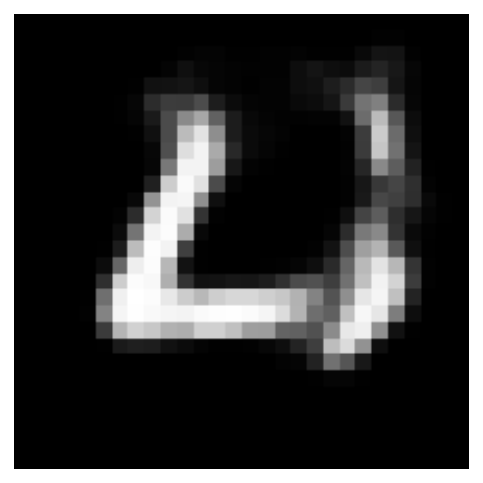} &
        \includegraphics[width=0.1\linewidth]{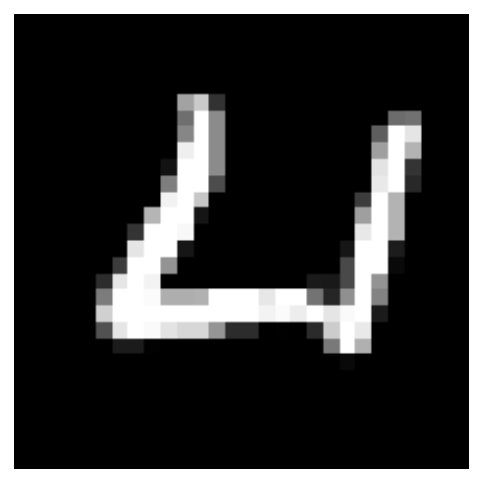} &
        \includegraphics[width=0.1\linewidth]{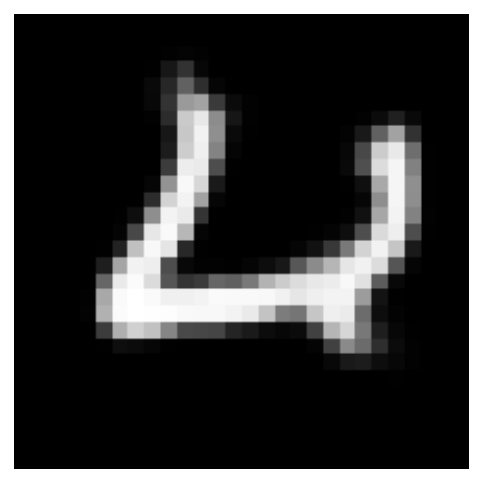} &
        \includegraphics[width=0.1\linewidth]{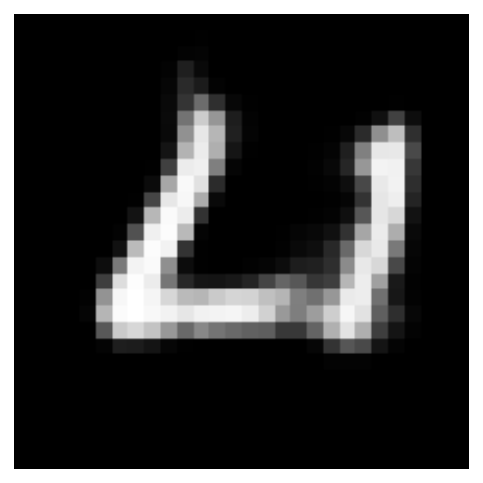}\\ 
        0.6 & \includegraphics[width=0.1\linewidth]{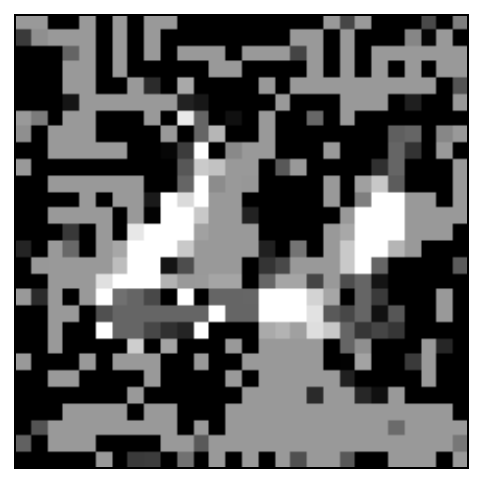} &
        \includegraphics[width=0.1\linewidth]{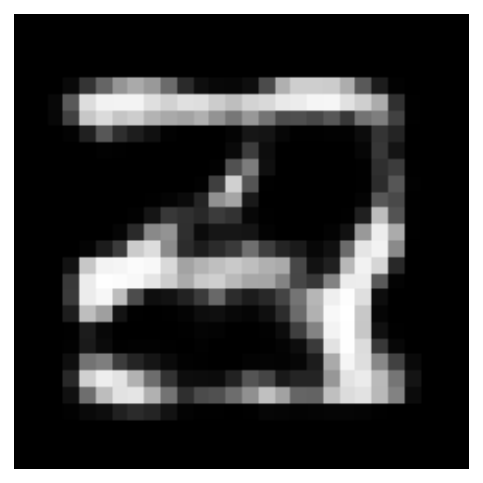} &
        \includegraphics[width=0.1\linewidth]{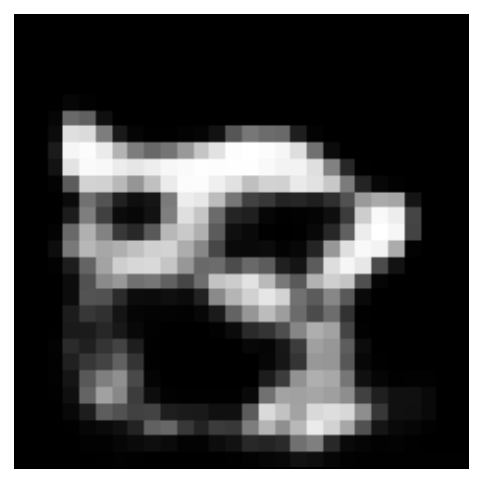} &
        \includegraphics[width=0.1\linewidth]{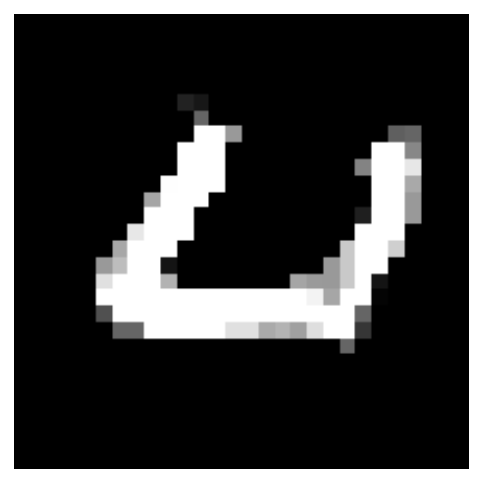} &
        \includegraphics[width=0.1\linewidth]{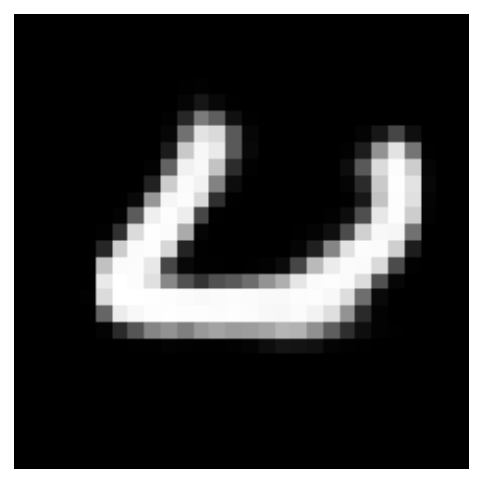} &
        \includegraphics[width=0.1\linewidth]{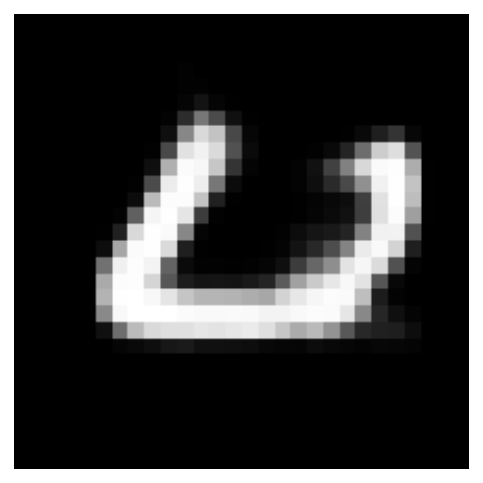} \\ 
        0.8 & \includegraphics[width=0.1\linewidth]{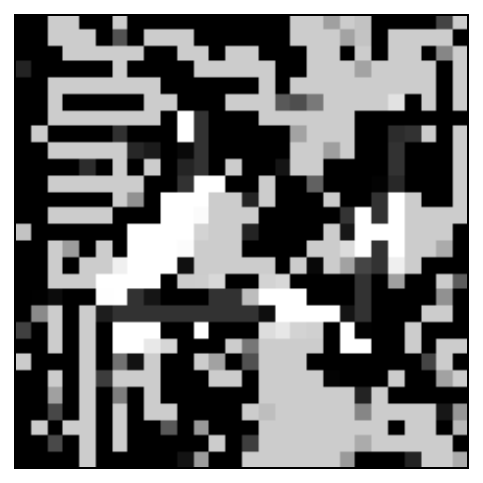} &
        \includegraphics[width=0.1\linewidth]{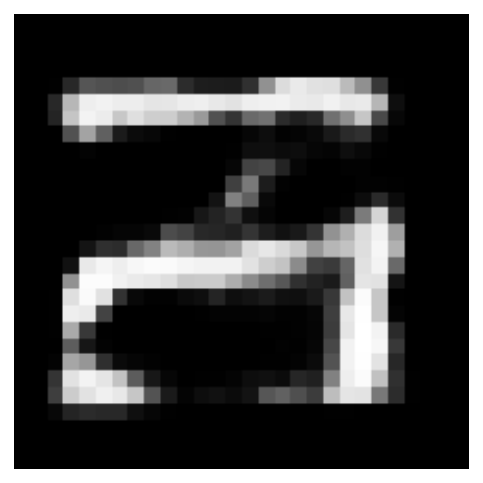} &
        \includegraphics[width=0.1\linewidth]{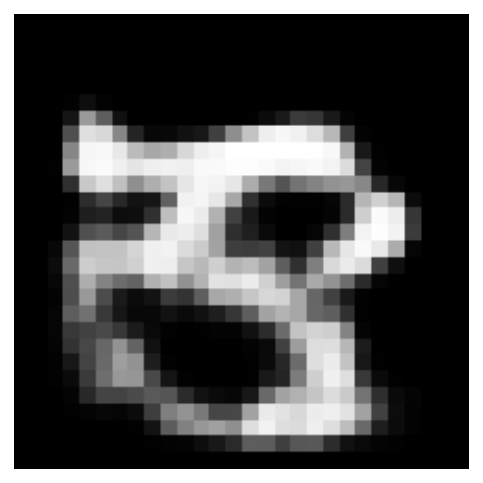} &
        \includegraphics[width=0.1\linewidth]{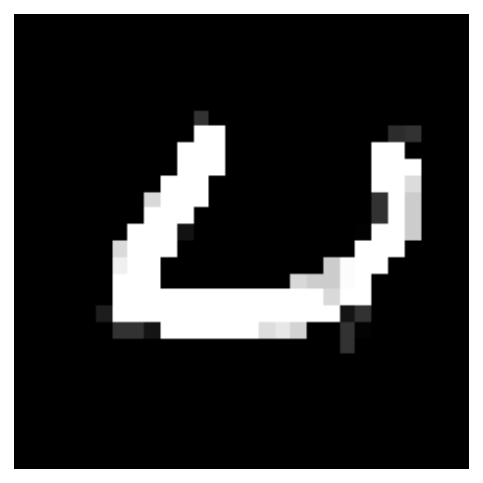} &
        \includegraphics[width=0.1\linewidth]{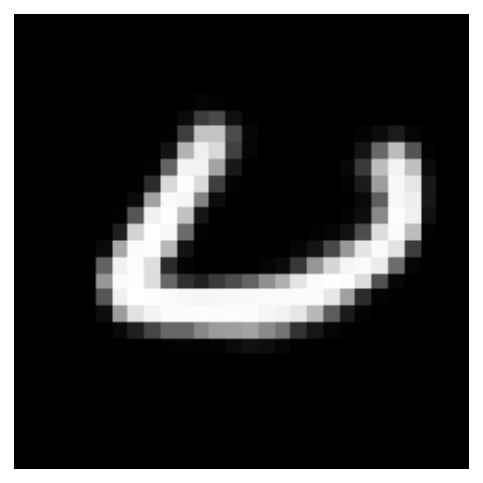} &
        \includegraphics[width=0.1\linewidth]{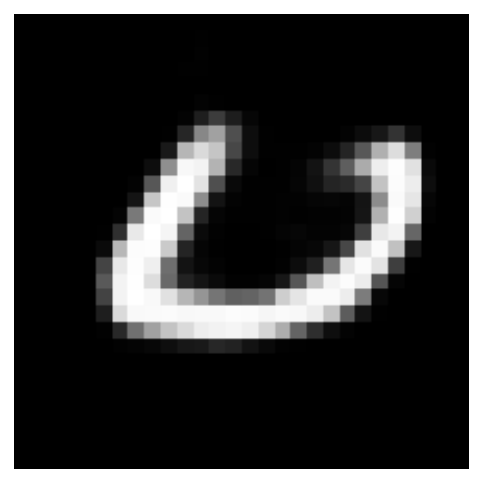} \\ 
        1.0 & \includegraphics[width=0.1\linewidth]{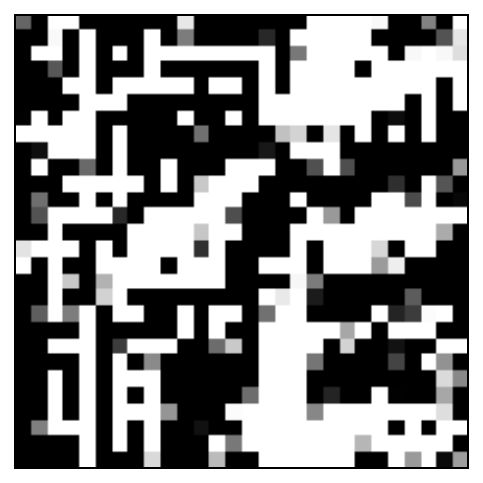} & 
         \includegraphics[width=0.1\linewidth]{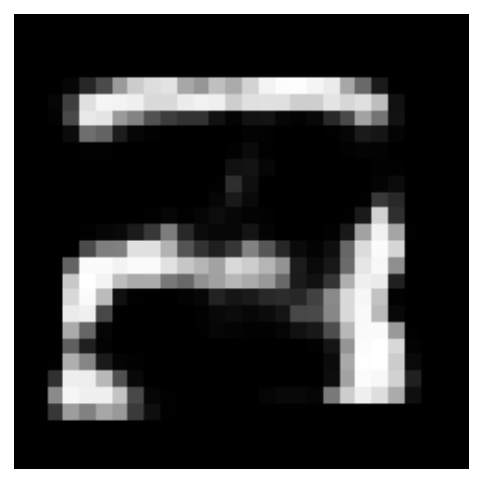} & 
          \includegraphics[width=0.1\linewidth]{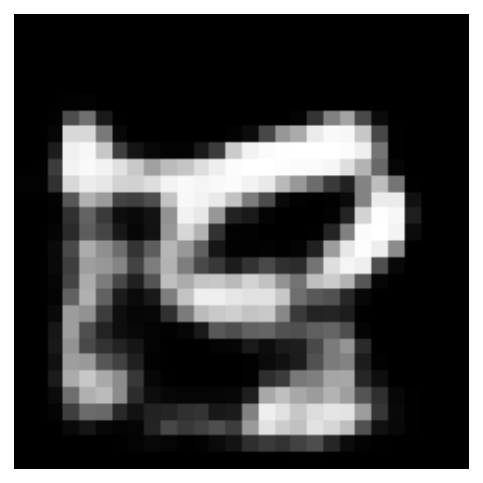} &
           \includegraphics[width=0.1\linewidth]{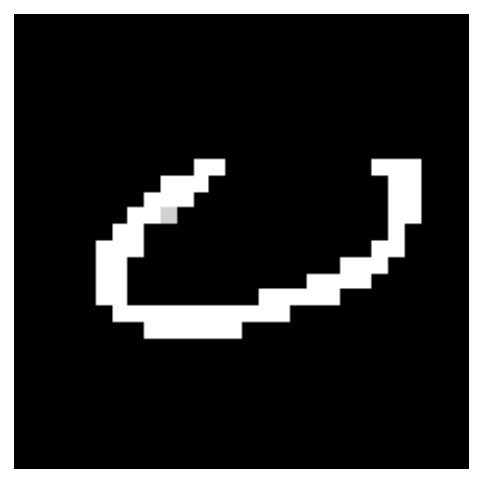} &
           \includegraphics[width=0.1\linewidth]{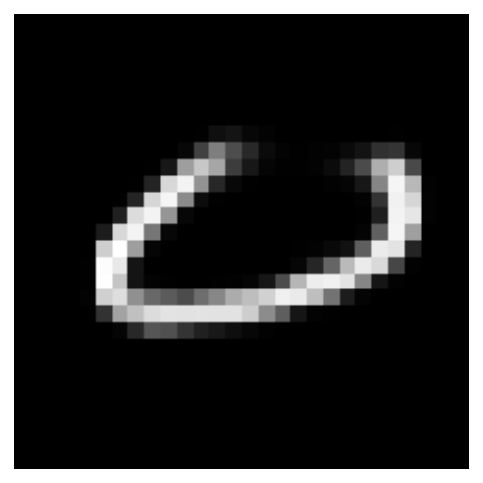} &
           \includegraphics[width=0.1\linewidth]{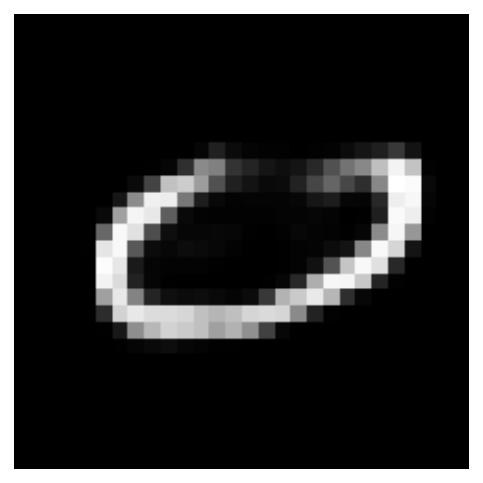} \\\midrule
         0.1 & \includegraphics[width=0.1\linewidth]{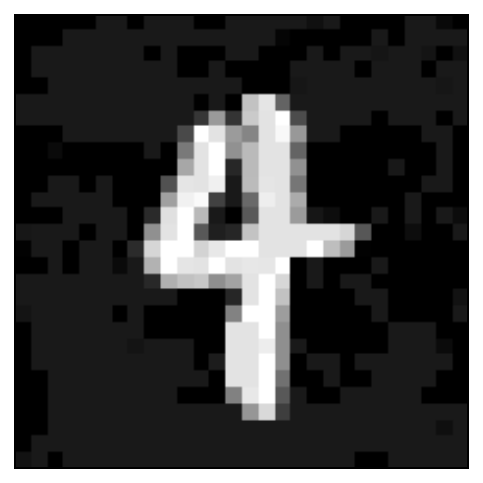} &
        \includegraphics[width=0.1\linewidth]{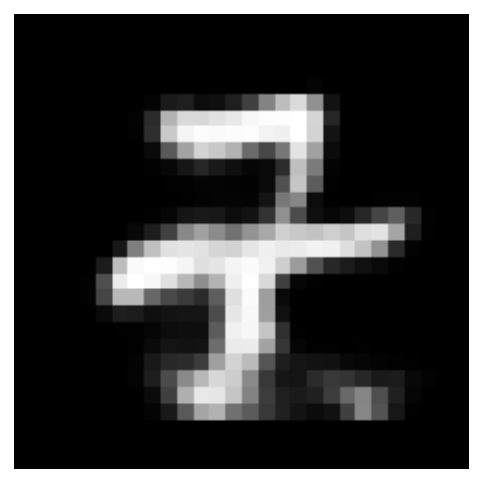} &
        \includegraphics[width=0.1\linewidth]{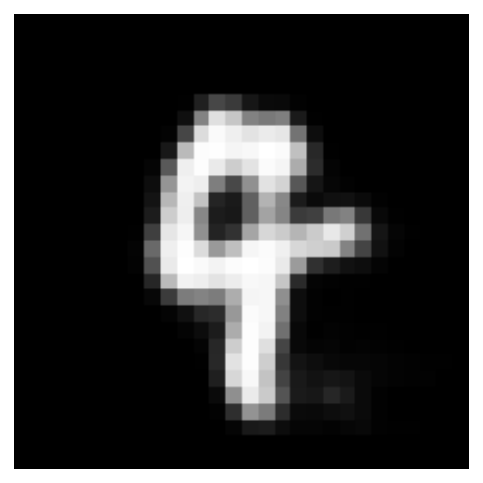} &
        \includegraphics[width=0.1\linewidth]{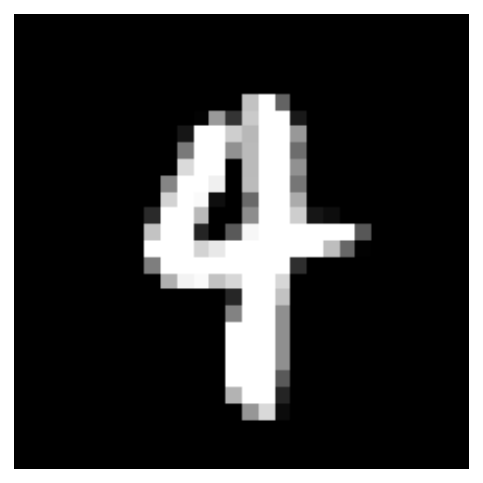} &
        \includegraphics[width=0.1\linewidth]{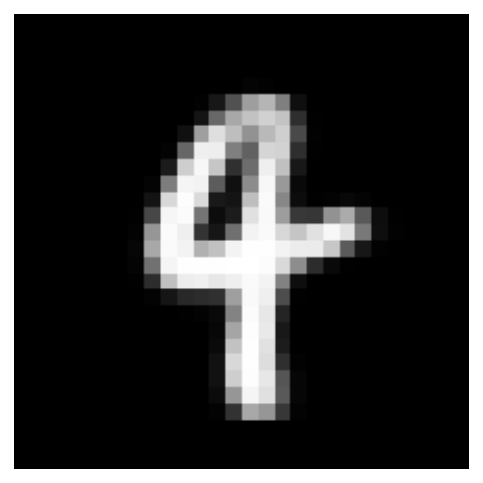} &
        \includegraphics[width=0.1\linewidth]{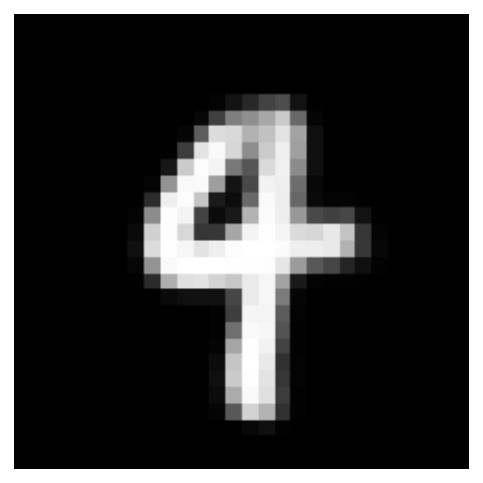}\\ 
        0.6 & \includegraphics[width=0.1\linewidth]{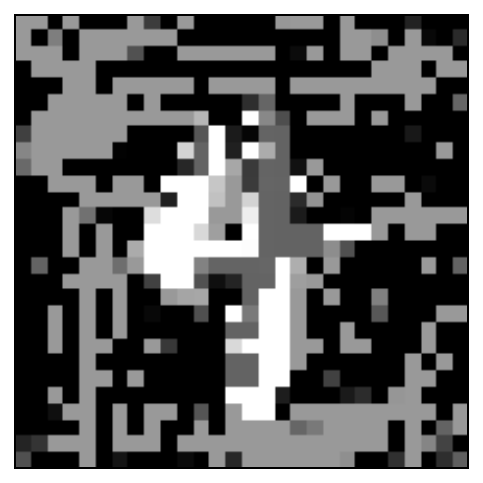} &
        \includegraphics[width=0.1\linewidth]{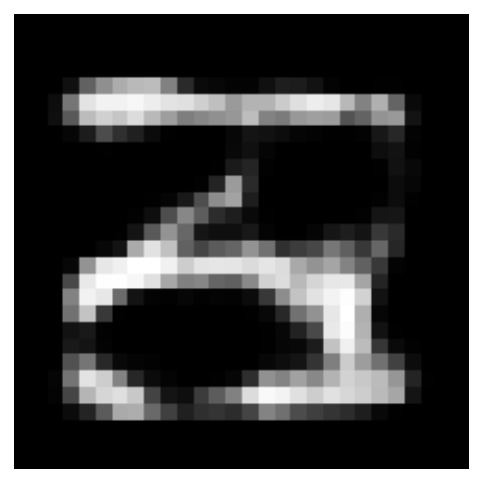} &
        \includegraphics[width=0.1\linewidth]{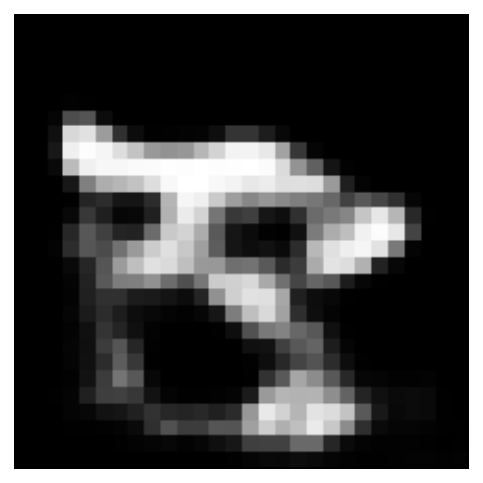} &
        \includegraphics[width=0.1\linewidth]{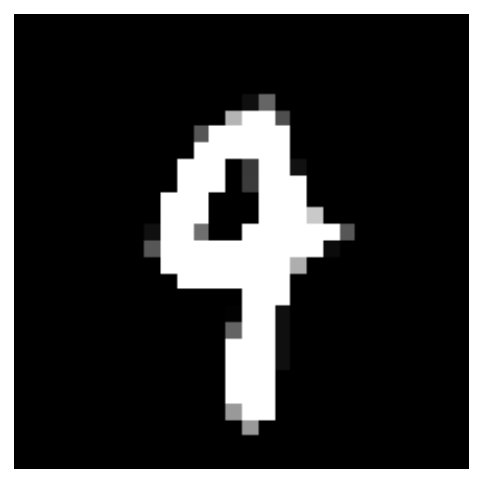} &
        \includegraphics[width=0.1\linewidth]{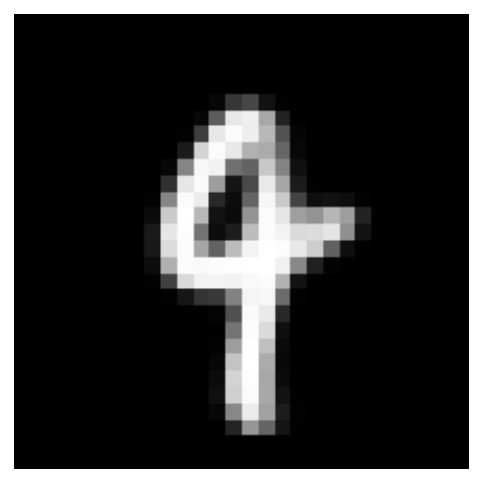} &
        \includegraphics[width=0.1\linewidth]{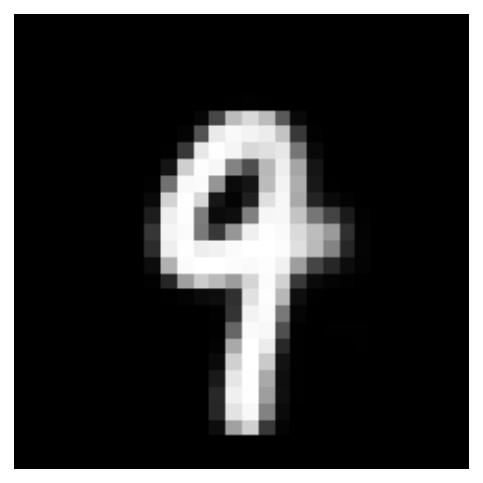} \\ 
        0.8 & \includegraphics[width=0.1\linewidth]{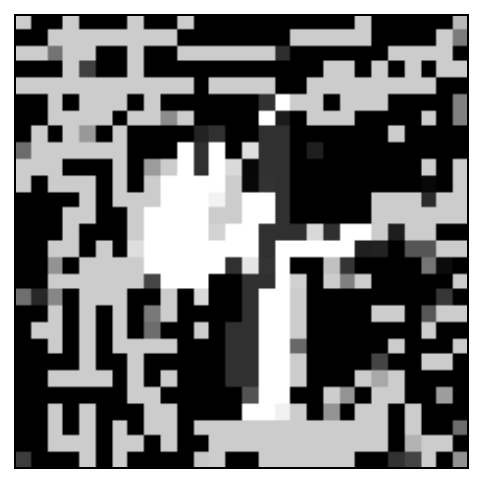} &
        \includegraphics[width=0.1\linewidth]{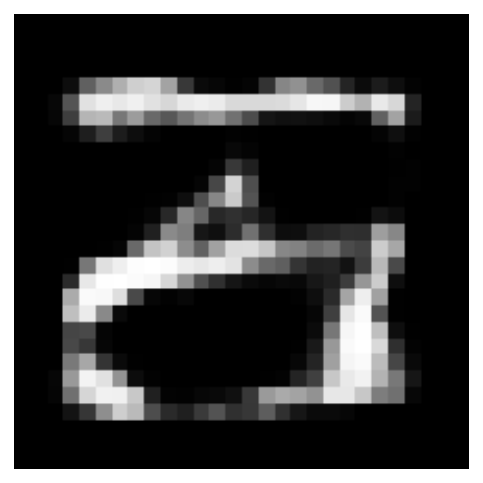} &
        \includegraphics[width=0.1\linewidth]{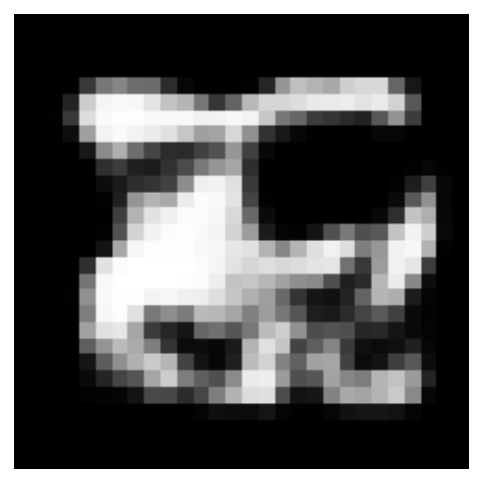} &
        \includegraphics[width=0.1\linewidth]{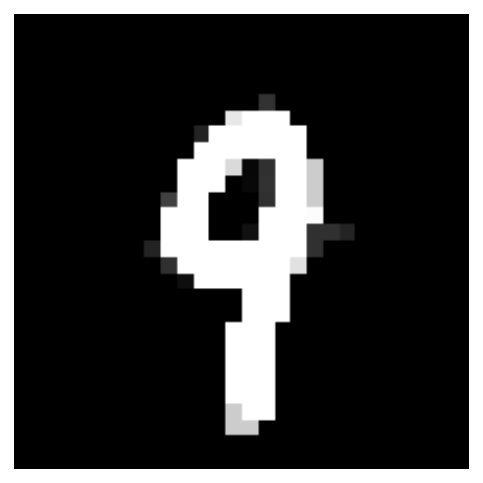} &
        \includegraphics[width=0.1\linewidth]{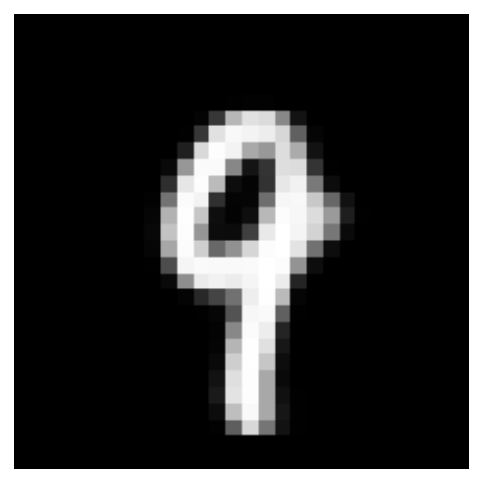} &
        \includegraphics[width=0.1\linewidth]{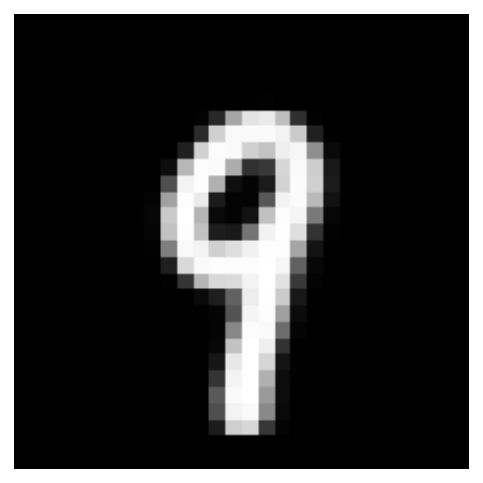} \\ 
        1.0 & \includegraphics[width=0.1\linewidth]{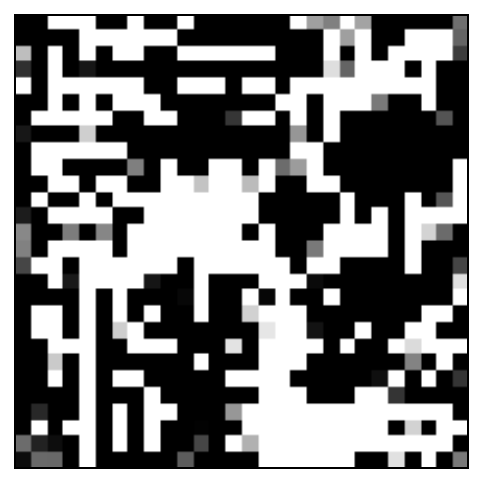} & 
         \includegraphics[width=0.1\linewidth]{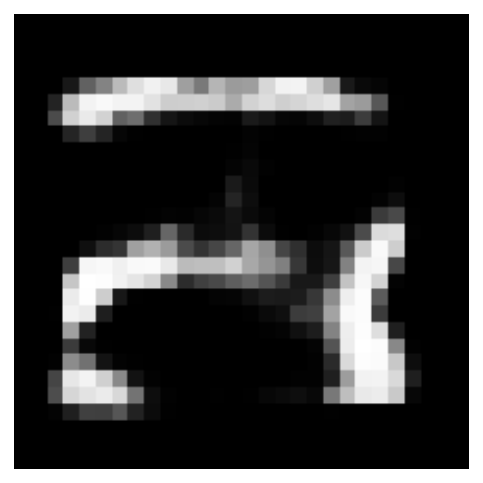} & 
          \includegraphics[width=0.1\linewidth]{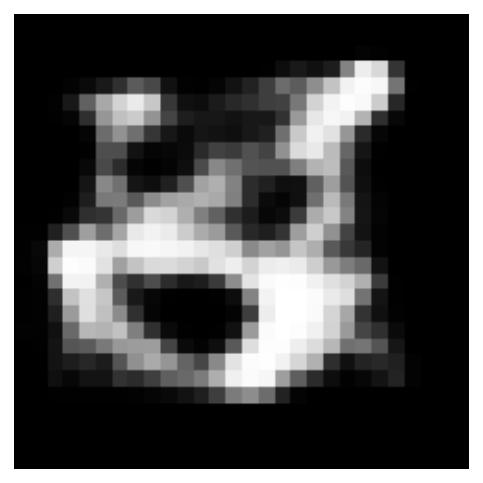} &
           \includegraphics[width=0.1\linewidth]{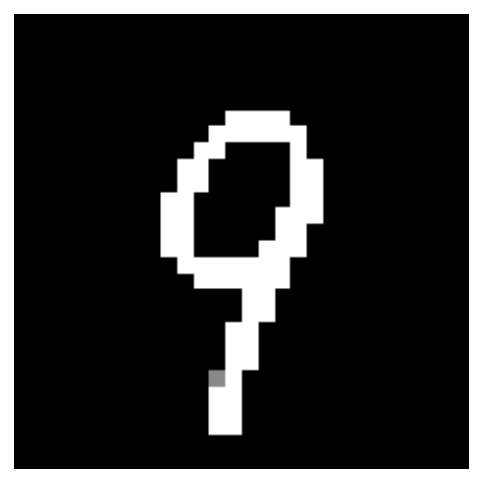} &
           \includegraphics[width=0.1\linewidth]{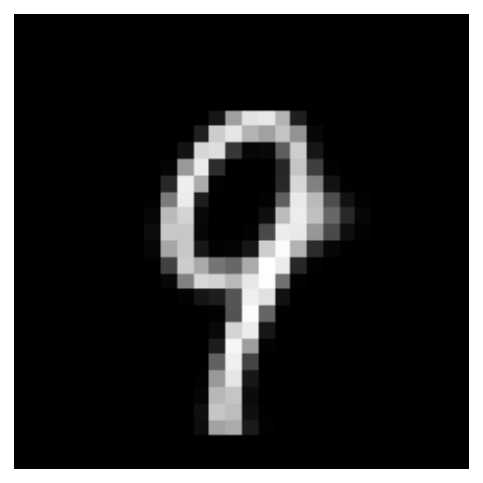} &
           \includegraphics[width=0.1\linewidth]{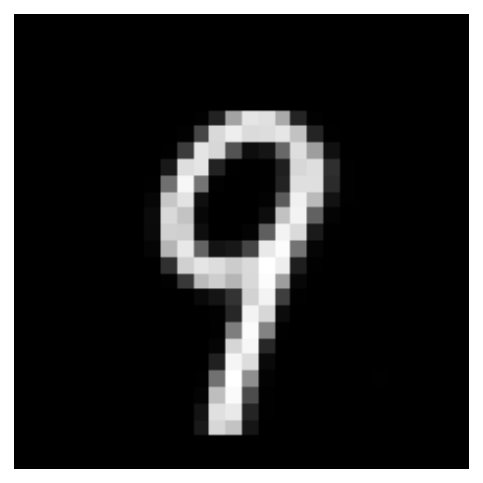} \\
    \end{tabular}
    \caption{Examples of adversarial point and their reconstructions, when attacker does not use MCMC (left) and when attacker uses MCMC(right).}
    \label{fig:mcmc_attack_example}
\end{figure}

\newpage



\subsection{Which attack radius should be considered?}\label{appendix:attack_radius}

In out experiments, we use attacks with the radius $0.1$ and $0.2$ for all the models except for CelebA dataset, where radii $0.05$ and $0.1$ were considered. Here, we provide additional experiment to justify this choice. In Figure \ref{fig:radius_metrics} (a) we show the similarity between the reference point and the adversarial point. We observe that for CelebA the similarity drops faster than for the MNIST. Further, if we look at the example plotted in Figure \ref{fig:radius_adversarial_examples}, we can clearly notice that with the radius $0.2$ CelebA image is already containing a lot of noise. At the same time, we observe (Figure \ref{fig:radius_metrics} (b)) that reconstruction similarity, which indicates the success of the attack, drops relatively fast when the radius of the attack increases. 

\begin{figure}[ht]
    \centering
    \begin{tabular}{cc}
        \includegraphics[width=0.35\columnwidth]{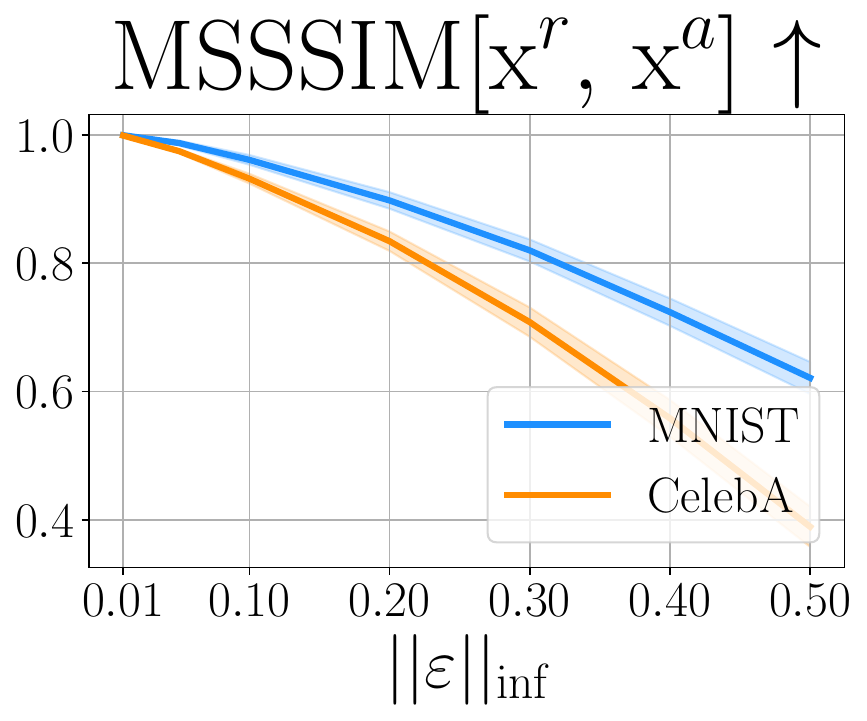} & \qquad
        \includegraphics[width=0.35\columnwidth]{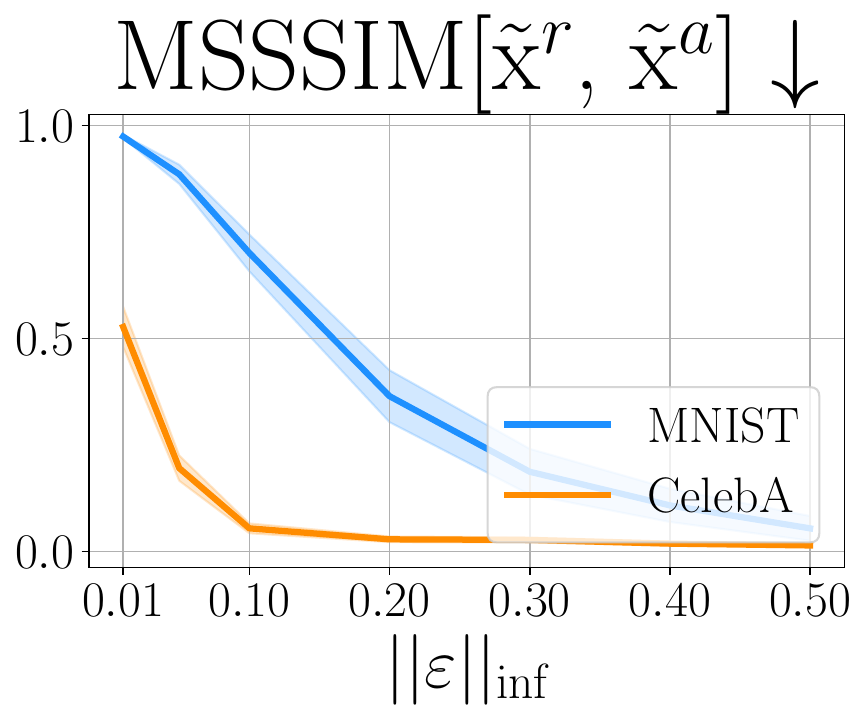} \\
        (a) Reference and adversarial point similarity &
        (b) Reconstruction similarity \\
    \end{tabular}
    \caption{Average images similarity (a) before it is passed to VAE and (b) after image is encoded and decoded back. We consider unsupervised attack on the encoder with the radiuses ranging from 0.01 to 0.5 for MNIST and CelebA datasets.}
    \label{fig:radius_metrics}
\end{figure}

\begin{figure}[h]
    \centering
    \resizebox{.99\textwidth}{!}{
    \begin{tabular}{c|ccccccc}
        $\rvx^r$ & \multicolumn{7}{|c}{$\rvx^a$} \\
        \includegraphics[width=0.11\linewidth]{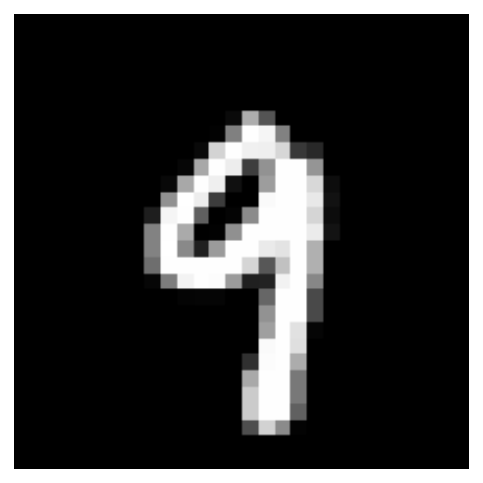} &
        \includegraphics[width=0.11\linewidth]{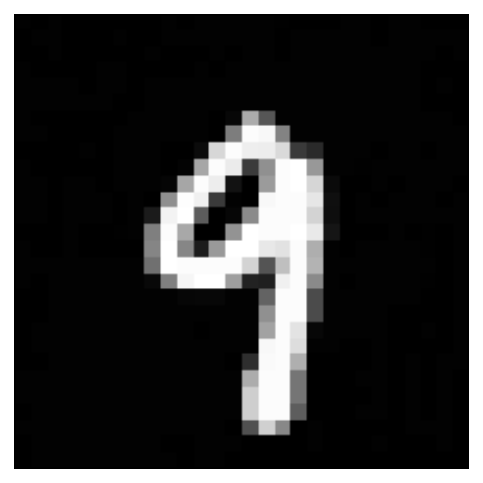} &
        \includegraphics[width=0.11\linewidth]{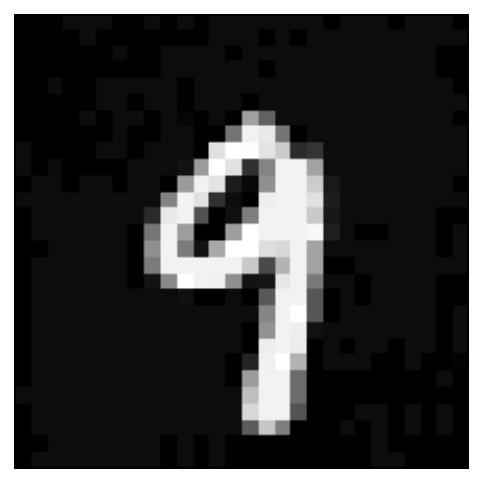} &
        \includegraphics[width=0.11\linewidth]{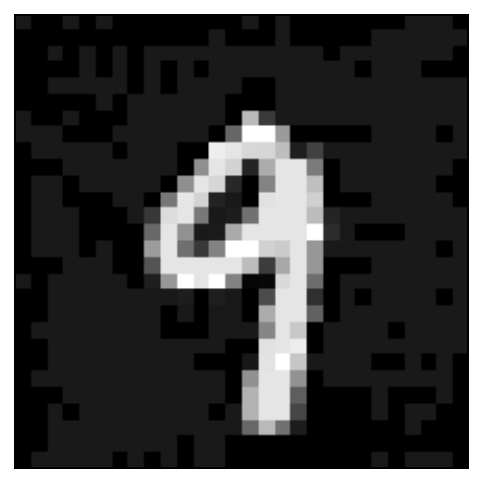} &
        \includegraphics[width=0.11\linewidth]{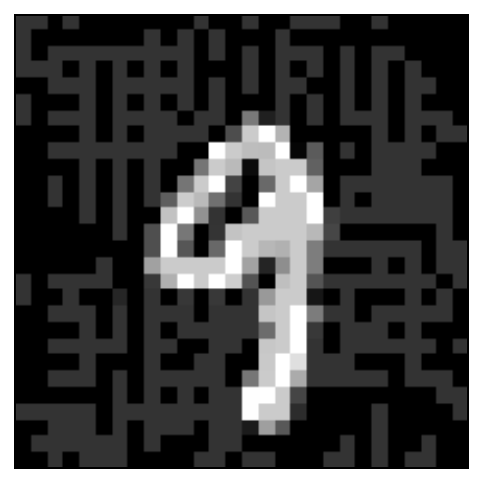} &
        \includegraphics[width=0.11\linewidth]{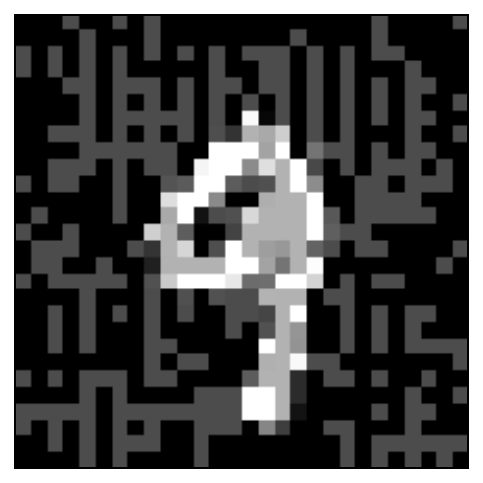} &
        \includegraphics[width=0.11\linewidth]{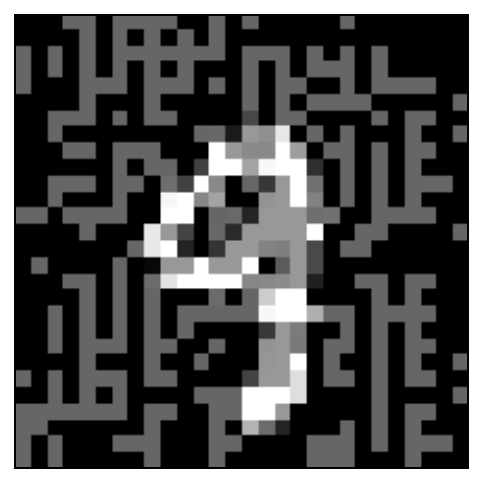} &
        \includegraphics[width=0.11\linewidth]{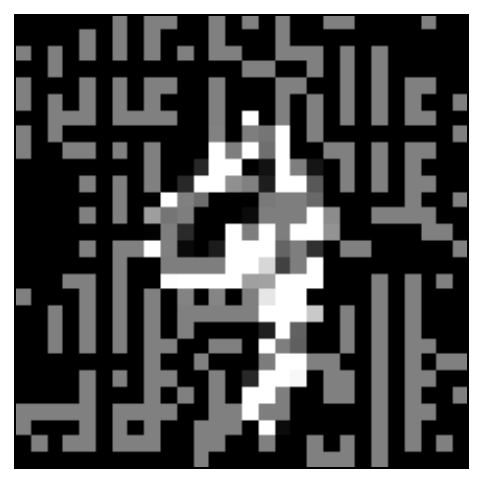} \\
        \includegraphics[width=0.11\linewidth]{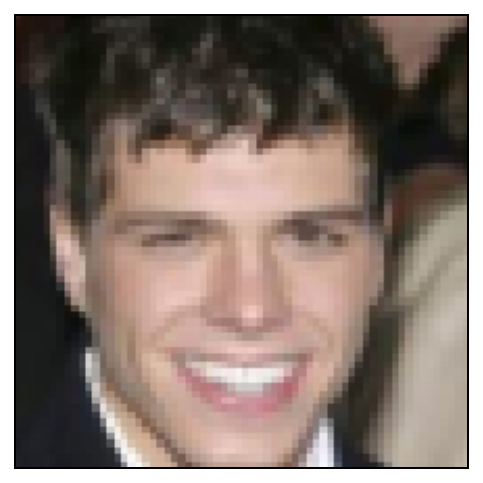} &
        \includegraphics[width=0.11\linewidth]{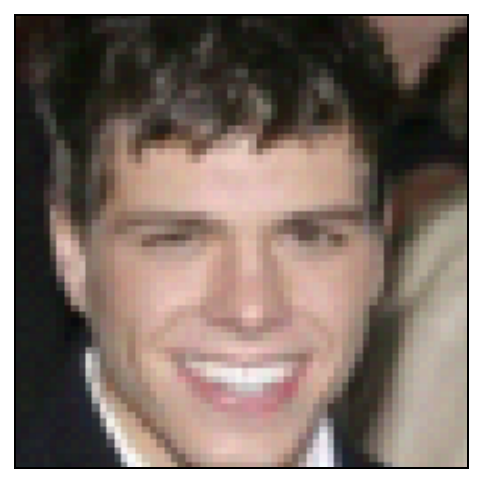} &
        \includegraphics[width=0.11\linewidth]{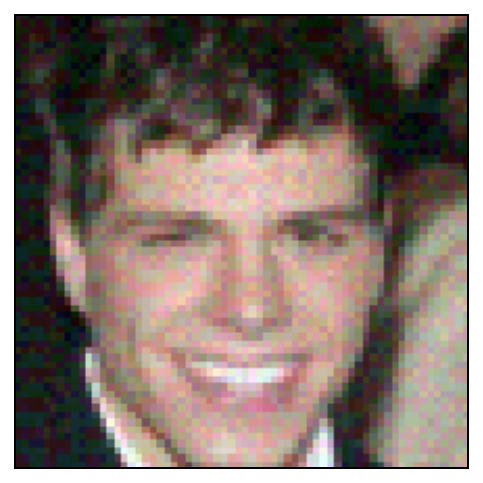} &
        \includegraphics[width=0.11\linewidth]{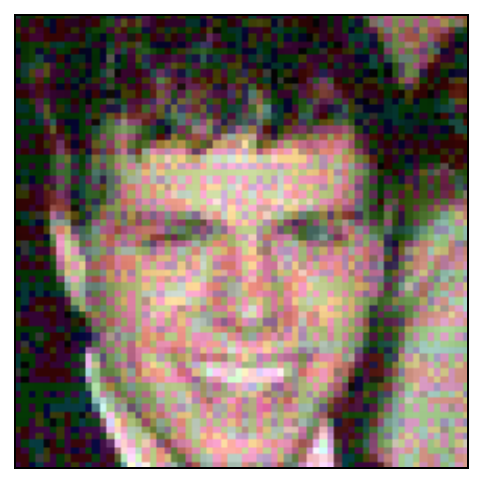} &
        \includegraphics[width=0.11\linewidth]{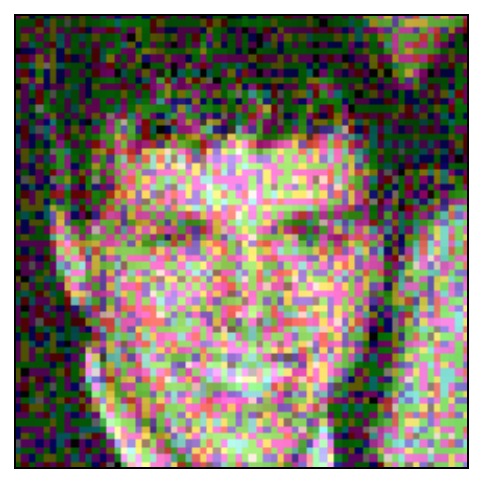} &
        \includegraphics[width=0.11\linewidth]{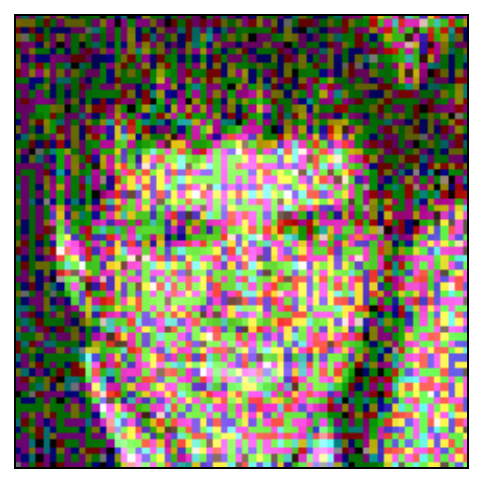} &
        \includegraphics[width=0.11\linewidth]{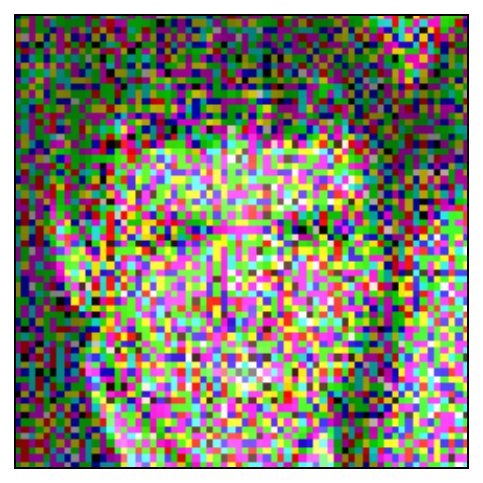} &
        \includegraphics[width=0.11\linewidth]{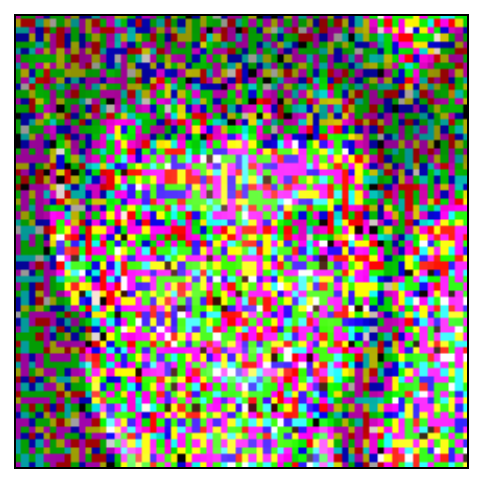} \\
        $\|\varepsilon\|_{\inf} = 0.0$ & $\|\varepsilon\|_{\inf} = 0.01$ & $\|\varepsilon\|_{\inf} = 0.05$  &  $\|\varepsilon\|_{\inf} = 0.1$ &  $\|\varepsilon\|_{\inf} = 0.2$& $\|\varepsilon\|_{\inf} = 0.3$ & $\|\varepsilon\|_{\inf} = 0.4$ &  $\|\varepsilon\|_{\inf} = 0.5$\\
    \end{tabular}}
    \caption{}
    \label{fig:radius_adversarial_examples}
\end{figure}

\newpage

\subsection{How many HMC steps are required for a defence?}\label{appendix:hmc_steps}
One of the main hyperparameters of the proposed approach is number of steps of MCMC that the defender does. We have conducted experiments with MNIST and Color MNSIT dataset to see how the robustness metrics change when we increase number of HMC steps from 0 to 200. As we can see from the Figure \ref{fig:hmc_step_metrics}, there is always a considerable jump between 0 steps (no defence) and 100 steps (lowest number of steps considered). However, as we continue making steps, we do not observe further improvement of the metrics. 
\begin{figure}[ht]
    \centering
    \begin{tabular}{cc}
        \includegraphics[width=0.35\columnwidth]{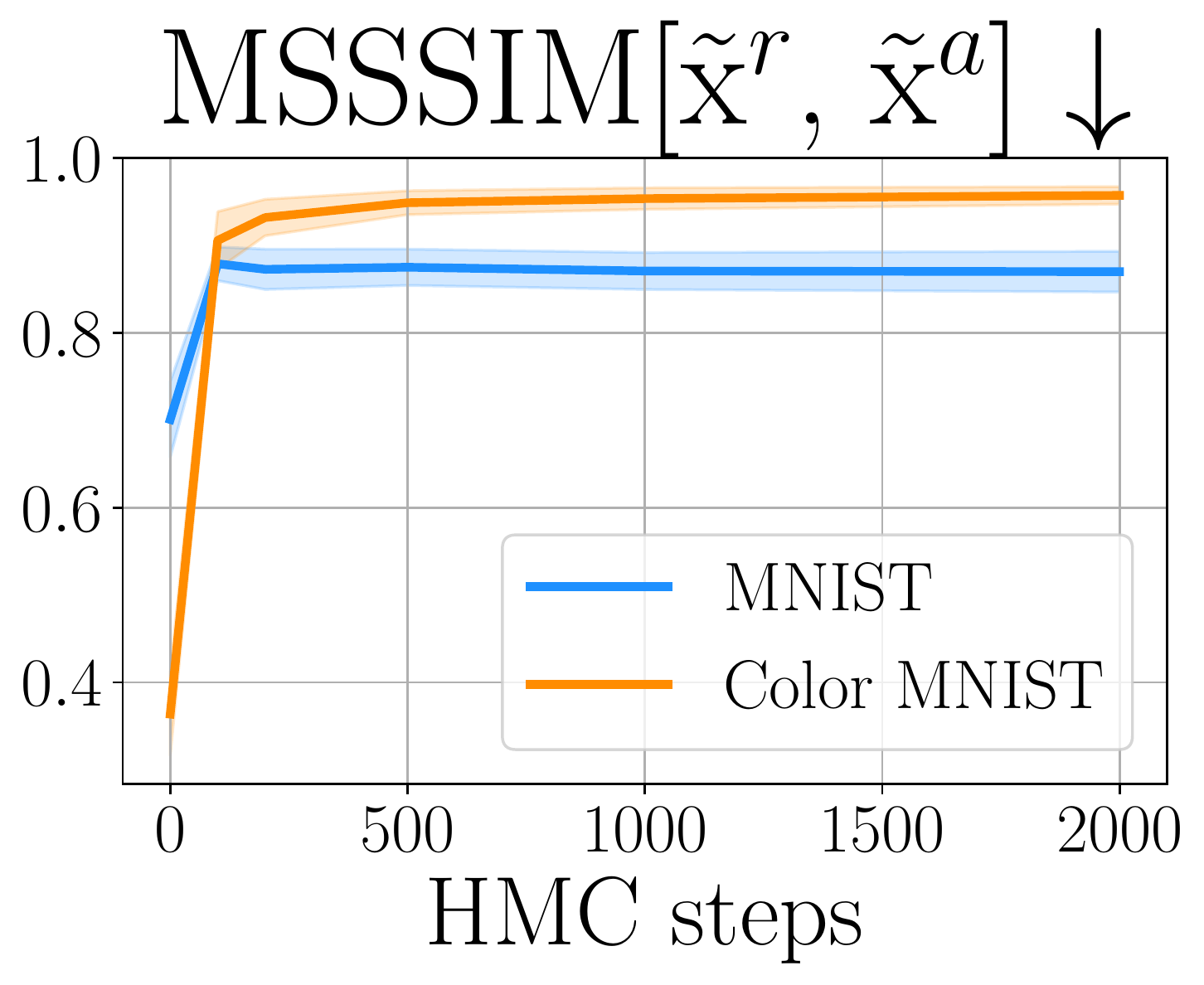} & \qquad
        \includegraphics[width=0.35\columnwidth]{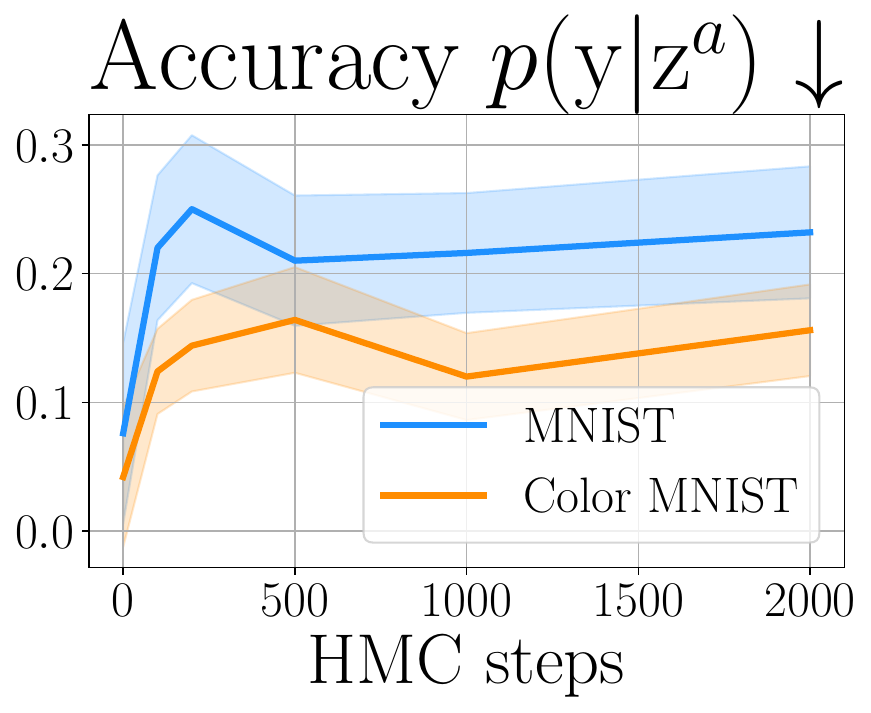} \\
        (a) Reconstruction similarity. &
        (b) Adversarial Accuracy (digit classification task). \\
    \end{tabular}
    \caption{Example of the reference point (leftmost column) and adversarial points for different raduises of the attack.}
    \label{fig:hmc_step_metrics}
\end{figure}
\newpage

\subsection{Comparison of objective functions} \label{appendix:objectives}
This section compares different objective functions that can be used to construct adversarial attacks on VAE. In general, in both supervised and unsupervised setting, we need to measure the difference between variational posterior in the adversarial point $q(\rvz|\rvx^a)$ and a point from the dataset (either a target or reference point). We consider a Gaussian encoder, and the simplest way to compare two Gaussian distributions is to measure the distance between their means. To take into account the variances, we can use the KL-divergence. It is a non-symmetric metric. Thus, we have two options: to use the forward or reverse KL. Finally, it is also possible to consider the symmetrical KL divergence that is an average between the two.

\begin{figure}[ht]
     \begin{subfigure}{0.47\columnwidth}
         \centering
         \includegraphics[width=0.68\columnwidth, valign=t]{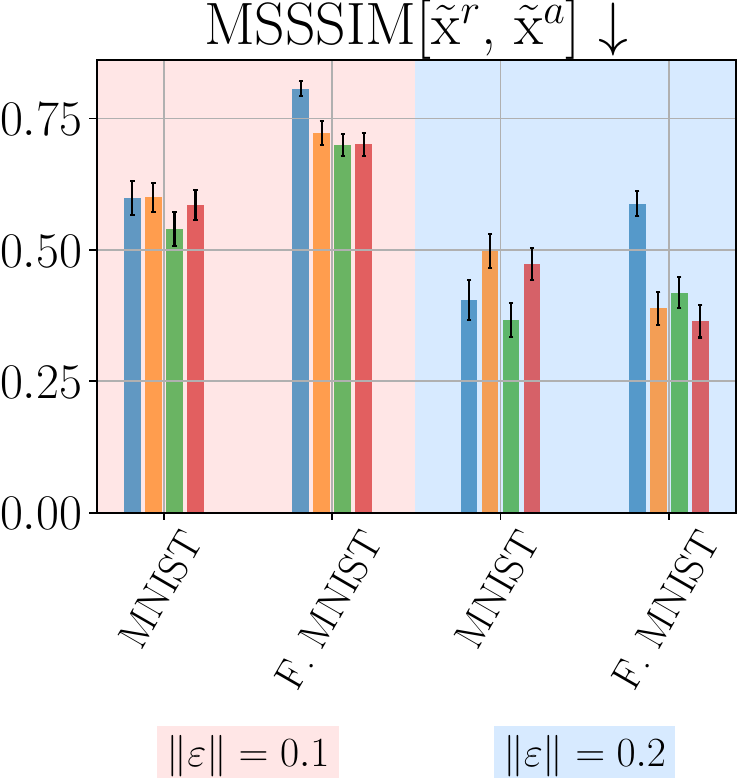}
         \caption{Reconstructions similarity: adversarial point and the reference.}
         \label{fig:objectives_0}
     \end{subfigure}%
     \begin{subfigure}{0.51\columnwidth}
         \centering
         \includegraphics[width=1.\columnwidth, valign=t]{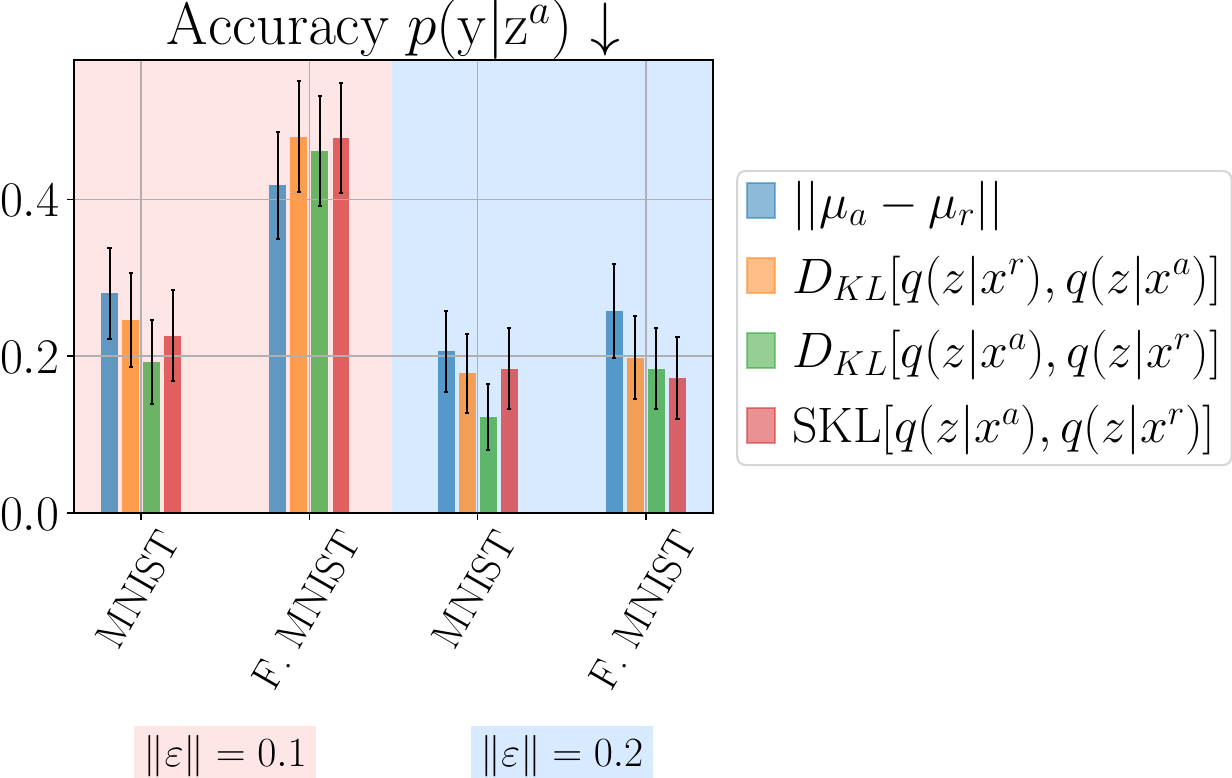}
         \caption{Adversarial Accuracy.}
         \label{fig:objectives_1}
     \end{subfigure}
     \vskip -5pt
\caption{Comparison of different objectives function used to train an attack. Arrows represent the direction of the successful attack.}
\label{fig:objectives}
\end{figure}

In Figure \ref{fig:objectives}, we measure how successful the attacks are in terms of the proposed metrics. We use arrows in the plot titles to indicate desirable values of the metric for a successful attack. We compare supervised and unsupervised attacks on VAE trained on MNIST and fashion MNIST datasets. We observe that there is no single objective function that consistently outperforms others.

\subsection{Inference Time} \label{appendix:inference_time}

\upd{
Even though our approach does not require changing the training procedure, it has influence on the inference time. In practice, this can be a limiting factor. Therefore, in Table \ref{tab:inferece_time} we report the computational overhead during the inference time. We measure the inference time in seconds per test point without HMC ($T=0$) and for different budgets ($T=\{100, 500, 1000\}$). 
}

\begin{table}[ht]
\begin{center}
\begin{small}
\begin{sc}
\caption{\upd{Inference wall-clock time of the VAE for various number of MCMC steps ($T$).}}
\label{tab:inferece_time}
\discussion{
\begin{tabular}{l|cccc}
\toprule
\multicolumn{1}{c|}{$T$} &  0 & 100 & 500 &  1000 \\
\midrule
 & \multicolumn{4}{c}{VAE} \\
MNIST & 0.0001 & 0.0099 & 0.0505 & 0.1011\\
Color MNIST & 0.0001& 0.0110 & 0.0553 & 0.1111\\
\midrule
& \multicolumn{4}{c}{NVAE}  \\
CelebA & 0.429 & 6.512 & 31.551 & 63.031\\
\bottomrule
\end{tabular}  
}
\end{sc}
\end{small}
\end{center}
\end{table}

\newpage

\section{Details of the experiments} \label{appendix:experimental_details}
\subsection{Training VAE models} \label{appendix:beta_vae_param}

\paragraph{Architecture} We use the same fully convolutional architecture with latent dimension 64 for MNIST, FashionMNISt and ColorMNIST datasets. In Table \ref{tab:mnist_arch}, we provide detailed scheme of the architecture. We use $\texttt{Conv(3x3, 1->32)}$ to denote convolution with kernel size $\texttt{3x3}$, $\texttt{1}$ input channel and  $\texttt{32}$ output channels. We denote stride of the convolution with $\texttt{s}$, padding with $\texttt{p}$ and dilation with $\texttt{d}$. The same notation applied for the transposed convolutions ($\texttt{ConvTranspose}$). ColorMNIST has 3 input channels, so the first convolutional layer in the encoder and the last of the decoder are slightly different. In this cases  values for ColorMNIST are report in parenthesis with the red color. 

\begin{table}[ht]
\caption{Convolutional architecture for VAE trained on MNIST, Fashion MNIST and ColorMNIST datasets.}
\label{tab:mnist_arch}
\begin{center}
\begin{tabular}{@{}lll@{}}
\toprule
                  & Encoder & Decoder  \\  \midrule
& \texttt{Conv(3x3, 1\textcolor{red}{(3)}->32, s=2, p=1)} & \texttt{ConvTranspose(3x3,64->128,s=1,p=0, d=2)} \\
& \texttt{ReLU()}& \texttt{ReLU()} \\
& \texttt{Conv(3x3, 32->64, s=2, p=1)} & \texttt{ConvTranspose(3x3,128->96,s=1,p=0)}\\
& \texttt{ReLU()} & \texttt{ReLU()}\\
& \texttt{Conv(3x3, 64->96, s=2, p=1)} &\texttt{ConvTranspose(3x3,96->64,s=1,p=1)} \\
& \texttt{ReLU()} &  \texttt{ReLU()} \\
& \texttt{Conv(3x3,96->128,s=2,p=1)} & \texttt{ConvTranspose(4x4,64->32,s=2,p=1)} \\
& \texttt{ReLU()} & \texttt{ReLU()} \\
& $\mu_z \leftarrow$  \texttt{Conv(3x3,128->64,s=2,p=1)} &  \texttt{ConvTranspose(4x4,31->1\textcolor{red}{(3)},s=2,p=1)} \\
& $\log \sigma^2_z \leftarrow$  \texttt{Conv(3x3,128->64,s=2,p=1)}&  $\mu_x \leftarrow$  \texttt{Sigmoid()} \textcolor{red}{(\texttt{Identity()})}\\ 
\bottomrule
\end{tabular}
\end{center}
\end{table}

\paragraph{Optimization} We use Adam to perform the optimization. We start from the learning rate $5e-4$ and reduce it by the factor of 2 if the validation loss does not decrease for 10 epochs. We train a model for 300 epochs with the batch size 128. In Table \ref{tab:mnist_vae_res}, we report the values of the test metrics for VAEs trained on MNIST, Fashion MNIST and Color MNIST. 

For calculating the FID score, we use \texttt{torchmetrics} library: \url{https://torchmetrics.readthedocs.io/en/latest/references/modules.html#frechetinceptiondistance}.

\begin{table}[ht]
\caption{Test performance of the $\beta$-VAE and $\beta$-TCVAE with different values of $\beta$.
Negative loglikelihood is estimated with importance sampling ($k = 1000$) as suggested in \cite{burda2015importance}.} 
\label{tab:mnist_vae_res}
\begin{center}
\begin{sc}
\begin{tabular}{ll|ccccccc}
\toprule
& & \multicolumn{2}{c}{MNIST} & \multicolumn{2}{c}{Fashion MNIST} & \multicolumn{3}{c}{Color MNIST} \\
& $\beta$ & $-\log p(\rvx)$    &  MSE &  $-\log p(\rvx)$    &  MSE &  $-\log p(\rvx)$    &  MSE & FID \\\midrule
&1    & \textbf{88.3} & 578.6 & \textbf{232.8} & 814.3 & \textbf{54.87}  & 261.3 & 2.09\\ \midrule
\small{\multirow{3}{*}{\STAB{\rotatebox[origin=c]{90}{$\beta$-VAE}}}}
&2    & 89.3  &  824.2   & 234.1 & 1021.1 &  55.6 & 365.6 & 2.4  \\
&5    & 100.6 &  1485.1  & 241.8 & 1457.8 &  63.6 & 586.1 & 2.5\\
&10   & 126.8 &  2498.9  & 248.7 & 1842.3 &  88.7 & 936.2 & 2.4\\ \midrule
\small{\multirow{3}{*}{\STAB{\rotatebox[origin=c]{90}{$\beta$-TCVAE}}} }
&2    & 89.3  &   828.4 & 233.6 & 980.4 & 55.8 & 366.4 & 3.0\\
&5    & 96.7  &  1325.4 & 238.2 & 1024.6 & 63.0 & 574.8 & 2.0\\
&10   & 107.2 &  1686.1 & 247.5 & 1570.0 & 76.5 & 806.2 & 2.2\\ \midrule
\end{tabular}
\end{sc}
\end{center}
\end{table}

\newpage
\subsection{Adversarial Attacks and Defence Hyperparameters} \label{appendix:vae_attack}
In Table \ref{tab:setup}, we report all the hyperparameter values that were used to attack and defend VAE models. 

In all the experiments we randomly select reference points from the test dataset. We also ensure that the resulting samples are properly stratified --- include an even number of points from each of the classes. For each reference point, we train 10 adversarial inputs with the same objective function but different initialization. 

We use projected gradient descent to learn the adversarial attacks. Optimization parameters were the same for all the datasets and models. They are presented in Table \ref{tab:setup}.

We choose HMC to defend the model against the trained attack. We perform $T$ steps of HMC with the step size $\eta$ and $L$ leapfrog steps. Where indicated, we adapt the step size after each step of HMC using the following formula:
\begin{equation}
    \eta_t = \eta_{t-1} + 0.01\cdot\frac{\alpha_{t-1} - 0.9}{0.9}\cdot\eta_{t-1},
\end{equation}
where $\alpha_t$ is the acceptance rate at step $t$. This way we increase the step size if the acceptance rate is higher than 90\% and decrease it otherwise. When adaptive steps size is used, a value in the table indicates the $\eta_0$.

\begin{table}[ht]
\caption{Full list of hyperparameters for attack construction and the defence.}
\label{tab:setup}
\begin{center}
\resizebox{.99\textwidth}{!}{
\begin{tabular}{llccc|cc}
\toprule
& & \multicolumn{3}{c}{VAE} & \multicolumn{2}{|c}{NVAE} \\
& &  MNIST   & Fashion MNIST & Color MNIST   & MNIST & CelebA\\\midrule
&\# of reference points & 50 & 50 & 50  & 50 & 20\\
&\# of adversarial points & 500 & 500 & 500 & 500 & 200 \\
&Radius norm ($\|\cdot\|_{p}$) & $\inf$ & $\inf$ &$\inf$&$\inf$&$\inf$ \\
&Radius & $\{0.1, 0.2\}$& $\{0.1, 0.2\}$& $\{0.1, 0.2\}$& $\{0.1, 0.2\}$& $\{0.05, 0.1\}$\\\midrule
\small{\multirow{4}{*}{\STAB{\rotatebox[origin=c]{0}{Optimization (PGD)}}}} &Optimizer &  \multicolumn{5}{c}{SGD}\\
 & Num. steps & \multicolumn{5}{c}{50}\\
 & $\varepsilon$ initialization & \multicolumn{5}{c}{$\mathcal{N}(0, 0.2\cdot I)$}\\
 & Learning rate (lr) & \multicolumn{5}{c}{$1$}\\\midrule
\small{\multirow{4}{*}{\STAB{\rotatebox[origin=c]{0}{Defence (HMC)}}}} 
& Num. steps ($T$) & 500 & 1000& 1000 & 2000 & 1000\\
& Step size $\eta $ & 0.1 & 0.05 & 0.05 & 1e-4 & 1e-4\\
& Num. Leapfrog steps ($L$) & 20 & 20 & 20 & 20 & 1\\
& Adaptive step size & True & True & True & True & False \\
\bottomrule
\end{tabular}}
\end{center}
\end{table}

\end{document}